\documentclass{elsarticle}
\usepackage{mathptmx}
\usepackage{amssymb}
\usepackage{amsmath}    
\usepackage{graphicx}
\usepackage{verbatim}
\usepackage{color} 

\usepackage[normalem]{ulem} 

\newtheorem{definition}{Definition}
\newtheorem{corollary}{Corollary}

\newtheorem{claim}{Proposition}
\newtheorem{example}{Example}
\newtheorem{lemma}{Lemma}

\usepackage{url, multicol} 

\usepackage{lineno,hyperref}
\modulolinenumbers[5]

\begin{document}

\newcommand{\hide}[1]{}
\newcommand{\otherquestions}[1]{}
\newcommand{\catom}[2]{(\{#1\}, \{#2\})}
\newcommand{\set}[1]{\{#1\}}
\newcommand{\pg}[1]{{\tt #1}}
\newcommand{\emptyclause}{\Box}

\def\st{\medskip\noindent}

\newcommand{\kleene}{\wedge}

\newcommand{\ee}[1] {
  \begin{enumerate}
    #1 
  \end{enumerate}
}

\newcommand{\ie}[1] {
  \begin{enumerate}
    #1 
  \end{enumerate}
}

\newcounter{statement}
\newcommand{\ml}[1]{
    \refstepcounter{statement}
    (\arabic{statement}) 
    \label{#1}
}
\newcommand{\mr}[1]{(\ref{#1})}

\begin{frontmatter}

\title{Vicious Circle Principle and Logic Programs with Aggregates
\footnote{This is an unpublished  long version of ``Vicious Circle Principle and Logic Programs with Aggregates" published in 2014 \cite{GelfondZ14}. It was completed in Apr 2015. The results on relation between $\mathcal{F}log$, $\mathcal{S}log$ and $\mathcal{a}log$ were independent of similar results reported in \cite{alviano2015stable}.}}



\author{Michael Gelfond and Yuanlin Zhang}
\address{Texas Tech University, Lubbock, Texas 79414, USA \\
\{michael.gelfond, y.zhang\}@ttu.edu}

\begin{abstract}
    The paper presents a knowledge representation language
    $\mathcal{A}log$ which extends ASP with aggregates. The goal is to
    have a language based on simple syntax and clear intuitive and
    mathematical  semantics. We give some properties of
    $\mathcal{A}log$, an algorithm for computing its answer sets, and
    comparison with other approaches.
\end{abstract}


\end{frontmatter}


\section{Introduction}
The development of answer set semantics for logic programs \cite{gl88,gl91} led 
to the creation of powerful knowledge representation language,
Answer Set Prolog (ASP), capable of representing recursive definitions,
defaults, effects of actions and other important phenomena of natural 
language. The design of algorithms for computing answer sets and their 
efficient implementations in systems called \emph{ASP solvers} \cite{nss02,dlv06,clasp07} allowed 
the language to become a powerful tool for building 
non-trivial knowledge intensive applications  
\cite{BrewkaET11,ErdemLL12}. 
There are a 
number of extensions of the 
ASP
which  also
contributed to this success. This paper is about one such extension --
\emph{logic programs with aggregates}. 

By \emph{aggregates} we mean
(possibly partial) functions 
defined on sets of objects of the domain.
Here is a typical example.
\begin{example}[Classes That Need Teaching Assistants]\label{e1}
Suppose that we have a complete list of students enrolled in 
a class $c$ that is represented by the following collection of 
atoms:
\begin{verbatim}
enrolled(c,mike). 
enrolled(c,john). 
... 
\end{verbatim}
Suppose also that we would like to define a new relation $need\_ta(C)$
that holds iff the class $C$ needs a teaching assistant. In this 
particular school $need\_ta(C)$ is true iff the number of students 
enrolled in the class is greater than $20$. The definition can be 
given by a simple rule in the language of logic programs with 
aggregates:
$$need\_ta(C) \leftarrow card\{X : enrolled(C,X)\} > 20$$
where $card$ stands for the cardinality function\footnote{
  A more accurate notation would be $card(\{X :
    enrolled(C,X)\}) $ but parentheses are skipped for readability.
}. Let us call the 
resulting program $P_0$. 
\end{example}
The program is simple, has a clear intuitive meaning, and can be run 
on some of the existing ASP solvers. However, the situation is more 
complex than that. 
Unfortunately,
currently there is no \emph{the} language of logic programs with 
aggregates. Instead there is a comparatively large collection of such 
languages with different syntax and, even more importantly, different 
semantics \cite{pdb07,nss02,SonP07,FaberPL11,gel02,KempS91}.
As an illustration consider the following example:
\begin{example}\label{e2}

Let $P_1$ consist of the following rule:
$$p(a) \leftarrow card\{X : p(X)\} = 1.$$
Even for this seemingly simple program, there are different opinions 
about its meaning. According to \cite{FaberPL11} the program has one answer 
set $A = \{\ \}$; according to \cite{gel02,KempS91} it has two answer sets: $A_1=\{\ \}$
and $A_2=\{p(a)\}$. 
\end{example}
In our judgment this and other similar ``clashes of intuition'' cause 
a serious impediment to the use of aggregates for knowledge 
representation and reasoning. 
In this paper we aim at addressing this problem by suggesting yet another logic 
programming language with aggregates, called $\mathcal{A}log$, which is based on the 
following design principles:
\begin{itemize}
\item the language should have a simple syntax and intuitive semantics 
  based on understandable informal principles, and 
\item  the informal semantics should have clear and elegant 
  mathematics associated with it. 
\end{itemize}
In our opinion existing extensions of ASP by aggregates 
often do not have clear intuitive principles underlying the semantics 
of the new constructs. 
Moreover, some of these languages violate such original foundational 
principles of ASP as the rationality principle. 
The problem is compounded by the fact 
that some of the semantics of aggregates use rather non-trivial 
mathematical constructions which makes it difficult to understand and explain their 
intuitive meaning. 


The semantics of $\mathcal{A}log$ is based on \emph{Vicious Circle 
  Principle} (VCP):  \emph{no object or property can be introduced by the 
  definition referring to the totality of objects satisfying this 
  property}. 
According to Feferman \cite{fef} the principle was first formulated by 
Poincare \cite{poin1906} in his analysis of paradoxes of set theory. 
Similar ideas were already successfully used in a collection of logic 
programming definitions of stratification including that of stratified 
aggregates (see, for instance, \cite{FaberPL11}. Unfortunately, 
limiting the language to stratified aggregates
eliminates some of the useful forms of circles (see Example \ref{e9} below). 
In this paper we give a new form of VCP which goes beyond stratification: 
 \emph{$p(a)$ cannot be introduced by the 
  definition referring to a set of objects satisfying $p$ if this set can contain $a$.} 
Technically, the principle is incorporated in our new 
definition of answer set (which coincides with the original definition for programs without aggregates). 
The definition is short and 
simple. We hope that, combined with a number of informal examples, it 
will be sufficient for developing an intuition necessary for the use 
of the language. 
The paper is organized as follows. 
In Section~\ref{alog}, we define the syntax and semantics of 
$\mathcal{A}log$.  We give some properties of $\mathcal{A}log$ programs in Section~\ref{prop} 
and present an algorithm for computing an 
answer set of an $\mathcal{A}log$ program in Section~\ref{solver}. 
A comparison with the existing work is done in Section~\ref{comp}.
In  Section~\ref{sec:set} we illustrate the elaboration tolerance of
our definitions by expanding $\mathcal{A}log$ by set constructs
playing the role similar to that of choice rules \cite{nss02} and conditional
literals \cite{lparse,syrjanen2009logic}.
We conclude the paper in Section~\ref{conclusion}. 

The first version of $\mathcal{A}log$ first appeared in the conference
proceedings \cite{GelfondZ14}.
The current paper contains a substantial amount of new material:

\begin{itemize}
\item The syntax and semantics of the original $\mathcal{A}log$
are expanded to allow aggregates defined on infinite sets, 
partially defined aggregates, and program rules constructed from
infinite collections of literals.
\item Mathematical properties of the original  $\mathcal{A}log$ are generalized to
  this case and substantial new results are obtained establishing
  relationships between $\mathcal{A}log$ and other logic programming
  languages with aggregates. Proof of correctness of the algorithm
  from Section \ref{solver} is presented.
\item Elaboration tolerance of the language design is illustrated by
  extending the language to include other set based constructs which
  are functionally similar to the choice rules and
  conditional literals but, in our opinion, have simpler
  syntax and semantics. 

\end{itemize}

\section{Syntax and Semantics of $\mathcal{A}log$}\label{alog}
We start with defining the syntax and intuitive semantics of the language. 
\subsection{Syntax}

Let $\Sigma$ be a (possibly sorted) signature with a finite collection 
of predicate and function symbols and (possibly infinite) collection of
object constants, and let $\mathcal{A}$ be a finite 
collection of symbols used to denote functions from
sets of 
terms of $\Sigma$ into integers.
Terms and literals over signature 
$\Sigma$ are defined as usual and referred to as \emph{regular}. 
Regular terms are called \emph{ground} if they contain no variables 
and no occurrences of symbols for arithmetic functions. Similarly for 
literals. 
An \emph{aggregate term} is an expression of 
the form 
\begin{equation}\label{agg-term}
f\{\bar{X}:cond\}  
\end{equation}
where $f \in \mathcal{A}$, $cond$ is a finite collection of regular 
literals, and $\bar{X}$ is a list of variables occurring in $cond$. 
We refer to an expression
\begin{equation}\label{set-name}
\{\bar{X}:cond\}  
\end{equation}
as a \emph{set name}.
An occurrence of a variable from $\bar{X}$ in (\ref{set-name}) 
is called \emph{bound} within (\ref{set-name}).
If the condition from 
(\ref{set-name}) contains no variables except those in $\bar{X}$ then 
it is read as \emph{the set of all objects of the 
  program satisfying $cond$}.  If $cond$ contains other 
variables, say $\bar{Y}=\langle Y_1,\dots,Y_n\rangle$, then 
$\{\bar{X}:cond\}$ defines the function mapping possible values 
$\bar{c} = \langle c_1,\dots,c_n\rangle$ of these variables into sets 
$\{\bar{X}:cond|^{\bar{Y}}_{\bar{c}}\}$ where 
$cond |^{\bar{Y}}_{\bar{c}}$ is the result of replacing 
$Y_1,\dots,Y_n$ by $c_1,\dots,c_n$. 

\noindent
By an \emph{aggregate atom} we mean an 
expression of the form 
\begin{equation}\label{agg-atom}
\langle aggregate\_term \rangle \langle arithmetic\_relation \rangle 
\langle arithmetic\_term  \rangle 
\end{equation}
where $arithmetic\_relation$ is $>, \geq, <, \leq,=$ or !=,  and  
$arithmetic\_term$
is constructed from variables and integers using 
arithmetic operations, $+$, $-$, $\times$, etc.  
An \emph{aggregate literal} is an aggregate atom, $A$, or its
negation, $\neg A$. Note that every negative aggregate literal can be
replaced by an equivalent aggregate atom, called its \emph{positive form}.
For instance, $f\{\bar{X}:cond\}  \leq N$ is the positive form of
$\neg f\{\bar{X}:cond\}  > N$. Similarly for other arithmetic
relations.

\st 
Regular and aggregate atomes are referred to as
  \emph{atoms}. Similarly for literals.

\noindent 
By \emph{e-literals} we mean literals possibly preceded by 
default negation $not$. The latter (former) are called \emph{negative}
(\emph{positive}) e-literals.

\st
A \emph{rule} of 
$\mathcal{A}log$ is an expression of the form 
\begin{equation}\label{rule}
head \leftarrow pos,neg 
\end{equation}
where $head$ is a disjunction of regular literals, $pos$ and $neg$ are 
collections of literals and literals preceded by 
$not$ respectively; as usual, $pos \cup neg$ will be referred to as
the \emph{body} of the rule. 
Note, that both, head and body can be infinite. 
All parts of the rule, including $head$, can be empty. 
An occurrence of a variable in (\ref{rule}) not bound within any set name
in this rule is called 
\emph{free} in (\ref{rule}). 
A rule of $\mathcal{A}log$ is called \emph{ground} 
if it contains no occurrences of free variables and no occurrences of arithmetic functions.

 \st 
A \emph{program} of $\mathcal{A}log$ is a collection of 
$\mathcal{A}log$'s rules. A program is \emph{ground} if its rules 
are ground. 

\st 
As usual for ASP based languages, 
rules of $\mathcal{A}log$ program with variables are 
viewed as collections of their ground instantiations. 
A \emph{ground instantiation} of a rule $r$ is the program 
obtained from $r$ by 
replacing free occurrences of variables in $r$ by 
ground terms of $\Sigma$ and evaluating all arithmetic functions. 
If the signature $\Sigma$ is sorted (as, for 
instance, in \cite{BalaiGZ13}) the substitutions should respect sort 
requirements for predicates and functions.

\noindent
Clearly the grounding of an
$\mathcal{A}log$ program is a ground program. 
The following examples 
illustrate the definition:

\begin{example}[Grounding: all occurrences of the set variable are bound] \label{e3}
Consider a program $P_2$ with variables:

\begin{verbatim}
q(Y) :- card{X:p(X,Y)} = 1, r(Y). 
r(a).  r(b).  p(a,b). 
\end{verbatim}
Here all occurrences of a set variable $X$ are bound; all occurrences 
of a variable $Y$ are free. 
The program's grounding, $ground(P_2)$, is 

\begin{verbatim}
q(a) :- card{X:p(X,a)} = 1, r(a). 
q(b) :- card{X:p(X,b)} = 1, r(b). 
r(a).  r(b).  p(a,b). 
\end{verbatim}
\end{example}
The next example deals with the case when some occurrences of the set
variable in a rule are free and some are bound.  

\begin{example}[Grounding: some occurrences of a set variable are free]\label{e4}
Consider an $\mathcal{A}log$ program $P_3$
\begin{verbatim}
r :- card{X:p(X)} >= 2, q(X). 
p(a).  p(b).  q(a). 
\end{verbatim}
Here the occurrence of $X$ in $q(X)$ is free. Hence the ground program 
$ground(P_3)$ is:
 \begin{verbatim}
r :- card{X:p(X)} >= 2, q(a). 
r :- card{X:p(X)} >= 2, q(b). 
p(a).  p(b).  q(a). 
\end{verbatim}
\end{example}

\subsection{Semantics}
To define the semantics of $\mathcal{A}log$ programs we first notice
that the \emph{standard definition of answer set from \cite{gl88} is
applicable to programs containing infinite rules}. Hence we can talk
about answer sets of $\mathcal{A}log$ programs not containing aggregates.
Now we expand this definition of answer sets to $\mathcal{A}log$ programs
with aggregates. The resulting 
definition captures the rationality principle - \emph{believe 
  nothing you are not forced to believe} \cite{GelK13} - and avoids vicious circles. 
As usual the definition of answer set is given for ground 
programs. 

We will need the following terminology: Let $S$ be a set of ground
regular literals and $f\{X:p(X)\} >n$  be a ground aggregate atom.
\begin{itemize}
\item $f\{X:p(X)\} >n$ is \emph{true} in 
$S$ if $f\{t: p(t) \in S\} > n$; 
\item $f\{X:p(X)\} >n$ is \emph{false}
in $S$ if  $f\{t: p(t) \in S\} \leq n$;
\item otherwise $f\{X:p(X)\} >n$ is
\emph{undefined}. 
\end{itemize}
Similarly for other arithmetic relations. 

If $A$
is a ground aggregate literal then
\begin{itemize}
\item $(\mbox{not } A)$ is \emph{true} in $S$
if $A$ is \emph{false} or
\emph{undefined} in $S$. 
\item Otherwise
$(\mbox{not } A)$ is \emph{false} in $S$.
\end{itemize}
A collection $U$ of e-literals is 
\begin{itemize}
\item \emph{true} in $S$ if 
every e-literal from $U$ is \emph{true} in $S$,
\item \emph{false} in $S$ if $U$ contains
an e-literal which is \emph{false} in $S$,
\item \emph{undefined} otherwise.
\end{itemize}
A disjunction of literals is true in $S$ if at least one of its
elements is \emph{true} in $S$.

\st
Finally a rule is \emph{satisfied} by $S$ if its head is \emph{true}
in $S$ or its body is not \emph{true} in $S$.

\begin{definition}[Aggregate Reduct]\label{reduct}
The \emph{aggregate reduct} of a ground program $\Pi$ 
of $\mathcal{A}log$ with respect to a set of 
ground regular literals $S$ is obtained from $\Pi$ by 
\begin{enumerate}
\item removing all rules containing aggregate e-literals
  which are false or undefined in $S$. 
\item removing all aggregate e-literals of the form ($\mbox{not } A$)
  such that  $A$ is undefined in $S$.
\item replacing all remaining aggregate e-literals of the form
($\mbox{not } A$) by $\neg A$ and, if necessary, removing double negation $\neg
\neg$.
\item replacing all negative aggregate literals by their positive form.
\item replacing every remaining aggregate atom $f\{X:p(X)\} \odot  n$ by 
 the set $\{p(t) : p(t) \in S\}$, where $p(t)$ is the result of 
  replacing variable $X$ by ground term $t$;
$\{p(t) : p(t) \in S\}$ is called the \emph{reduct} of $f\{X:p(X)\} \odot  n$  with respect to $S$. 
\end{enumerate}
\end{definition}
The first clause of the definition removes rules useless because of
the truth values of their aggregates.
The next three clauses deal with e-literals of the form ($\mbox{not }
A$) which are \emph{true} in $S$. There are two distinct cases: $A$ is
\emph{undefined} in $S$ and $A$ is \emph{false} in $S$. The former is dealt
with by simply removing  ($\mbox{not } A$) from the rule. The latter
requires more refined treatment in which ($\mbox{not } A$) is
replaced by an equivalent aggregate \emph{atom}
\footnote{Note, that
this is necessary for 
checking if the resulting rule avoids vicious circles.}. 
The 
last clause of the definition reflects the principle of avoiding 
vicious circles -- a rule with aggregate atom $f\{X:p(X)\} \odot  n$
in the body can only be used if ``the totality'' of all objects satisfying 
$p$ has already being constructed. Attempting to apply this rule to 
define $p(t)$ will either lead to contradiction or to turning the rule 
into tautology (see Examples \ref{e7} and \ref{e9}). 

Clearly, aggregate reducts do not contain aggregates. Now we can give
our main definition.
 

\begin{definition}[Answer Set]\label{ans-set}
A set $S$ of ground regular literals over the signature of a ground
program $\Pi$ of $\mathcal{A}log$ is an 
\emph{answer set} of $\Pi$ if $S$ is an answer set of the aggregate 
reduct of $\Pi$ with respect to $S$. 
\end{definition}
We will illustrate this definition by a number of examples. 
\begin{example}[Example \ref{e3} Revisited] \label{e5}
Consider grounding 
\begin{verbatim}
q(a) :- card{X:p(X,a)} = 1, r(a). 
q(b) :- card{X:p(X,b)} = 1, r(b). 
r(a).  r(b).  p(a,b). 
\end{verbatim}
of program $P_2$ from Example \ref{e3}. 
It is easy to see that, due to the first clause of Definition \ref{reduct},
the aggregate reduct of the program with 
respect to any set $S$ of ground literals not containing $p(a,b)$ consists of the program 
facts, and hence $S$ is not an answer set of $P_2$. However the program's 
aggregate reduct with respect to 
$A = \{q(b),r(a),r(b),p(a,b)\}$ 
consists of the program's facts and the rule 
\begin{verbatim}
q(b) :- p(a,b),r(b)  
\end{verbatim}
obtained by the fifth clause of Definition \ref{reduct}.
Hence $A$ is an answer set of $P_2$. 
\end{example}

\begin{example}[Example \ref{e4} Revisited] \label{e6}
Consider now the grounding 
 \begin{verbatim}
r :- card{X:p(X)} >= 2, q(a). 
r :- card{X:p(X)} >= 2, q(b). 
p(a).  p(b).  q(a). 
\end{verbatim}
of program $P_3$ from Example \ref{e4}. 
Any answer set $S$ of this program must contain its facts. Hence 
$\{X: p(X) \in S\} = \{a,b\}$. $S$ satisfies the body of the first 
rule and must also contain $r$. Indeed, due to clause five 
of Definition \ref{reduct},
the aggregate reduct of $P_3$
with respect to $S = \{p(a), p(b), q(a), r\}$ consists of the facts of 
$P_3$ and the rules 
\begin{verbatim}
r :- p(a),p(b),q(a). 
r :- p(a),p(b),q(b). 
\end{verbatim}
Hence $S$ is the answer set of $P_3$. 
\end{example}
Neither of the two examples above contained recursion through aggregates
and hence did not
require the application of VCP.
The next example shows how this 
principle influences our definition of answer sets and hence our reasoning. 
\begin{example}[Example \ref{e2} Revisited]\label{e7}
Consider a program $P_1$ from Example \ref{e2}. 
The program, consisting of a rule 
\begin{verbatim}
p(a) :- card{X : p(X)}=1 
\end{verbatim}
is grounded. It has two candidate answer sets, $S_1 = \{\ \}$ and $S_2 
= \{p(a)\}$. The aggregate reduct of the program with respect to $S_1$
is the empty program. Hence, $S_1$ is an answer set of $P_1$. 
The program's aggregate reduct with respect to $S_2$ however is 
\begin{verbatim}
p(a) :- p(a). 
\end{verbatim}
The answer set of this reduct is empty and hence $S_1$ is the only 
answer of $P_1$. 
\end{example}
Example \ref{e7} shows how the attempt to define $p(a)$ in terms of 
totality of $p$ turns the defining rule into a tautology. 
The next example shows how it can lead to inconsistency of a program. 
\begin{example}[Vicious Circles through Aggregates and 
  Inconsistency] \label{e8}
Consider a program $P_4$:
\begin{verbatim}
p(a). 
p(b) :- card{X:p(X)} > 0. 
\end{verbatim}
Since every answer set of the program must contain $p(a)$,
the program has two candidate answer sets: $S_1 = \{p(a)\}$ and 
$S_2 = \{p(a),p(b)\}$. The aggregate reduct of $P_4$ with respect to 
$S_1$ is 
\begin{verbatim}
p(a). 
p(b) :- p(a). 
\end{verbatim}
The answer set of the reduct is $\{p(a),p(b)\}$ and hence $S_1$ is not 
an answer set of $P_4$. The reduct of $P_4$ with respect to $S_2$ is 
\begin{verbatim}
p(a). 
p(b) :- p(a),p(b). 
\end{verbatim}
Again its answer set is not equal to $S_2$ and hence $P_4$ is 
inconsistent (i.e., has no answer sets). The inconsistency is the direct 
result of an attempt to violate the underlying principle of the 
semantics. 
Indeed, the definition of $p(b)$ refers to the set of objects
satisfying $p$. This set can contain $b$ which is prohibited by our version
of VCP.

One can, of course, argue that $S_2$ can be viewed as 
a reasonable collection of beliefs which can be formed by a rational 
reasoner associated with $P_4$. After all, we do not need the totality 
of $p$ to satisfy the body of the rule defining $p(b)$. It is 
sufficient to know that $p$ contains $a$. This is indeed true but 
this reasoning depends on the knowledge which is not directly 
incorporated in the definition of $p(b)$. 
If one were to replace $P_4$ by 
\begin{verbatim}
p(a). 
p(b) :- card{X:p(X), X != b} > 0. 
\end{verbatim}
then, as expected, the vicious circle principle will not be violated 
and the program will have unique answer set $\{p(a),p(b)\}$. 
\end{example}
Next we give a simple but practical example of a program 
which allows recursion through aggregates but avoids vicious circles. 
\begin{example}[Defining Digital Circuits]\label{e9}
Consider part of a logic program formalizing propagation of binary signals 
through simple digital circuits. We assume that a circuit does not 
have a feedback, i.e., a wire receiving a signal from a gate cannot be 
an input wire to this gate. 
The program may contain a simple 
rule 
\begin{verbatim}
  val(W,0) :-
        gate(G, and),
        output(W, G),
        card{W: val(W,0), input(W, G)} > 0. 
\end{verbatim}
(partially) describing propagation of symbols through an \emph{and}
gate. Here $val(W,S)$ holds iff the digital signal on a wire $W$ has 
value $S$.  Despite its recursive nature the definition of $val$
avoids vicious circle. To define the signal on an output wire $W$ of an 
\emph{and} gate $G$ one needs to only construct a particular subset of 
input wires of $G$. Since, due to absence of feedback in our circuit,
$W$ can not belong to the latter set our definition is reasonable. 
To illustrate that our definition of answer set produces the intended 
result let us consider program $P_5$ consisting of the above rule and a 
collection of facts:
\begin{verbatim}
gate(g, and). 
output(w0, g). 
input(w1, g). 
input(w2, g).  
val(w1,0). 
\end{verbatim}

\noindent The grounding, $ground(P_5)$, of $P_5$ consists of
the above facts and the three rules of the form
\begin{verbatim}
  val(w,0) :-
        gate(g, and),
        output(w, g),
        card{W: val(W,0), input(W, g)} > 0. 
\end{verbatim}
where $w$ is $w_0$, $w_1$ ,and $w_2$.

\noindent Let $S=\{gate(g, and), val(w1, 0), val(w0,0), output(w0, g), input(w1, g),
  input(w2, g)\}$. 
The aggregate reduct of $ground(P_5)$
with respect to $S$ is the 
collection of facts  and the rules 
\begin{verbatim}
  val(w,0) :-
        gate(g, and),
        output(w, g),
        input(w1, g), 
        val(w1, 0).         
\end{verbatim}
where $w$ is $w_0$, $w_1$, and $w_2$.

\noindent The answer set of the reduct is $S$ and hence $S$ is an 
answer set of $P_5$. As expected it is the only answer set. 
(Indeed it is easy to see that other candidates 
do not satisfy our definition.) 
 
\end{example}

Next several example demonstrate the behavior of our semantics for
aggregates defined on infinite sets and for partially defined aggregates.
\begin{example}[Aggregates on Infinite Sets] \label{new1}
Consider a program $E_1$ consisting of the following rules:

\begin{verbatim}
even(0). 
even(I+2) :- even(I). 
q :- min{X : even(X)} = 0. 
\end{verbatim}

\noindent 
It is easy to see that the program has one answer set,
$S_{E_1} = \{q,0,2,\dots\}$. Indeed, the reduct of $E_1$ with respect to 
$S_{E_1}$ is the infinite collection of rules 

\begin{verbatim}
even(0). 
even(2) :- even(0). 
even(4) :- even(2). 
... 
q :- even(0),even(2),even(4)... 
\end{verbatim}

\noindent
The last rule has the infinite body constructed in the last step of 
Definition \ref{reduct}. Clearly, $S_{E_1}$ is a subset minimal collection 
of ground literals satisfying the rules of the reduct (i.e. its answer
set). Hence  $S_{E_1}$ is an answer set of $E_1$.
\end{example}
\begin{example}[Programs with Undefined Aggregates]\label{new2} 
Now consider a program $E_2$ consisting of the rules:
\begin{verbatim} 
even(0).  
even(I+2) :- even(I).  
q :- count{X : even(X)}> 0. 
\end{verbatim}

\noindent
This program has one answer set, $S_{E_2}= \{even(0),even(2),\dots\}$.  
To see that notice that every set satisfying the rules of $E_2$ must
contain even numbers. Since our aggregates range over natural
numbers, the aggregate $count$ is not defined on the set $count\{X :
even(X)\}$. This means that the body of the last rule is
undefined. According to clause one of Definition \ref{reduct} this rule is removed.
The reduct of $E_2$ with respect to $S_{E_2}$ is

\begin{verbatim}
even(0). 
even(2) :- even(0). 
even(4) :- even(2). 
... 
\end{verbatim}   
and hence $S_{E_2}$ is the answer set of $E_2$\footnote{Of course this
  is true only because of our (somewhat arbitrary) decision to limit
  aggregates of $\mathcal{A}log$ to those ranging over natural
  numbers. We could, of course, allow aggregates mapping sets into
  ordinals. In this case the body of the last rule of $E_2$ will be
  defined and the only answer set of $E_2$ will be $S_{E_1}$.}.
\end{example}

\noindent Next example combines undefined aggregates with the
default negation:

\begin{example}[Undefined Aggregates and Default Negation]\label{new3}
Consider a program $E_3$ consisting of the rules:

\begin{verbatim}
even(0). 
even(I+2) :- even(I). 
q :- not count{X : even(X)} > 0. 
\end{verbatim}

\noindent
This program has one answer set, 
$S_{E_3}= \{q,even(0),even(2),\dots\}$. 
To see that note that $count\{X : even(X)\}$ is undefined, and hence,
by the definition of
the reduct, the body of the last rule will be removed. The reduct is

\begin{verbatim}
even(0). 
even(2) :- even(0). 
even(4) :- even(2). 
... 
q.
\end{verbatim}   
and hence $S_{E_3}$ is an answer set of $E_3$. 
Clearly, the program  
has no other answer sets.  
\end{example}

\begin{example}[Undefined Aggregates and Default Negation (continued)]\label{new4}
Consider a program $E_4$ consisting of the rules:

\begin{verbatim}
p(1) :- not p(0). 
p(I+1) :- p(I). 
p(0) :- not count{X : p(X)} > 0. 
\end{verbatim}

\noindent Clearly, there are two candidate answer sets of $E_4$:  $S^1 =
\{p(0)\}$ and $S^2 = \{p(1),p(2),\dots\}$. Since the aggregate is defined,
$count\{X : p(X) > 0\} $ is \emph{true} in $S^1$, the body of the third rule
is \emph{false} in $S^1$ and the aggregate reduct of $E_4$ with 
respect to $S^1$ is 
\begin{verbatim}
p(1) :- not p(0). 
p(2) :- p(1). 
... 
\end{verbatim}
Hence, $S^1$ is not an answer set of $E_4$.

\st To compute the aggregate reduct of $E_4$ with 
respect to $S^2$ it is sufficient to notice that in $S^2$ 
$count\{X : p(X) > 0\} $ is undefined and hence the body of the last rule
is removed. The reduct is 
\begin{verbatim}
p(1) :- not p(0). 
p(2) :- p(1). 
... 
p(0).
\end{verbatim}
and hence $S^2$ is not an answer set of $E_4$ either.
This is not surprising since the program violates the vicious circle principle.

\end{example}

\begin{example}[Aggregates on Infinite Sets and Default Negation] \label{new5}
Finally, consider a program $E_5$ consisting of the rules:
\begin{verbatim}
even(0). 
even(I+2) :- even(I). 
q :- not min{X : even(X)} > 0. 
\end{verbatim}
The program has one answer set,
$S_{E_5} = \{q,even(0),even(2),\dots\}$. Indeed, since\\ 
$min\{X : even(X)\} > 0$ is false in $S_{E_5}$, according to clause three from the
definition of the aggregate reduct 
the body of the last rule will be first replaced by $\neg min\{X :
even(X)\} >  0$ and next, by clause four, by $min\{X : even(X)\} \leq 0$. Hence, 
the reduct of $E_5$ with respect to 
$S_{E_5}$ is 
\begin{verbatim}
even(0). 
even(2) :- even(0). 
even(4) :- even(2). 
... 
q :- even(0),even(2),even(4)... 
\end{verbatim}
and therefore $S_{E_5}$ is an answer set of $E_5$. Clearly, the program 
has no other answer sets. 
\end{example}

\section{Properties of $\mathcal{A}log$ Programs} \label{prop}
In this section we give some basic properties of $\mathcal{A}log$ programs. 
Propositions \ref{p1} and \ref{p2} ensure that, as in regular ASP,
answer sets of $\mathcal{A}log$ program are formed using the program 
rules together with the rationality principle. 
Proposition \ref{split} is the $\mathcal{A}log$ version of the
Splitting Set Theorem -- basic 
technical tool used in theoretical investigations of ASP and its
extensions. 
Proposition \ref{compl} shows that complexity of entailment in 
$\mathcal{A}log$ is the same as that in regular ASP. 
The results of this section for $\mathcal{A}log$ programs with
finite rules were presented in \cite{GelfondZ14}. The presence of infinite
rules require some modifications of the corresponding arguments. The
proofs can be found in the appendix.

\begin{claim}[Rule Satisfaction and Supportedness]\label{p1} Let 
  $A$ be an answer set of a ground $\mathcal{A}log$ program $\Pi$. 
Then 
\ee{
  \item $A$ satisfies every rule $r$ of $\Pi$. 
  \item If $p \in A$ then there is a rule $r$ from $\Pi$  such 
    that the body of $r$ is satisfied by $A$ and $p$ is the only atom 
    in the head of $r$ which is true in $A$. (It is often said that 
    rule $r$ supports atom $p$.) 
}
\end{claim}
\begin{claim}[Anti-chain Property]\label{p2}
Let $A_1$ be an answer set of an $\mathcal{A}log$ program $\Pi$. 
Then there is no answer set $A_2$ of $\Pi$ such that $A_1$ is a proper 
subset of $A_2$. 
\end{claim}
Before formulating the next theorem we need some terminology.
We say that a ground literal $l$ occurs in a ground rule $r$ \emph{explicitly} if
$l$ or $\mbox{not } l$ belongs to the head
or to the body of $r$; $l$ occurs in $r$ \emph{implicitly} if there is
a set name $\{\bar{X} : cond\}$ occurring in an aggregate literal from
$r$ and $l$ is a ground instance of some literal from $cond$; $l$
\emph{occurs} in $r$ if it occurs in $r$ explicitly or
implicitly.

\begin{claim}[Splitting Set Theorem]\label{split}
Let $\Pi_1$ and $\Pi_2$ be ground programs of $\mathcal{A}log$ such that no  
ground literal occurring in $\Pi_1$
is a head literal of $\Pi_2$  and $S$ be a set of ground
  literals containing those occurring in the rules of $\Pi_1$ and not
  containing those occurring in the heads of rules of $\Pi_2$.
A set $A$ of literals is an answer set of $\Pi_1 \cup \Pi_2$
iff $A \cap S$  is an answer set of $\Pi_1$ and $A$ is an answer set 
of $(A \cap S) \cup \Pi_2$. 
\end{claim}

\noindent As for the complexity of finding an answer set of an $\mathcal{A}log$ program, we restrict to programs with only finite number of rules and finite number of literals in any rule, which are called {\em finite} $\mathcal{A}log$ programs. 


\begin{claim}[Complexity]\label{compl}
The problem of checking if a ground atom $a$ belongs to all answer 
sets of a finite $\mathcal{A}log$ program is $\Pi_2^P$ complete. 
\end{claim}
\section{An Algorithm for Computing Answer Sets}\label{solver}
In this section we briefly outline an algorithm, called $\mathcal{A}solver$,
for computing answer 
sets of $\mathcal{A}log$ programs. 
We follow the tradition and limit our attention to programs 
without classical negation. Hence, in this section we consider
only programs of this type. 

\st The change in perspective requires introduction of some additional
terminology. By \emph{e-atom} we mean a 
regular atom or an aggregate atom possibly preceded by $not$; 
e-atoms $p$ and $not\ p$ are called \emph{contrary};
a set of e-atoms is called \emph{consistent} if it contains no
contrary e-atoms;
a \emph{partial interpretation} $I$ over signature $\Sigma$ is a 
consistent set of regular e-atoms of this signature.
By $atoms(I)$ we denote the collection of atoms from $I$.
A regular e-atom $l$ is \emph{true} 
in $I$ if $l \in I$; it is  \emph{false} in $I$ if $I$ contains an
e-atom contrary to $l$;
otherwise 
$l$ is \emph{undecided} in $I$; 
An aggregate e-atom is \emph{true} in $I$ if 
it is true in $atoms(I)$; otherwise, the atom is \emph{false} in $I$.

\st The head of a rule is \emph{satisfied} by $I$ if at least one of its atoms is 
true in $I$; the body of a rule is  \emph{satisfied} by $I$ if all of 
its e-atoms are true in $I$. 
A \emph{rule is satisfied} by $I$ if its head is satisfied by $I$ or 
its body is not satisfied by $I$. 

\st
The following notions will play substantial role in the algorithm:
\begin{definition}[Strong Satisfiability 
  and Refutability]\label{strong-satisfiablity} 
 
\begin{itemize}
\item An e-atom is \emph{strongly satisfied} by a partial 
interpretation $I$ if it is true in every partial interpretation 
containing $I$; 
\item An e-atom is \emph{strongly refuted} by
$I$ if it is false or undefined in every partial interpretation 
containing $I$; 

\item an e-atom which is neither strongly satisfied nor 
strongly refuted 
by $I$ is \emph{undecided} by $I$. 

\item A set $S$ of e-atoms is \emph{strongly satisfied} 
by $I$ if all e-atoms in $S$ are strongly satisfied by $I$; 

\item $S$ is  \emph{strongly 
  refuted} by $I$ if for every 
  partial interpretation $I^\prime$ containing $I$, some e-atom of $S$ is false 
in $I^\prime$. 
\end{itemize}
\end{definition}

\noindent 
For instance, an e-atom $p(1)$ is strongly satisfied (refuted) by any
interpretation $I$ in which $p(1)$ is true (false); 
$p(1)$ and $count\{X:p(X)\} > 2$ are undecided in $I = \emptyset$;
an atom  $card\{X:p(X)\} > n$ which is true in a partial interpretation $I$ is 
strongly satisfied by $I$; an atom $card\{X:p(X)\} < n$ which is false in $I$ is 
strongly refuted by $I$; an atom $card\{X:even(X)\} < n$ is strongly
refuted by any partial interpretation $I$; an e-atom $not\ (card\{X:even(X)\} < n)$ 
is strongly satisfied by any $I$;
and a set $\{f\{X: p(X)\} > 5, f\{X:
p(X)\} < 3\}$ is strongly refuted by any $I$.

\st 
Now we are ready to define our algorithm, $\mathcal{A}solver$.
It consists of three functions: $Solver$, $Cons$, and 
$IsAnswerSet$. 
The main function, $Solver$, is similar to that used in standard ASP 
algorithms (See, for instance, $Solver1$
from \cite{GelK13}). But unlike these functions which normally have two 
parameters - partial interpretation $I$ and program $\Pi$ - $Solver$
has two additional parameters, $TA$ and $FA$ containing aggregate e-atoms
that must be true and false respectively in the answer set 
under construction.

\st Function $Solver$ returns a pair $\langle A, true\rangle$
where $A$ is an answer
set of $\Pi$ such that for every regular atom $p$ and every aggregate
e-literal $a$
\begin{itemize}
\item if $p \in I$ then $p \in A$;
\item if $not~ p \in I$ the $p \not\in A$;
\item if $a \in TA$ then $a$ is true in $A$;
\item if $a \in FA$ then $a$ is false in $A$;
\end{itemize}

\noindent
If no such $A$ exists the function returns $false$.
The $Solver$'s description is given in the appendix. 

\st The second function, $Cons$, computes the consequences of 
its parameters - a program $\Pi$, a partial interpretation $I$, and two 
above described sets $TA$ and $FA$ of aggregate e-atoms. 
Due to the presence of aggregates the function is sufficiently 
different from a typical $Cons$ function of ASP solvers so we 
describe it in some detail. The new values of $I$, $TA$ and $FA$ containing 
the desired consequences is computed 
by application of the following {\bf inference rules}: 

\begin{enumerate}
  \item If the body of a rule $r$ is strongly satisfied by $I$ and all 
    atoms in the head of $r$ except $p$ are false in $I$ then $p$ must be in $I$.   
  \item If an atom $p \in I$ belongs to the head of exactly one rule $r$
    of $\Pi$ then every other atom from the head of $r$ must have its contrary
    in $I$, the regular e-atoms from the body of $r$ must be in 
    $I$, and its aggregate e-atoms must be in $TA$.

  \item If every atom of the head of a rule $r$ is false in $I$, and $l$ is 
the only premise of $r$ which is either an undecided regular e-atom or 
an aggregate e-atom not in $FA$, and  the rest of the body is strongly 
satisfied by $I$, then 
  \ee{
    \item if $l$ is a regular e-atom, then the contrary of $l$ must be in $I$,
	\item if $l$ is an aggregate e-atom, then it must be in $FA$. 
  }
  \item \label{firstSupportRule}   
  If the body of every rule with $p$ in the head is strongly refuted by $I$, then $(not\ p)$ must be in $I$. 
\end{enumerate}


\st Given an interpretation $I$, a program $\Pi$, inference
rule $i \in [1..4]$ and $r \in \Pi$, let function $iCons(i,I,\Pi,r)$ 
return $<\delta I, \delta TA, \delta FA>$ where 
$\delta I$, $\delta TA$ and $\delta FA$ are the results of applying
inference rule $i$ to $r$. (Note, that inference rule $4$ does not 
really use $r$ -- it is added for the uniformity of notation). 

\st We also need the following terminology.
We say that $I$ is \emph{compatible} with $TA$ if  $TA$
is not strongly refuted by $I$; $I$ is \emph{compatible} with $FA$
if no e-atom from $FA$ is strongly satisfied by $I$. 
A set $A$ of regular atoms is 
\emph{compatible} with $TA$ and $FA$ if the set 
$compl(A) = \{p : p \in A\} \cup 
\{not~ a: a \notin A\}$ is \emph{compatible}
with $TA$ and $FA$; $A$ is compatible with $I$ if $I \subseteq 
compl(A)$. 
The algorithm $Cons$ is listed below.


\begin{tabbing}
abcde\=ab\=ab\=ab\aa\=aa\=\kill 
{\bf function} Cons \\
\> {\bf input}: \>\>\>partial interpretation $I_0$, sets $TA_0$ and $FA_0$
of aggregate e-atoms compatible with $I_0$, \\
\> \> \> \> and program $\Pi_0$ with signature $\Sigma_0$; \\
\> {\bf output}: \\
\> \> $\langle I, TA, FA, true\rangle$ where \\
\> \> \> \> $I$ is a partial interpretation such that $I_0 \subseteq I$, \\ 
\> \> \> \> $TA$ and $FA$ are sets of aggregate e-atoms such that $TA_0 \subseteq TA$ and $FA_0 \subseteq FA$, \\
\> \> \> \> $I$ is compatible with $TA$ and $FA$, and\\
\> \> \> \> for every set $A$ of regular ground atoms satisfying $\Pi_0$,\\
 \> \> \> \>  $A$ is compatible with $I_0$, $TA_0$, and $FA_0$  iff\\
\> \> \> \>  $A$ is compatible with $I$, $TA$ and $FA$.\\
\> \> $\langle I_0, TA_0, FA_0, false\rangle$ if no such $I, TA, FA$ exist.\\
\> {\bf var} $I, T$: set of e-atoms; $TA, FA$: set of aggregate e-atoms; $\Pi$: program; \\
1. \> Initialize $I$, $\Pi$, $TA$ and $FA$ to be $I_0$, $\Pi_0$, $TA_0$ and $FA_0$ respectively; \\
2. \> {\bf repeat} \\
3. \> \> $T$ := $I$; \\
4. \> \> Non-deterministically select an inference rule $i$ from (1)--(4); \\
5. \> \> {\bf for} every $r \in \Pi_0$ {\bf do}\\ 
6. \> \> \>\> $<\delta I, \delta TA, \delta FA>$ := $iCons(I,\Pi_0,i,r)$; \\
7. \> \> \>\> $I$ := $I \cup \delta I$, $TA$ := $TA \cup \delta TA$, $FA$ := $FA \cup \delta FA$; \\
8. \> {\bf until} $I$ = $T$; \\
9. \> {\bf if} $I$ is consistent and $TA$ and $FA$ are compatible with 
$I$  {\bf then} \\
10. \> \> {\bf return} $<I, TA, FA, true>$; \\
11. \> {\bf else} \\
12.  \> \> {\bf return} $<I_0, TA_0, FA_0, false>$; \\
\end{tabbing}

\medskip The third function, $IsAnswerSet$, of 
$\mathcal{A}solver$ checks if the collection of atoms
  of an interpretation $I$ is an answer set 
of program $\Pi$. It computes the aggregate reduct of $\Pi$ with 
respect to $atoms(I)$ and applies usual checking algorithm (see, for 
instance, \cite{KochLP03}).

\begin{claim}[Correctness of the Solver]
If, given a program $\Pi_0$, a partial interpretation $I_0$, and sets 
$TA_0$ and $FA_0$ of aggregate e-atoms Solver($I_0, TA_0, FA_0,
\Pi_0$) returns $\langle A, true\rangle$ then $A$
is 
an answer set of $\Pi_0$ compatible 
with $I_0$, $TA_0$ and $FA_0$. If there is no such answer set,
the solver returns $false$.
\end{claim}

\noindent To illustrate the algorithm consider a program $\Pi$
\begin{verbatim}
:- p(a).
p(a) :- card{X:q(X)} > 0.
q(a) or p(b).
\end{verbatim}
and trace $Solver(\Pi,I,TA,FA)$ where $I$, $TA$, and $FA$ are empty.
$Solver$ starts by calling $Cons$ which computes the consequences:
$not\ p(a)$ (from the first rule of the program), $\{card\{X:q(X)\} > 0\}=false$
(from the second rule of the program) and $not\ q(b)$ (from the fourth
inference rule), and returns $true$, $I=\{not\ q(b), not\ p(a)\}$ and new
$FA=\{card\{X:q(X)\} > 0\}$; $TA$ is unchanged. 
$Solver$ then guesses $q(a)$ to be true, i.e., 
$I=\{not\ q(b), not\ p(a), q(a)\}$, and calls $Cons$ again.  $Cons$ 
does not produce any new consequences but finds that $FA$ is 
not compatible with $I$ (line 9 of the algorithm). So, it returns
$false$, which causes $Solver$ to set $q(a)$ to be false, i.e., 
$I=\{not\ q(b), not\ p(a)$, $not\ q(a)\}$. $Solver$ then calls $Cons$ 
again which returns $I=\{not\ q(b), not\ p(a), not\ q(a), p(b)\}$. 
$Solver$ finds that $I$ is complete and calls $IsAnswerSet$ which returns true. 
Finally, $Solver$ returns $I$ as an answer set of the program.

\section{Comparison with Other Approaches}\label{comp}
There are a large number of proposed extensions of ASP by aggregates
based on different intuitions and different mathematical techniques
used in the definition of their semantics,  e.g., \cite{KempS91,gel02,nss02,Marek04,Marek2004set,Pelov04,PelovDB04,PelovT04,Ferraris05,FerrarisL05,pdb07,LeeLP08,ShenYY09,LiuPST10,FaberPL11,PontelliST11,WangLZY12,HarrisonLY14}.
To the best of our knowledge none of them incorporates the vicious
circle principle and hence their semantics are different from that of 
$\mathcal{A}log$. Perhaps more surprisingly, even for programs 
with finite rules, the syntax of 
$\mathcal{A}log$ also differs substantially from syntax of most other logic 
programming languages allowing aggregates 
(with the exception of that in \cite{gel02}). 

To illustrate the difference between $\mathcal{A}log$ and other
aggregate extensions of ASP we concentrate on two
languages, \cite{FaberPL11} and \cite{SonP07}, which can be viewed as
good representatives of this line of work.
We refer to these languages as
$\mathcal{F}log$ and $\mathcal{S}log$  respectively. 
Due to multiple equivalence
results discussed in these two papers our comparison can be extended  
to cover most of the existing approaches. 

\subsection{Comparison with $\mathcal{F}log$}

\subsubsection{Syntax and Semantics of $\mathcal{F}log$}
We start with a short overview of a slightly simplified version of $\mathcal{F}log$. 
The simplifications are chosen to facilitate the presentation and are
made explicit whenever possible. We believe that the simplified version preserves the essence of the 
language.

\medskip
Similar to $\mathcal{A}log$, aggregates of $\mathcal{F}log$ can be
viewed as functions from
\emph{sets} of constants to constants. 
$\mathcal{F}log$ allows two different ways to name sets:
\emph{symbolic names} which correspond to set names of 
$\mathcal{A}log$, and \emph{ground sets} -- expressions of the form 
$\{ c_1 : Conj_1,\dots,c_k : Conj_k $ where $c$'s are constants 
and $Conj$'s are collections of ground regular atoms. Intuitively,
$\{X:p(X,Y\}$ denotes a set $S=\{c_1,\dots,c_n\}$ of constants 
such that for every constant $c$, $c \in S$ iff there is a constant $y$ such that 
$p(c,y)$ is true. (Note that this is different from $\mathcal{A}log$
in which variable $Y$ is free, i.e. not bound by an existential
quantifier.)
A ground set denotes the set of its constants that satisfy their
conditions\footnote{$\mathcal{F}log$ allows ground sets whose elements
  are of the form $(\bar{v} : Conj)$ where $v$ is a vector. Such a set
  represents the multisets consisting of the first coordinates of vectors
  satisfying their corresponding conditions.}.

Both, symbolic names and ground sets are 
are referred to as \emph{set terms}. An \emph{aggregate function} of $\mathcal{F}log$
is an expression of the form $f(A)$ where $A$ is a set term and $f$ is 
the name of an aggregate. An aggregate function can be viewed as a 
(possibly partial) function mapping sets of constants to a 
constant. Aggregate atoms of $\mathcal{F}log$ are of the form 
$\langle aggregate\_term \rangle \langle arithmetic\_relation \rangle 
\langle value \rangle$ where $value$ is a number or a variable. 
An $\mathcal{F}log$ {\em rule} is of the form 
$$head \leftarrow pos, neg$$
where $head$ is the disjunction of a \emph{finite} set of regular \emph{atoms},
$pos$ and $neg$ are \emph{finite} collections of \emph{atoms} and \emph{atoms} preceded 
by $not$ respectively. The {\em body} of the rule is $pos \cup neg$. 
An {\em $\mathcal{F}log$ program} is a set of $\mathcal{F}log$ rules. 

\st 
As usual, semantics of $\mathcal{F}log$ is defined for ground 
programs. The $\mathcal{F}log$ notion of grounding is, however,  
very different from that of $\mathcal{A}log$. It is based on an 
important distinction between so called local and global variables.

\st A {\em local variable} of a rule $r$ is a variable of $r$ that 
appears solely in some aggregate term occurring in $r$. 
A variable of a rule $r$ is {\em 
  global} if it is not a local variable of $r$. 
A substitution from the set of local variables of an occurrence $S$ of 
 symbolic name in a rule $r$ is called \emph{local for $S$ in 
$r$}. 

\noindent
Given an occurrence of a symbolic name $S = \{X : Conj\}$ in a rule $r$
containing no variables global in $r$, the \emph{ground instantiation} of 
$S$ is the set $inst(S,r)$ of \emph{instances of $S$} -- ground terms
of the form
$$\gamma(Vars) : \gamma(Conj)$$ 
where $ \gamma$ is a local substitution for $S$ in $r$. 


\st A {\em ground instance} of a rule $r$ is obtained by two steps: 
\begin{enumerate}
\item replace 
the occurrences of global variables of $r$ by constants, and 
\item replace each occurrence of a symbolic name $S=\{x: p(x)\}$ in $r$
by its ground instantiation.
\end{enumerate}
\emph{Ground instantiation of a rule} is the collection of its ground instances. 
Note that for ease of presentation, we assumed that there is only 
one bound variable in a set name. (This is a substantial technical simplification
since it allows us to avoid difference between $\mathcal{A}log$ and $\mathcal{F}log$
treatment of aggregates defined on sets of vectors. The former allows
aggregates to be arbitrary functions on such sets. The later only
allows functions defined on first coordinates of their vectors.)

\noindent
The following example illustrates the notion of $\mathcal{F}log$ grounding



\begin{example}[$\mathcal{F}log$ grounding]\label{new6}
Consider a program $P_2$ from Example \ref{e3}. 
\begin{verbatim}
q(Y) :- card{X:p(X,Y)} = 1, r(Y). 
r(a).  r(b).  p(a,b). 
\end{verbatim}
(We will refer to the first rule of $P_2$ as rule (1).)
$P_2$ can be viewed as a program of $\mathcal{F}log$.
Variable $X$ of (1) occurs solely in an aggregate term 
$card\{X:p(X,Y)\}$ of this rule and hence is local. Variable $Y$ 
occurs in $r(Y)$ and hence is global. To produce 
ground instances of (1) we first replace global variable $Y$ by
constants of the language:
\begin{verbatim}
q(a) :- card{X:p(X,a)} = 1, r(a). 
q(b) :- card{X:p(X,b)} = 1, r(b). 
\end{verbatim}
Symbolic name $\{X:p(X,a)\}$ has no global variables.
Its ground instantiation is defined and is equal to $\{a:p(a,a), b:p(b,a)\}$.
Similarly for  $\{X:p(X,b)\}$ the second rule.
Thus, the ground instantiation of rule (1) is the following collection of
$\mathcal{F}log$ rules:
\begin{verbatim}
q(a) :- card{a:p(a,a),b:p(b,a)} = 1, r(a). 
q(b) :- card{a:p(a,b),b:p(b,b} = 1, r(b). 
\end{verbatim}
\end{example}


\medskip
As usual, an $\mathcal{F}log$ program with variables is viewed as a
shorthand for its ground instantiation -- the union of ground
instantiations of its rules.
Hence the semantics of $\mathcal{F}log$ will be given for ground programs.
We start with defining satisfiability of various program constructs 
with respect to a set $A$ of ground regular atoms. Satisfiability 
relation between $A$ and a construct $C$ will be denoted by 
$A \models C$. Often we will also read $A \models C$ as ``$C$ is \emph{true}
in $A$''.

\begin{itemize}
\item The definition of satisfiability of 
\emph{regular e-atoms} in $\mathcal{F}log$ is the same as in $\mathcal{A}log$. 

\item Let $T= f(\{c_1 : conj_1,\dots, c_k : conj_k)$ be a ground
  aggregate term, $A$ be a set of ground regular atoms, and 
$S = \{c_i : 1 \leq i \leq k \mbox{ and } A \models conj_i\}$:
\begin{itemize}
\item 
$f(T) \odot n$ is satisfied by $A$ if $f(S) \odot n$\footnote{Recall, that a partially defined aggregate $f$
  satisfies condition $f(S) > n$ if $f(S)$ is defined and its value is 
  greater than $n$. Similarly for other relations.}.

\item $not~ f(T) \odot n$ is satisfied by $A$ if 
  $f(S)$ is defined and its value $y$ does not satisfy 
  the relation $y \odot n$.

\end{itemize}





\item A rule $r$ is {\em satisfied} by $A$ 
if either the head of $r$ is satisfied by $A$ or 
some element of the body of $r$ is not satisfied by $A$. 
\end{itemize}

\st 
Given an $\mathcal{F}log$ ground program $P$ and a set $A$ of ground 
regular atoms we say that 

\begin{itemize}
\item $A$ is a {\em model} of $P$ if all rules of $P$ are satisfied by $A$.

\item The {\em $\mathcal{F}log$ reduct} of $P$ with respect to $A$, 
denoted by $R_{\mathcal{F}}(P, A)$, is the program obtained from $P$ 
by removing every rule whose body contain an element not satisfied by $A$


\item $A$ is an {\em $\mathcal{F}log$ answer set} of $P$ if $A$ is a subset minimal model of $R_{\mathcal{F}}(P, A)$. 

\end{itemize}

\begin{example}[Answer Set of an $\mathcal{F}log$ Program]\label{new7}
Consider the grounding of $\mathcal{F}log$ program $P_2$ from Example \ref{new6}:
\begin{verbatim}
q(a) :- card{a:p(a,a),b:p(b,a)} = 1, r(a). 
q(b) :- card{a:p(a,b),b:p(b,b} = 1, r(b). 
r(a).  r(b).  p(a,b). 
\end{verbatim}
and a set $A=\{r(a), r(b),  p(a,b), q(b)\}$.
The $\mathcal{F}log$ reduct, $R_{\mathcal{F}}(P_2, A)$ is
\begin{verbatim}
q(b) :- card{a:p(a,b),b:p(b,b} = 1, r(b). 
r(a).  r(b).  p(a,b). 
\end{verbatim}
Clearly, $A$ is the subset minimal model of this reduct and hence $A$
is an answer set of $P_2$. It is easy to see that $P_2$ has no other
answer sets.

\end{example}

\subsubsection{Comparison between $\mathcal{A}log$ and $\mathcal{F}log$}
In this section we limit our attention to programs with variables
whose rules satisfy syntactic requirements of both, $\mathcal{A}log$
and $\mathcal{F}log$, i.e. to programs with finite rules not containing
classical negation, non-aggregate function symbols, and ground set
of $\mathcal{F}log$. 

Even though the rules look the same there are very substantial
difference between the two languages. They are related primarily
to differences in
\begin{itemize} 
\item treatment of variables
occurring in aggregate terms,
\item understanding of grounding,
\item understanding of the meaning of default negation of aggregate
  atoms,
\item treatment of recursion through aggregates.
\end{itemize} 
We will illustrate these differences by a number of examples.

\begin{example}[Variables in Aggregate Terms: Global versus Bound]\label{new7}
Consider an $\mathcal{A}log$ program $P_3$
\begin{verbatim}
r :- card{X:p(X)} >= 2, q(X). 
p(a).  p(b).  q(a). 
\end{verbatim}
from Example \ref{e4}. 
Here the occurrence of $X$ in $q(X)$ is \emph{free} (while the
occurrences of $X$ within a set name $card\{X:p(X)\}$ are bound. Hence
the $\mathcal{A}log$ grounding of $P_3$ is:
 \begin{verbatim}
r :- card{X:p(X)} >= 2, q(a). 
r :- card{X:p(X)} >= 2, q(b). 
p(a).  p(b).  q(a). 
\end{verbatim}
and, therefore, $P_3$ has one answer set, $A = \{p(a),  p(b),  q(a),r\}$.

\st If, however, we view the same program as a program of 
$\mathcal{F}log$ the situation changes. Instead of using 
usual logical concept of bound and free occurrence of a variable
$\mathcal{F}log$ uses very different concepts of global and local
variable of a rule. Since there is an occurrence of $X$ 
outside of the set name (in $q(X)$) the variable is bound, and hence
the $\mathcal{F}log$ grounding of $P_3$ is:
 \begin{verbatim}
r :- card{a:p(a)} >= 2, q(a). 
r :- card{b:p(b)} >= 2, q(b). 
p(a).  p(b).  q(a). 
\end{verbatim}
The bodies of the rules are not satisfied by $A = \{p(a),  p(b),
q(a)\}$ and hence it is the only answer set of $P_3$ in
$\mathcal{F}log$.

\end{example}

\st We believe that the $\mathcal{A}log$ notion of grounding has a
substantial advantage over that of $\mathcal{F}log$ (and many
other languages and systems which adopted the same syntax).
It avoids the need for introduction of fairly complex notions of a local variable
and a ground set with its unusual membership of pairs. But more
importantly it allows to define the intuitive (and formal) meaning of a set name of
$\mathcal{A}log$ independently from its occurrence in a rule. 
In $\mathcal{F}log$, a name $f\{X:p(X)\}$ has different intuitive
readings in rules:
 \begin{verbatim}
r :- f{X:p(X)} > 1, q(a). 
\end{verbatim}
and
 \begin{verbatim}
r :- f{X:p(X)} > 1, q(X). 
\end{verbatim}
The program consisting of the first rule, used together with a set of facts:
\begin{verbatim}
p(a).   p(b).   q(a).
\end{verbatim}
has answer set $\{p(a),   p(b),   q(a), r\}$. The answer set of a program
obtained by combining the second rule with the same facts is
$\{p(a),   p(b),   q(a)\}$.

\st As the result of its definition of grounding, in $\mathcal{A}log$, every 
set name $\{X : p(X)\}$ can be replaced by a name $\{Y : p(Y)\}$
without change in the meaning of the program.
As the above example shows this is clearly not the case in
$\mathcal{F}log$, which makes
declarative reading of aggregate terms substantially more
difficult.
In general we believe that \emph{independence of an intuitive meaning of a language
construct from the context in which it appears should be a desirable
feature of a knowledge representation language}.

\st
The other 
difference in reading of symbolic names is related to the treatment of 
variable $Y$ in $\{X: p(X,Y)\}$ even if both, $X$ and $Y$ are local in
the corresponding rule of $\mathcal{F}log$.
In $\mathcal{F}log$ the variable is bound by an unseen 
existential quantifier. If all the variables are local then 
$S=\{X:p(X,Y)\}$ is really $S_1 = \{X: \exists Y \ p(X,Y)\}$. In 
$\mathcal{A}log$, $Y$ is free. Both approaches are reasonable but we 
prefer to deal with the different possible readings by introducing an explicit 
existential quantifier as in Prolog \textcolor{red}{\cite{lloydr-1}}. It is easy semantically and, to
save space, we do not discuss it in the paper.


\st
 $\mathcal{A}log$ also differs from $\mathcal{F}log$ in 
 understanding of the meaning of default negation in front of 
aggregate atoms. To see the difference consider an aggregate $f$
which is \emph{undefined} on a set $S$ (e.g. $sum\{a,b\}$).
According to the semantics of  $\mathcal{A}log$ and atom
$not~ sum\{a,b\}=1$ is \emph{true} in any interpretation $I$.
This view is in sink with the generic epistemic reading of a statement $no~ p$:  
the statement is true if ``there is no reason to believe that $p$ is
true''. Clearly, since $sum\{a,b\}$ is undefined one certainly has no
reason to believe that it is equal to $1$.
In fact, one knows that it is not.

\st According to the semantics of $\mathcal{F}log$, however, statement
$not~ sum\{a,b\}=1$ is not true. For $no~f(S) = 1$ to be true according
to the $\mathcal{F}log$ semantics $sum\{a,b\}$ must be defined and
its value must be different from $1$.

\st We are not sure why this choice was made by the designers of $\mathcal{F}log$.
As far as far as we can see in this situation default negation of
$\mathcal{F}log$ behaves more like the classical negation, $f(S) \not=
1$, which is true exactly when $f(S)$ is defined and different from $1$.

\st Finally we mention
semantic difference which are due to the difference between informal
principles underlying both semantics. 

\begin{example}[Vicious Circles in $\mathcal{F}log$] \label{e11a} 
Consider the following program, $P_6$, adopted from  \cite{SonP07}:
\begin{verbatim}
p(1) :- p(0). 
p(0) :- p(1). 
p(1) :- count{X: p(X)} != 1. 
\end{verbatim}
We can view $P_6$ as a ground $\mathcal{A}log$ program. It is easy to
see that it has no answer sets. Clearly, $\emptyset$, $\{p(0)\}$, and
$\{p(1)\}$ are not answer sets since they do not satisfy the rules of
the program. Now consider $A=\{p(0),p(1)\}$.
 Since the aggregate reduct of $P_6$ with respect to
$A$ is
\begin{verbatim}
p(1) :- p(0). 
p(0) :- p(1). 
p(1) :- p(0), p(1).
\end{verbatim}
and $A$ is not an answer set of the reduct it is also not an answer set
of $P_6$.
The result is to be expected 
since the program's definition of $p(1)$
is given in terms of fully defined set $\{X:p(X)\}$, i.e., the 
definition contains a vicious circle. 

Now let us view $P_6$ as a program of $\mathcal{F}log$.
The program is not ground.
Its $\mathcal{F}log$ grounding is
\begin{verbatim}
p(1) :- p(0). 
p(0) :- p(1). 
p(1) :- count{0: p(0, 1 : p(1))} != 1. 
\end{verbatim}
which is unchanged by the reduct operation. $A$ is a minimal set
satisfying these rules and therefore it is an answer set of $P_6$.
Notice, that \emph{belief in $p(1)$ sanctioned by the 
semantics of $\mathcal{F}log$ is self-supported}. 
Informal argument justifying this result $\mathcal{F}log$ 
may go something like this:
Clearly, $A$ satisfies the rules of the program. Since no proper subset of $A$
satisfies these rules, minimality principle is satisfied and $A$ is an
answer set of $P_6$. The semantics does not take into consideration
justifications for the membership of an element in the set.
This seems to be understood and intended by the designers of $\mathcal{F}log$.
In fact, Faber et al seem to express something similar in their \emph{black box 
principle}: ``\emph{when checking stability they [aggregate literals] are
either present in their entirety or missing altogether}''.
We are having some difficulties in understanding this principle and
find $\mathcal{A}log$ refusal to form rational beliefs on the basis 
of rules of $P_6$ more intuitive. In fact we believe that
\emph{avoidance of self-referential beliefs is an important principle
of rational behavior formalized by the original ASP}. After all, an ASP
program
\begin{verbatim}
p :- p.
:- not p.
\end{verbatim}
does not have answer sets even though $\{p\}$ is the minimal set 
satisfying its rules and $p$ is self-supported by the first rule of
the program. In ASP, and in $\mathcal{F}log$, absence of
self-referencing beliefs is not explicitly mentioned as the underlying
principle of the semantics since it follows from the other principles.

\st It is also worth mentioning that
in this particular example we are in agreement with $\mathcal{S}log$
which requires that the value of an aggregate atom can be computed
before the rule with this atom in the body can be used in the
construction of an answer set.
It is, of course, possible that further study will produce more convincing
arguments in favor of one or another informal principles incorporated
in the semantics of  these languages.

\end{example}





\hide{
\st Consider the grades of several homework of a list of students taking a course. The continuous performance of a student is good if none of her homework grade is less than $60$. 

\begin{verbatim}
grade(peter, hw1, 70).
...
grade(peter, hw5, 90).
...
grade(zion, hw1, 50).

goodPerformance(S) :- not min{G, Hw:, grade(S, Hw, G)} < 60.
\end{verbatim}
(??We allow list of bound vars in a set name. But no details on how to apply 
aggregate function to a set of tuples??)

Assume we have a student $john$ who missed all homework. According to the program and $\mathcal{A}log$ semantics, $john$ has good continuous performance. By $\mathcal{F}log$, we do not know the performance of $john$. 
\hfill $\Box$. }

\st Despite all the differences between the two languages there is
some close relationship between them. 
The following result shows that under certain conditions, 
an $\mathcal{A}log$ answer set of a program is an $\mathcal{F}log$
answer set of the program. 

\begin{definition}[Compatible Programs]
Consider a program $\Pi$ which, syntactically, can be viewed as a
program of both, $\mathcal{A}log$ and $\mathcal{F}log$.
The program is called $\mathcal{AF}$-\emph{compatible}
it satisfies the following conditions 
\begin{itemize}
\item no variables have 
bound occurrence in two different aggregate terms of the same rule, 
\item no local variables in any rule are free, 
\item no aggregate functions are partial, 
\item every variable having a free occurrence in an aggregate term of a rule
$r$ also appears 
in some regular literal of $r$ outside of rule's aggregate atoms.
\end{itemize}
\end{definition}

\begin{claim}[From $\mathcal{A}log$ Answer Sets to $\mathcal{F}log$ Answer Sets] \label{prop:a2flog}
If $\Pi$ is  $\mathcal{AF}$ compatible program 
and $A$ is an $\mathcal{A}log$ answer set of $\Pi$ then it is an $\mathcal{F}log$ answer set of $\Pi$.  
\end{claim}
In other words, entailment of $\mathcal{F}log$ is more cautious than
that of $\mathcal{A}log$, i.e. every $\mathcal{F}log$ consequence of $\Pi$
is also its $\mathcal{A}log$ consequence.
As the following example shows the reverse is not true even 
for consistent $\mathcal{A}log$ programs.
An $\mathcal{A}log$ program
\begin{verbatim}
p(1) :- count{0: p(0, 1 : p(1))} != 1,b.
b or c. 
\end{verbatim}
has an answer set $\{c\}$ and hence is consistent. Viewed as an
$\mathcal{F}log$ program it has one more answer set $\{b,p(1)\}$.
As expected, the difference between semantics disappears if a program
is stratified with respect to aggregates, i.e. has no recursion through aggregates:

\begin{definition}[\cite{FaberPL11}] \label{prop:leveling}
An $\mathcal{A}log$ program $P$ is {\em   aggregate stratified} if 
there is a level mapping $||~||$ from the predicates of $P$ to natural numbers, such that for each rule $r \in P$ and for each prediate $a$ occurring in the head of $r$, the following holds: 
\ie{
  \item for each predicate $b$ occurring in the body of $r$, $||b|| \le ||a||$,
  \item for each predicate $b$ occurring in an aggregate atom of $r$, $||b|| < ||a||$, and 
  \item for each predicate $b$ occurring in the head of $r$, $||b|| = ||a||$. 
}

\end{definition}

 \begin{claim}[Equivalence of $\mathcal{A}log$ and
   $\mathcal{F}log$ Semantics for Aggregate Stratified Programs] \label{prop:stratA2F}
If $\Pi$ is  an aggregates stratified $\mathcal{AF}$-compatible program then
$A$ is an $\mathcal{A}log$ answer set of $\Pi$ iff it is an $\mathcal{F}log$ answer set of $\Pi$.  
\end{claim}

\subsection{Comparison with $\mathcal{S}log$} 

\subsubsection{Syntax and Semantics of  $\mathcal{S}log$}


As for $\mathcal{F}log$, we also provide a slightly simplified version of $\mathcal{S}log$ here. 
The simplifications are chosen to facilitate the presentation and are
made explicit whenever possible. We believe that the simplified version preserves the essence of the 
language.

\st An {\em $\mathcal{S}log$ aggregate atom} is of the form $f\{X: p(X)\} \odot n$ where $n$ is a variable or number and $f$ is a function maps a collection of sets to integers. The variable $X$ is called {\em grouped variable} and other variables in $p(X)$ are called {\em free variables}. As we can see that grouped variables in $\mathcal{S}log$ corresponds to bound variables in $\mathcal{A}log$, and free variables in $\mathcal{S}log$ corresponds to those in $\mathcal{A}log$.

\st An {\em $\mathcal{S}log$ rule} is of the form $$head \leftarrow aggs, pos, neg$$ 
where $head$ contains at most one regular atom, $pos$ and $neg$ are finite collections of regular atoms and regular atoms preceded by $not$ respectively, and $aggs$ a set of aggregate atoms. The {\em body} of the rule is $agg \cup pos \cup neg$. 
An {\em $\mathcal{S}log$ program} is a finite set of $\mathcal{S}log$ rules.

\st A {\em ground instance} of a rule of a program $P$ is the result of replacing free variables by ground terms of $P$. 
We use $ground(P)$ to denote the set of ground instances of all the rules of $P$. 

\st Consider a set $S$ of ground regular atoms and an aggregate atom $agg$. $Base(agg)$ denotes the set of the ground instantiations of atoms occurring in the set name of $agg$. 
We define $ta(agg, S) = \{l: l \in S, l \mbox{ occurs in } agg\}$, i.e., $S \cap Base(agg)$, and $fa(agg,S) = Base(agg) - S$.

\st The definition of a set $S$ of ground atoms satisfying a ground aggregate atom $agg$, denoted by $S \models agg$ in $\mathcal{S}log$ follows that in $\mathcal{A}log$. 
An {\em aggregate solution} of a ground aggregate atom $agg$ is a pair $\langle S_1, S_2 \rangle$ of disjoint subsets of $Base(agg)$ such that for every set $S$ of regular ground atoms, if $S_1 \subseteq S$ and $S \cap S_2 = \emptyset$ then $S \models agg$.

\st Given a program $P$ and a set $S$ of ground regular atoms, the {\em $\mathcal{S}log$ reduct} of $P$ with respect to $S$, denoted by $^S\!P$, is defined as 

$^S\!P$ = 
  \begin{minipage}[t]{0.85\textwidth}
	\{ 
	 \begin{minipage}[t]{\textwidth}
	  $head(r) \leftarrow pos(r), aggs(r): r \in ground(P)$, $pos(r)$ is the set of regular atoms of the body of $r$, $aggs(r)$ is the set of aggregate atoms of $r$, and $S \cap neg(r) = \emptyset$ where $neg(r)$ is the set of negative regular atoms of $r$ \}.
	 \end{minipage}
  \end{minipage}

\st Let $a$ be an atom and $I$, $S$ be two sets of ground regular atoms. The {\em conditional satisfaction} of $a$ with respect to $I$ and $S$, denoted by $(I, S) \models a$, is defined as
\ie{
  \item If $a$ is a regular atom, $(I, S) \models a$ if $I \models a$,
  \item If $a$ is an aggregate atom, $(I, S) \models a$ if $\langle I \cap S \cap Base(a), Base(a) - S \rangle$ is an aggregate solution of $a$. 
}

\noindent Given a set $A$ of ground atoms (regular or aggregate), $(I, S) \models A$ denotes that for every atom $a \in A$, $(I, S) \models a$.  

\st Given an $\mathcal{S}log$ program $P$ and a set $S$ of ground regular atoms,
for any collection $I$ of ground regular atoms of $P$, the {\em consequence operator} on $P$ and $S$, denoted by $K_S^P$, is defined as

$$K_S^P(I) = \{head(r) : r \in~ ^S\!P \mbox{ and } (I, S) \models body(r)\}.$$ 

\st A set $S$ of ground regular atoms is an {\em $\mathcal{S}log$ answer set} of a $\mathcal{S}log$ program $P$ if $S = lfp(K_S^P)$.

\subsubsection{Difference}

In terms of syntax, $\mathcal{S}log$ allows multisets. It does not allow infinite number of atoms in a rule, infinite number of rules in a program, disjunction in the head of rules, classical negations, or partial aggregate functions.  

\st As for semantics, the absence of answer set of $P_6$ in  $\mathcal{S}log$ may suggest that
it adheres to our formalization of the VCP. 
The next example shows that it is not the case.

\begin{example}[VCP and Constructive Semantics of aggregates]\label{e12}
Let us consider a program $P_7$.
\begin{verbatim}
p(a) :- count{X:p(X)} > 0.
p(b) :- not q.
q :- not p(b).
\end{verbatim}
As shown in \cite{SonP07} the program has two $\mathcal{S}log$ answer sets, $A=\{q\}$ and
$B = \{p(a),p(b)\}$. An informal argument used to construct $B$ may look as
  follows: ``Clearly, belief in $p(b)$ in justified by the second rule
  of the program. Hence, atom $count\{X:p(X)\} > 0$ is satisfied 
independently of the final extent of $p$, which, by the first
rule, justifies belief in $p(a)$.''

\st If, however, $P_7$ is viewed as a program of  $\mathcal{A}log$,
it will have only one answer set, $A$. This happens because the
$\mathcal{S}log$ construction
of $B$  uses knowledge about properties
of the  \emph{aggregate atom} of the first rule (in this case the
monotonicity of $>$). In contrast, the semantics of
$\mathcal{A}log$ only takes into account the meaning of the  \emph{parameter
of the aggregate
term}. Both approaches can, probably, be successfully defended but, in
our opinion, the $\mathcal{S}log$ semantics has a disadvantage of being less general 
(it is only applicable to non-disjunctive programs), and more complex mathematically.
\end{example}
 
\subsubsection{Similarity}


The $\mathcal{A}log$ answer sets and the $\mathcal{S}log$ answer sets of programs without involving multisets are related in the following way.

\begin{claim}[From $\mathcal{A}log$ Answer Sets to $\mathcal{S}log$ Answer Sets] \label{prop:a2slog}
Consider an $\mathcal{S}log$ program $P$ without multisets in its rules. If $A$ is an $\mathcal{A}log$ answer set of $P$, it is an $\mathcal{S}log$ answer set of $P$.  
\end{claim}


\st For the stratified programs, $\mathcal{S}log$ semantics coincides with $\mathcal{A}log$ semantics .

\begin{claim}[Equivalence of $\mathcal{A}log$ and
   $\mathcal{S}log$ Semantics for Aggregate Stratified Programs]\label{prop:stratA2S}
Given an aggregate stratified $\mathcal{S}log$ program $P$ without
multisets, $A$ is an $\mathcal{S}log$ answer set of $P$ iff $A$
is an $\mathcal{A}log$ answer set of $P$. 
\end{claim}

\noindent Propositions \ref{prop:stratA2F}  and \ref{prop:stratA2S}
have a useful consequence. They allow to prove that for stratified
programs without disjunction all three semantics coincide.

\begin{corollary}[Equivalence of $\mathcal{F}log$, $\mathcal{S}log$,
  $\mathcal{A}log$ for Stratified Non-disjunctive Programs]
Given an aggregate stratified $\mathcal{S}log$ program $P$, a set $A$ of ground regular atoms is an $\mathcal{S}log$ answer set of $P$ iff $A$ is an $\mathcal{F}log$ answer set of $P$ iff $A$ is 
an $\mathcal{A}log$ answer set of $P$.
\end{corollary}

\subsection{Algorithm}

A key difference between our algorithm and those in the existing work 
\cite{FaberPLDI08,GebserKKS09} is that the other work needs rather 
involved methods to ground the aggregates while our algorithm does not 
need to ground the aggregate atoms. As a result, the ground program
used by our algorithm may be smaller, and our algorithm is simpler.  

\section{More on the Set Constructs of $\mathcal{A}log$} \label{sec:set}
In this section we introduce several advanced constructs of 
$\mathcal{A}log$ which are somewhat analogues to conditional literals
and choice rules of Lparse and Clingo. The new constructs are aimed at
modeling the same natural language phenomena as the latter but differ 
substantially from them in syntax and the intuitive meaning.

\st 
{\bf Subsets statements in the bodies of $\mathcal{A}log$  rules.}

\st
We start by expanding the syntax of $\mathcal{A}log$ by introducing
binary relations $=$, $\leq$ and $<$ on sets of $\mathcal{A}log$.
The two latter relations are read as \emph{subset} and \emph{proper 
  subset} respectively. An expression of the form
\begin{equation}\label{set-atom}
A_1 \odot A_2,
\end{equation}
where $A_1$ and $A_2$ are set names and $\odot$ is one of the
relations $=$, $\leq$ and $<$, is referred to as a \emph{set atom}.
A ground set literal $\{\bar{X}_1:cond_1(\bar{X}_1)\} < \{\bar{X}_2:cond_2(\bar{X}_2)\} $ 
is \emph{true} in a set of ground regular literals $S$ if 
$\{\bar{t}_1:S \models cond_1(t_1)\}$ is a proper subset of
$\{\bar{t}_2: S \models cond_2(t_2)\} $. Similarly for other relations.
To simplify the writing of the corresponding
atoms we will often abbreviate $ \{\bar{X}: p(X)\} \odot 
\{\bar{X}: q(X)\}$ by 
\begin{equation}\label{set-atom-a}
p \odot \{\bar{X}: q(\bar{X})\}. 
\end{equation}
We illustrate the use of set atoms in the bodies of program rules
by the following example:

\begin{example}[Set atoms in the rule body]\label{e15}
Consider a knowledge base containing two complete lists of atoms:
\begin{verbatim}
taken(mike,cs1).         required(cs1).
taken(mike,cs2).         required(cs2).
taken(john,cs1).
\end{verbatim}
Set atoms allow for a natural definition of the new relation, 
$ready\_to\_graduate(S)$, which holds if student $S$ has taken all the
required classes from the second list:   
\begin{verbatim}
ready_to_graduate(S) :- {C:taken(S,C)} <= {C: required(C)}. 
\end{verbatim}
The intuitive meaning of the rule is reasonably clear.
Together with the closed world assumption:
\begin{verbatim}
-ready_to_graduate(S) :- not ready_to_graduate(S).
\end{verbatim}
The rules imply that Mike is ready to graduate while John is not.
\end{example}

\st
Semantics of set literals in the body of the rule is based on the
observation of the similarity between set atoms and aggregates.
Intuitively, a set atom (\ref{set-atom}) can be viewed as a boolean
aggregate defined on a pair of sets. As a result syntactically it can be involved in
a vicious circle definition and is a subject to the avoidance of
vicious circles principle. Technically, this is captured by the
following definition:

\begin{definition}[Set Atom Reduct and New Definition of Answer Set]\label{reduct3}
The \emph{set atom reduct} of a ground program $\Pi$ 
of $\mathcal{A}log$ with respect to a set of 
ground regular literals $S$ is obtained from $\Pi$ by 
\begin{enumerate}
\item removing all rules with bodies containing set atoms 
  which are not true in $S$. 
\item replacing every remaining set atom 
$\{\bar{X}_1:cond_1\} \odot  \{\bar{X}_2:cond_2\}$ in the body of rules  by 
  the sets $\{p(\bar{t}) : p(\bar{t}) \in ground(cond_1) \cap 
    S\}$ and $\{p(\bar{t}) : p(\bar{t}) \in ground(cond_2) \cap 
    S\}$
\end{enumerate}
$S$ is an \emph{answer set} of $\Pi$ if $S$ is an answer set of the 
set atom reduct of $\Pi$ with respect to $S$.
\end{definition}

\st
It is not difficult to check that, as expected, the answer set of the program
from Example \ref{e15} contains $ready\_to\_graduate(mike)$
and $\neg ready\_to\_graduate(john)$.
The next example shows how the semantics deals with definitions
containing vicious circles.

\begin{example}[Set atoms in the rule body]\label{e15}
Consider a program $P_8$
\begin{verbatim} 
p(a) :- p <= {X : q(X)}. 
q(a). 
\end{verbatim}
in which $p(a)$ is defined in terms of the complete extent of the set
$\{X: p(X)\}$. In accordance with the vicious circle principle no answer
set of this program can contain $p(a)$. There are only two candidates
for answer sets of $P_8$: $S_1 = \{q(a)\}$ and $S_2 =
\{q(a),p(a)\}$. The set atom reduct of $P_8$ with respect to $S_1$
is 
\begin{verbatim} 
p(a) :- q(a). 
q(a). 
\end{verbatim}
while set atom reduct of $P_8$ with respect to $S_2$ is 
\begin{verbatim} 
p(a) :- p(a),q(a). 
q(a). 
\end{verbatim}
Clearly, neither $S_1$ nor $S_2$ is an answer set of $P_8$. As
expected, the program is inconsistent.
\end{example}

\st 
{\bf Subsets statements in the rules heads}.

\st
In addition to allowing set atoms in bodies of rules 
we allow a limited form of such atoms in the rules heads,
where they would play the role similar to that of the choice rules of Lparse and Clingo.
Syntactically, we expand rules of $\mathcal{A}log$ by \emph{
subset introductions}: 
\begin{equation}\label{subset-def}
p \leq \{\bar{X}: q(\bar{X})\} \leftarrow body.
\end{equation}
The rule reads as, given the $body$,  \emph{let $p$ be an arbitrary
  subset of the set $\{\bar{X}: q(\bar{X})\}$}.
Similarly for rules
\begin{equation}\label{subset-def-a}
p < \{\bar{X}: q(\bar{X})\} \leftarrow body. 
\end{equation}
and
\begin{equation}\label{subset-def-b}
p= \{\bar{X}: q(\bar{X})\} \leftarrow body. 
\end{equation}

\begin{example}[Subset introduction rule]\label{e20}
According to this intuitive reading the program $P_9$:
\begin{verbatim}
q(a). 
p <= {X:q(X)}. 
\end{verbatim}
has answer sets $A_1 = \{q(a)\}$ where the set $p$ is empty and 
 $A_2 = \{q(a),p(a)\}$ where $p = \{a\}$. 
\end{example}
The formal definition of answer sets of programs allowing introduction 
of subsets which captures this intuition is given via a notion of 
\emph{subset introduction reduct}. (The definition is similar to that 
presented in \cite{gel02}). 

\begin{definition}[Subset Introduction Reduct]\label{reduct4}
The \emph{subset introduction reduct} of a ground program $\Pi$ 
of $\mathcal{A}log$ with respect to a set of 
ground regular literals $S$ is obtained from $\Pi$ by 
\begin{enumerate}
\item replacing every subset introduction rule of $\Pi$ which head is
  not true in $S$ by
$$\leftarrow body.$$
\item replacing every subset introduction rule of $\Pi$ which head is
  true in $S$ by
$$p(\bar{t}) \leftarrow body$$
for each $p(\bar{t}) \in S$.
\end{enumerate}
$S$ is an \emph{answer set} of $\Pi$ if it is an answer set of 
 subset introduction reduct of $\Pi$ with respect to $S$.
\end{definition}

\begin{example}[Subset Introduction Rule]\label{e25} 
Consider a program  $P_9$ from Example \ref{e20}.
The reduct of this program with respect to $A_1 = \{q(a)\}$  is
\begin{verbatim}
q(a). 
\end{verbatim}
and hence $A_1$ is an answer set of $P_9$.
The reduct of $P_9$ with respect to  $A_2 = \{q(a),p(a)\}$ is
\begin{verbatim}
q(a). 
p(a).
\end{verbatim}
and hence $A_2$ is also an answer set of $P_9$.
There are no other answer sets.
\end{example}
Our last example shows how subset introduction rule with equality can
be used to represent synonyms:
\begin{example}[Introducing Synonyms]\label{e26} 
Suppose we have a list of cars represented by atoms formed by
predicate symbol $car$. 
\begin{verbatim}
car(a).
car(b).
\end{verbatim}
The following rule
\begin{verbatim}
carro = {X:car(X)} :- spanish.
\end{verbatim}
allows to introduce a new name for this list for spanish speaking
people. 
Clearly, $car$ and $carro$ are synonyms. 
Since the set introduction reduct of the 
program $P_{10}$ consisting of the rules above and a fact
\begin{verbatim}
spanish.
\end{verbatim}
is 
\begin{verbatim}
spanish.
car(a). 
car(b). 
carro(a). 
carro(a). 
\end{verbatim}
the answer set $A$
of the program consists of atoms $\{spanish, car(a), car(b), carro(a),
carro(b)\}$. 
\end{example}

\noindent If $p$ from the subset introduction rule (\ref{subset-def})
 does not occur in the head of any other rule of the program,
rule (\ref{subset-def}) is very close in its meaning to the
choice rule 
$$\{p(\bar{X}) : q(\bar{X})\} \leftarrow body$$
introduced in \cite{nss02}.
However if this condition does not hold the meaning is different:

\begin{example}[Subset Introduction and Choice Rules]\label{e30}
Consider an $\mathcal{A}log$ program
\begin{verbatim}
q(a). 
q(b). 
r(a). 
P <= {X : q(X)}. 
P <= {X : r(X)}. 
\end{verbatim}
and a Clingo program
\begin{verbatim}
q(a). 
q(b). 
r(a). 
{p(X) : q(X)}. 
{p(X) : r(X)}. 
\end{verbatim}
The former has two answer sets
$A_1 = \{q(a), q(b), r(a)\}$ with $p = \{\ \}$ and $A_2 = \{q(a), q(b), r(a), p(a)\}$
with $p = \{a\}$. In both cases $p$, in accordance with the intuitive
meaning of  subset introduction in $\mathcal{A}log$  is a subset of both, $\{X:q(X)\}$
and $\{X:r(X)\}$. The latter program, however,
has two extra answer sets:   $A_3 = \{q(a), q(b), r(a), p(b)\}$ and 
$A_4 =  \{q(a), q(b), r(a), p(a), p(b)\}$ which do not have this property.
Hence, in general, the choice rule $\{p(X) : q(X)\}$ cannot be read as
\emph{let $p$ be a subset of $q$}.
\end{example}

We conclude by a short discussion of the relationship between the
semantics of the new language with the Rationality Principle.  
At the first glance it may seems that this principle is no longer
satisfied, since an answer set of a program with subset introduction
is not necessarily minimal. It can be a proper subset of another answer set of the
same program (see, for instance, $A_1$ and $A_2$ in the above example).
We claim, however, that this impression is false. First of all 
answer sets must satisfy rules of a program. The rule requires $p$ be
an \emph{arbitrary} subset of $\{X:q(X)\}$ -- hence each such subset
should belong to some answer set of a program.
  
\begin{claim}[Rule Satisfaction and Supportedness]\label{p1aa} Let 
  $A$ be an answer set of a ground $\mathcal{A}log$ program $\Pi$. 
Then 
\begin{itemize}
  \item $A$ satisfies every rule $r$ of $\Pi$. 
  \item If $p(\bar{t}) \in A$ then there is a rule $r$ from $\Pi$  such 
    that the body of $r$ is satisfied by $A$ and $p(\bar{t})$ is the only atom 
    in the head of $r$ which is true in $A$ or the head contains an
    atom $p \odot \{\bar{X} : q(\bar{X})\}$ such that $q(\bar{t}) \in A$\textcolor{blue}{\}}.
\end{itemize}
\end{claim}

\section{Conclusion and Future Work}\label{conclusion}

We presented an extension, $\mathcal{A}log$, of ASP which allows for the
representation of and reasoning with aggregates. We believe that the language satisfies
design criteria of simplicity of syntax and formal and informal
semantics. There are many ways in which this work can be continued.
The first, and simplest, step is to expand $\mathcal{A}log$ by
allowing choice rules similar to those of \cite{nss02}. This can be
done in a natural way by combining ideas from this paper and that from
\cite{gel02}. 
We also plan to investigate mapping of
$\mathcal{A}log$ into logic programs with arbitrary propositional formulas.
There are many interesting and, we believe, important questions 
related to optimization of the $\mathcal{A}log$ solver from Section \ref{solver}.
After clarity is reached in this area one will, of course, try to
address the questions of implementation.

\section{Acknowledgment} We would like to thank 
Amelia Harrison, Patrick Kahl, Vladimir Lifschitz, and
Tran Cao Son for useful comments. The authors' work was partially supported by NSF grant
IIS-1018031 and CNS-1359359.



\section{Appendix}

In this appendix, we give the proofs and the needed  background  for the properties of $\mathcal{A}log$ programs and the results comparing $\mathcal{A}log$ with existing approaches. 

\subsection{Properties of $\mathcal{A}log$ Programs}
\setcounter{claim}{0}

In this appendix, given an ${\mathcal{A}}log$ program $\Pi$, a set $A$ of literals and a rule $r \in \Pi$, 
we use $R_\mathcal{A}(r, A)$ to denote the rule obtained from $r$ in the aggregate reduct of $\Pi$ with respect to $A$. 
$R_\mathcal{A}(r,A)$ is $nil$, called an {\em empty rule}, if $r$ is discarded in the aggregate reduct. 
We use $R_\mathcal{A}(\Pi, A)$ to denote the aggregate reduct of $\Pi$, i.e.,  $\{R_\mathcal{A}(r, A): r \in \Pi \mbox{ and } R_\mathcal{A}(r, A) \neq nil\}$. 
 
\st
To prove Proposition \ref{p1} we will need two auxiliary lemmas.

\begin{lemma}\label{aux1}
Let $\Pi$ be a ground $\mathcal{A}log$ program which contains no
occurrences of aggregates, $A$ be an answer set of $\Pi$ and $R$ be
the set of all rules of $\Pi$ whose bodies are satisfied by $A$.
Then $A$ is a minimal 
(with respect to set-inclusion) set of literals which satisfies $R$.
\end{lemma} 

\st 
Proof. 

\st 
The fact that $A$ satisfies $R$ follows immediately from the 
definition of answer set. 
To prove minimality 
assume that 

\st 
(1) $B \subseteq A$ 

\st 
(2) $B$ satisfies $R$

\st 
and show that 

\st
(3) $B$ satisfies the reduct $\Pi^A$.

\st 
Consider a rule $head \leftarrow body^A$ from $\Pi^A$ such that 

\st 
(4) $B$ satisfies $body^A$. 

\st 
Since $body^A$ contains no default negation,
(1) and (4) imply that

\st 
(5) $A$ satisfies $body^A$. 

\st 
By definition of a reduct, 

\st 
(6) $A$ satisfies $body$

\st 
and hence the rule

\st 
(7) $head \leftarrow body$ is in $R$. 

\st 
By (2) and (7), $head$ is satisfied by $B$. 

\st 
Therefore,
$B$ satisfies $head \leftarrow body^A$ 
and, hence, (3) holds.

\st 
Since $A$ is an answer set of  $\Pi^A$,
(3) implies that

\st
(8) $B = A$

\st
which concludes the proof.
\hfill $\Box$

\begin{definition}[Supportedness] 
\label{support} 
Let $A$ be an answer set of a ground program $\Pi$ of
$\mathcal{A}log$. We say that a literal $p \in A$
is supported by a rule $r$ from $\Pi$  if
the body of $r$ is satisfied by $A$ and $p$ is the only literal 
    in the head of $r$ which is true in $A$.
\end{definition}
To show supportedness for $\mathcal{A}log$ with aggregates we need to
first prove this property for $\mathcal{A}log$ programs not containing
aggregates (similar result for disjunctive programs with finite rules 
can be found in \cite{bg94}).
\begin{lemma}[Supportedness for Programs without Aggregates]\label{aux2}
Let $A$ be an answer set of a ground program $\Pi$ of
$\mathcal{A}log$ which contains no occurrences of aggregates. 
Then 
\ee{
  \item $A$ satisfies every rule $r$ of $\Pi$. 
  \item If $p \in A$ then there is a rule $r$ from $\Pi$  
which supports $p$.
}
\end{lemma}
Proof:
\begin{enumerate}
\item  The first clause follows immediately from the definition of 
an answer set of a program without aggregates.

\item Let $p \in A$. 

\st
To prove the existence of a rule of $\Pi$ supporting $p$
we consider the set $R$ of all rules of $\Pi$ whose 
bodies are satisfied by $A$. 

\st
Suppose $p$ does not belong to the head of any rule from $R$.
Then $A - \{p\}$ also satisfies rules of $R$.
(Indeed, suppose that $A - \{p\}$ satisfies the body of a rule from 
$R$. Then, by definition of $R$, the rule's head is satisfied by $A$.
Since the head does not contain $p$, it is also satisfied by $A
- \{p\}$.)
This contradicts
the minimality condition from Lemma \ref{aux1}.)

\st Suppose now that for every rule $r \in R$ which contains an occurrence
of $p$ in the head $head(r) \cap A \not= \{p\}$. But then $A -
\{p\}$ would again satisfy $R$  which would contradict 
Lemma \ref{aux1}. This concludes the proof of the Lemma \ref{aux2}. \hfill $\Box$
\end{enumerate}

\st
Now we are ready to prove proposition \ref{p1}.

\begin{claim}[Rule Satisfaction and Supportedness] \label{prop:support} Let 
  $A$ be an answer set of a ground $\mathcal{A}log$ program $\Pi$. 
Then 
\ee{
  \item $A$ satisfies every rule $r$ of $\Pi$. 
  \item If $p \in A$ then there is a rule $r$ from $\Pi$  such 
    that the body of $r$ is satisfied by $A$ and $p$ is the only atom 
    in the head of $r$ which is true in $A$. 
}
\end{claim}

\noindent
Proof: Let 

\st (1) $A$ be an answer set of $\Pi$.

\st We first prove that $A$ satisfies every rule $r$ of $\Pi$. 
Let $r$ be a rule of $\Pi$ such that 

\st (2) $A$ satisfies the body of
  $r$. 
  
\st Statement (2) implies that every aggregate atom, if there is any, of the body of $r$ is satisfied by $A$.  By the definition of the aggregate reduct, there must be a non-empty rule
  $r^\prime \in R_\mathcal{A}(\Pi, A)$ such that

\st (3)  $r^\prime = R_\mathcal{A}(r,A)$.  

\st By the definition of aggregate reduct, $A$ satisfies the body of $r$ iff it satisfies that of $r^\prime$. Therefore, (2) and (3) imply that 

\st (4) $A$ satisfies the body of $r^\prime$. 

\st By the definition of answer set of $\mathcal{A}log$, (1) implies that 

\st (5) $A$ is an answer set of $R_\mathcal{A}(\Pi,A)$.

\st Since $R_\mathcal{A}(\Pi,A)$ is an ASP program, (3) and (5) imply that 

\st (6) $A$ satisfies $r^\prime$.  

\st Statements (4) and (6) imply $A$ satisfies the head of $r^\prime$ and thus the head of $r$ because  $r$ and and $r^\prime$ have the same head.

\st Therefore $r$ is satisfied by $A$, which concludes our
proof of the first part of the proposition.

\st We next prove the second part of the proposition. Consider $p \in A$.  (1) implies that $A$ is an
  answer set of $R_\mathcal{A}(\Pi,A)$. By Lemma \ref{aux2}
there is a  rule $r^\prime \in R_\mathcal{A}(\Pi,A)$ such that 

\st (7) $r^\prime$ supports $p$. 

\st Let $r \in \Pi$ be a rule such that $r^\prime = R_\mathcal{A}(r, A)$. By the definition of aggregate reduct, 
  
\st (8) $A$ satisfies the body of $r$ iff 
  $A$ satisfies that of $r^\prime$. 
  
\st Since $r$ and $r^\prime$ have the same heads, (7) and (8) imply that  
  rule $r$ of $\Pi$ supports $p$ in $A$, which concludes the proof of the second part of the proposition.
\hfill $\Box$

\begin{claim}[Anti-chain Property]
Let $A_1$ be an answer set of an $\mathcal{A}log$ program $\Pi$. 
Then there is no answer set $A_2$ of $\Pi$ such that $A_1$ is a proper 
subset of $A_2$. 
\end{claim}

\noindent
Proof:  Let us assume that there are $A_1$ and $A_2$ such that

\st
(1) $A_1 \subseteq A_2$ and

\st
(2) $A_1$ and $A_2$ are answer sets of $\Pi$

\st
and show that $A_1=A_2$. 

\st
Let $R_1$ and $R_2$ be the aggregate reducts of $\Pi$ with respect to
$A_1$ and $A_2$ respectively. Let us first show that $A_1$ satisfies the
rules of $R_2$. Consider

\st
(3) $r_2 \in R_2$.

\st
By the definition of aggregate reduct there is $r \in \Pi$ such that 

\st
(4) $r_2 = R_\mathcal{A}(r,A_2)$. 

\st
Consider 

\st
(5) $r_1 = R_\mathcal{A}(r,A_1)$.

\st
If $r$ contains 
no aggregate atoms then 

\st
(6) $r_1 = r_2$.  

\st
By (5) and (6), $r_2 \in R_1$ 
and hence, by (2) $A_1$ satisfies $r_2$.

 \st
Assume now that $r$ contains one aggregate term, $f\{X:p(X)\}$, i.e. $r$ is of the form 

\st 
(7) $h \leftarrow B,C(f\{X:p(X)\})$

\st
where $C$ is some property of the aggregate.

\st 
Then $r_2$ has the form 
 
\st 
(8) $h \leftarrow B,P_2$

\st 
where 

\st 
(9) $P_2 =  \{p(t) : p(t) \in A_2\}$ and $f(P_2)$ satisfies condition $C$.

\st
Let

\st
(10) $P_1 =  \{p(t) : p(t) \in A_1\}$



\st
and consider two cases: 

\st
(11a)  $R_\mathcal{A}(r,A_1) = \emptyset$.

\st
In this case $C(f(P_1))$ does not hold. Hence, $P_1 \not= P_2$.
Since $A_1 \subseteq A_2$ we have that $P_1 \subset P_2$, the body
of rule (8) is not satisfied by $A_1$, and hence the rule (8) is.

\st
(11b) $R_\mathcal{A}(r,A_1) \not= \emptyset$.

\st 
Then $r_1$ has the form 
 
\st 
(12) $h \leftarrow B,P_1$

\st 
where 

\st 
(13) $P_1 =  \{p(t) : p(t) \in A_1\}$ and $f(P_1)$ satisfies condition $C$.

\st
Assume that $A_1$ satisfies the body, $B,P_2$, of rule (8). 
Then 

\st
(14) $P_2 \subseteq A_1$

\st
This, together with (9) and (10) implies

\st
(15) $P_2 \subseteq P_1$.

\st
From (1),
(9), and (10) we have $P_1 \subseteq P_2$.  Hence

\st
(16) $P_1 = P_2$. 

\st
This means that $A_1$ satisfies the body of $r_1$
and hence it satisfies $h$ and, therefore, $r_2$.

\st
Similar argument works for rules containing multiple aggregate atoms
and, therefore, $A_1$ satisfies $R_2$. 

\st
Since $A_2$ is a minimal set satisfying $R_2$ and $A_1$ satisfies
$R_2$ and $A_1 \subseteq A_2$ we have that $A_1 = A_2$.

\st
This completes our proof. \hfill $\Box$

\begin{claim}[Splitting Set Theorem]
Let 
$\Pi_1$ and $\Pi_2$ be ground programs of $\mathcal{A}log$ such that
no ground literal
occurring in $\Pi_1$ is a head literal of $\Pi_2$, and 
$S$ be a set of ground
  literals containing those occurring in the rules of $\Pi_1$ and not
  containing those occurring in the heads of rules of
  $\Pi_2$\footnote{We refer to such $S$ as a \emph{splitting set} of $\Pi_1
    \cup \Pi_2$.}.
Then 

\st
(3) $A$ is an answer set of $\Pi_1 \cup \Pi_2$

\st
iff

\st
(4a)  $A \cap S$  is an answer set of $\Pi_1$ and

\st
(4b) $A$ is an answer set of $(A \cap S) \cup \Pi_2$.
\end{claim}

\st
Proof. 

\st 
First consider the case when $\Pi_1$ and $\Pi_2$ contain no aggregates. 
It is easy to check that, in this case, we can use the proof from
\cite{ErdoganL03}, since the infinite number of literals in the rules does not affect the
arguments used in this proof.
Proof of the proposition for programs with aggregates may be easily reduced to this case.
To see that first notice that, by the definitions of an answer set and
an aggregate reduct 

\st 
(3) holds iff

\st
(5) $A$ is an answer set of $R_\mathcal{A}(\Pi_1,A) \cup R_\mathcal{A}(\Pi_2,A)$.

\st
By definition, the aggregate reduct may only replace aggregate literals of a rule $r$ by
regular literals implicitly occurring in this rule. Hence no literal occurring
in $R_\mathcal{A}(\Pi_1,A)$ is a head literal occurring in 
$R_\mathcal{A}(\Pi_2,A)$. Moreover, the head literals of $\Pi_1$ and $\Pi_2$ 
may only be removed by the aggregate reduct and hence $S$ is a
splitting set of $R_\mathcal{A}(\Pi_1,A) \cup R_\mathcal{A}(\Pi_2,A)$.
This means that
$R_\mathcal{A}(\Pi_1,A)$, $R_\mathcal{A}(\Pi_2,A)$, and $S$ also 
satisfy conditions of the Splitting Set Theorem.  Since 
$R_\mathcal{A}(\Pi_1,A)$ and $R_\mathcal{A}(\Pi_2,A)$
contain no aggregate atoms this implies that (5) holds
iff 

\st
(6a)  $A \cap S$  is an answer set of $R_\mathcal{A}(\Pi_1,A)$

\st
and

\st
(6b) $A$ is an answer set of $(A \cap S) \cup R_\mathcal{A}(\Pi_2,A)$.

\st 
Now we show that (4a) holds iff (6a) holds. By the definition of an 
answer set we have that (4a) holds iff 

\st 
(7) $A \cap S$ is an answer set of $R_\mathcal{A}(\Pi_1,A \cap S)$.

\st 
Note that the truth of an aggregate literal $L$ of $\Pi_1$ depends only on the
truth of regular literals occurring in it. Since all such regular
literals belong to $S$, $L$ is true in $A$ iff $L$ is true in $A \cap
S$. Moreover, the aggregate
reduct replaces aggregate literals of $\Pi_1$ which are true in $A$ only by literals from $S$ and hence
we have that

\st 
(8) $R_\mathcal{A}(\Pi_1,A) = R_\mathcal{A}(\Pi_1,A \cap S)$.

\st
and, from (7) and (8),

\st
(9)  (4a) iff (6a).

\st
It remains to show that  (4b) holds iff (6b) holds.

\st 

\st
By the definition of an aggregate reduct , 

\st
(10)  $(A \cap S) \cup R_\mathcal{A}(\Pi_2,A) = R_\mathcal{A}((A \cap S) \cup \Pi_2,A)$

\st
and hence, by the definition of an answer set we have

\st
(11) (4b) iff (6b)

\st
which completes the proof of our theorem. \hfill $\Box$


\begin{claim}[Complexity]
The problem of checking if a ground atom $a$ belongs to all answer 
sets of a finite $\mathcal{A}log$ program is $\Pi_2^P$ complete. 
\end{claim}

\noindent We restrict this result of finite 
$\mathcal{A}log$ program only as we did in our previous work whose proof can be found in the appendix of \cite{GelfondZ14}.

\subsection{Correctness of the Algorithm}

We list the algorithm $Solver$ below for computing an answer set of an $\mathcal{A}log$ program. 
Given a set $A$ of atoms, recall $compl(A) = \{a: a \in A\} \cup \{not~ a: a \not \in A \}$. 
\begin{tabbing}
abcde\=ab\=ab\=ab\aa\=aa\=\kill
{\bf function} Solver \\
\> {\bf input}: \>\>\> partial interpretation $I_0$, sets $TA_0$ and $FA_0$ of
aggregate e-atoms, and \\
\>\>\> \>  an $\mathcal{A}log$ program $\Pi_0$ such that any answer set of  $\Pi_0$ compatible with $I_0$\\
\>\>\> \>  is compatible with $TA_0$ and $FA_0$; \\
\> {\bf output}: \>\>\> $<I, true>$ where $I$ is an answer set of  $\Pi_0$ compatible with $I_0$; \\
\>\>\> \> $<I_0, false>$ if no such answer set exists; \\
{\bf var} 
$I$: set of e-atoms; $TA,FA$: set of aggregate e-atoms; $X$: boolean; \\
{\bf begin} \\
2. \> $I$ := $I_0$; \\
3. \> $<I, TA, FA, X>$ := Cons($I$, $TA_0$, $FA_0$, $\Pi_0$);\\
4. \> {\bf if} $X$ = false {\bf then}  \\
5. \> \> {\bf return} $<I_0, false>$; \\
6. \> {\bf if} no regular atom is undecided in $I$ {\bf then} \\
7. \> \> {\bf if} IsAnswerSet($I$, $\Pi_0$) {\bf then} \\
8. \> \> \> {\bf return} $<I, true>$; \\
9. \> \> {\bf else} {\bf return} $<I_0, false>$; \\
10. \> select a ground atom $p$ undecided in $I$; \\
11. \> $<I, X>$ := Solver($I \cup \{p\}, TA, FA, \Pi_0$); \\
12. \> {\bf if} $X$ = true {\bf then} \\
13. \> \> {\bf return} $<I, X>$; \\
14. \> {\bf return} Solver(($I - \{p\}$) $\cup$ \{not $p$\}, TA, FA, $\Pi_0$); \\
{\bf end};
\end{tabbing}

\noindent The function IsAnswerSet is listed below. 

\begin{tabbing}
ab\=ab\=ab\=ab\aa\=aa\=\kill
{\bf function} IsAnswerSet \\
\> {\bf input}: interpretation $I$ and program $\Pi$; \\
\> {\bf output}: {\em true} if $I$ is an answer set of $\Pi$; {\em false} otherwise; \\
{\bf begin} \\
1. \> Compute the aggregate reduct $R$ of $\Pi$ wrt $I$; \\
2. \> {\bf If} $atoms(I)$ is an answer set of $R$; \\
\> \> {\bf return} {\em true} \\
\> {\bf else} {\bf return} {\em false} \\
{\bf end};
\end{tabbing}

\noindent Koch et al. \cite{KochLP03} present an algorithm to translate the test in line 2 into a SAT problem. So, we don't provide refinement of line 2 here.   


\begin{lemma}[Local Property of $Cons$] \label{lm:lowerBound}
Let $I_1$, $TA_1$ and $FA_1$  be the value of $I$, $TA$ and $FA$ respectively before the execution of line 7, and $I_2$, $TA_2$ and $FA_2$ the value of $I$, $TA$ and $FA$ respectively after the execution of line 7. For any answer set $A$ of $\Pi_0$, it is compatible with $I_1$, $TA_1$ and $FA_1$ iff it is compatible with $I_2$, $TA_2$ and $FA_2$. 
\end{lemma}

\st Proof.

\st $\Longrightarrow$: Assuming 

\st \ml{lb00} $A$ is compatible with $I_1$, $TA_1$ and $FA_1$,

\st we prove in part 1, $I_2 \subseteq compl(A)$, and in part 2, $A$ is compatible with $TA_2$ and $FA_2$.

\begin{enumerate}

\item For any e-atom $l \in I_2$, there are two cases.

\st Case 1: $l \in I_1$. Since $I_1 \subseteq compl(A)$, $l \in compl(A)$. 

\st Case 2: $l \notin I_1$. $l$ must be obtained from one of the four inference rules (Line 6 of $Cons$). 

\st Case 2.1: $l$ is obtained by the first inference rule, i.e., there is a rule $r$ of $\Pi_0$ whose body is strongly satisfied by $I_1$ and all atoms except $l$ in the head of $r$ are false in $I_1$. Since $A$ is compatible with $I_1$, $A$ satisfies the body of $r$ and every atom in the head of $r$ except $l$ is false in $A$. Since $A$ is an answer set of $\Pi_0$, $A$ satisfies $r$ and thus $l$ must be in $A$, i.e., $l \in compl(A)$. 

\st Case 2.2: $l$ is obtained from the rule $r$ of $\Pi_0$ by the second inference rule. By the inference rule, $l$ occurs in the body of $r$ or its contrary
in the head of $r$, and there is an atom $p \in I_1$ of the head of $r$ such that $r$ is the only rule of $\Pi_0$ whose head contains $p$.  Since $I_1 \subseteq compl(A)$, $p \in A$. 
Consider two cases. 

In the first case, $l \in body(r)$. We prove $l \in compl(A)$ by contradiction. Assume $l$ is false in $A$. If $l$ is a negative regular e-atom, then $R_{\mathcal{A}}(\Pi_0, A)^A$ does not contain any rule whose head containing $p$. So, $p \notin A$, contradicting $p \in A$. 
If $l$ is a positive regular e-atom,  let rule $r' \in R_{\mathcal{A}}(\Pi_0, A)^A$ be the reduct of $r$. Since $l \notin A$, $A - \{p\}$ satisfies $r'$ and all other rules of $R_{\mathcal{A}}(\Pi_0, A)^A$. Since $p \in A$, $A - \{p\} \subset A$, contradicting that $A$ is an answer set of $R_{\mathcal{A}}(\Pi_0, A)^A$.

In the second case, $\bar{l} \in head(r)$ where $\bar{l}$ is the contrary of $l$. We prove $l \in compl(A)$ by contradiction. Assume $\bar{l} \in A$. There must be an $r' \in R_{\mathcal{A}}(\Pi_0, A)^A$ which is obtained from $r$. Since $\bar{l} \in A$, $A-\{p\}$ satisfies $r'$ and all other rules of $R_{\mathcal{A}}(\Pi_0, A)^A$. It contradicts that $A$ is an answer set of $R_{\mathcal{A}}(\Pi_0, A)^A$ because $A -\{p\} \subset A$. 

\st Case 2.3: $l$ is obtained from $r$ of $\Pi_1$ by the third inference rule. By the inference rule, every atom of the head of  $r$ is false, $\bar{l}$ is the only premise which is undecided, and every other premise of $r$ is strongly satisfied by $I_1$. 
Since $I_1 \subseteq compl(A)$, 

\st \ml{lb2} every atom of the head of $r$ is false wrt $compl(A)$, and 

\st \ml{lb3} every e-atom $l_1$ of the body of $r$, except $\bar{l}$, is satisfied by $compl(A)$ too.

\st We prove $l \in compl(A)$ by contradiction. Assume $l \notin compl(A)$, i.e, 

\st \ml{lb5} $\bar{l} \in compl(A)$. 

\st By \mr{lb2} -- \mr{lb5}, there is $r' \in R_{\mathcal{A}}(\Pi_0, A)^A$ which is obtained from $r$, and 
$A$ satisfies the body of $r'$. However, no atom of the head of $r'$ is true in $A$. Therefore, $A$ does not satisfy $r'$, contradicting that $A$ is an answer set of $R_{\mathcal{A}}(\Pi_0, A)^A$. Hence $l \in I(A)$.  

\st Case 2.4: $l$ is obtained by the forth inference rule. By this inference rule, $l$ must be of the form $(not~ p)$ where $p$ is an atom, and the body of every rule of $\Pi_0$ with $p$ in its head is strongly refuted 
by $I_1$. 
For any rule $r$ with $p$ in its head, there are three cases. The first case is that some positive regular e-atom $l \notin A$. The second case is that some negative e-atom $l \notin compl(A)$. The last case is that some aggregate e-atom is false or undefined in $compl(A)$. 
Therefore, in $R_{\mathcal{A}}(\Pi_0, A)^A$, all rules with $p$ in their heads are obtained from rules with the first case. 
Hence, $A - \{p\}$ satisfies all rules with $p$ in its head and all other rules of $R_{\mathcal{A}}(\Pi_0, A)^A$, which contradicts that $A$ is an answer set of $R_{\mathcal{A}}(\Pi_0, A)^A$ because $A - \{p\} \subset A$. 

\st In summary, from Case 2.1 to 2.4, we have  $l \in compl(A)$. 

\st

\item We now prove that $A$ is compatible with $TA_2$ and $FA_2$. 

\st For any aggregate e-atom $agg \in TA_2$,  we will prove \mr{lb8}. Consider two cases.

\st Case 1:  $agg \in TA_1$. Since $A$ is compatible with $TA_1$ and $FA_1$, \mr{lb8} holds. 

\st Case 2: $agg \notin TA_1$. In this case, it is added to $TA_2$ only by the second inference rule. By the inference rule, $agg$ occurs in the body of $r$, and there is an atom $p \in I_1$ of the head of $r$ such that $r$ is the only rule of $\Pi_0$ whose head contains $p$. Since $I_1 \subseteq compl(A)$, $p \in A$. 
Assume \mr{lb8} doesn't hold. There is no rule in $R_{\mathcal{A}}(\Pi_0, A)^A$ with $p$ in the head because $r$ is discarded from $R_{\mathcal{A}}(\Pi_0, A)^A$.  So, $A - \{p\}$ satisfies all rules of $R_{\mathcal{A}}(\Pi_0, A)^A$. Since $p \in A$, $A - \{p\} \subset A$, contradicting that $A$ is an answer set of $R_{\mathcal{A}}(\Pi_0, A)^A$. Therefore, the following statement holds.

\st \ml{lb8} $agg$ is not strongly refuted by $A$. 

\st Next, for any aggregate atom $agg \in FA_2$,  we will prove \mr{lb9}. Consider two cases.

\st Case 1: $agg \in FA_1$. Since $A$ is compatible with $TA_1$ and $FA_1$, \mr{lb9} holds. 

\st Case 2: $agg \notin FA_1$. $agg$ is added to $FA_2$ only by the third inference rule. By the inference rule, there is a rule $r$ of $\Pi_0$ such that all atoms of its head is false in $I_1$, $agg$ is the only aggregate e-atom in the body of $r$ that is not in $FA_1$, and 

\st \ml{lb80} the rest premises of $r$ are strongly satisfied by $I_1$. 

\st We prove \mr{lb9} by contradiction. Assume \mr{lb9} is false, i.e., 

\st \ml{lb81} $agg$ is strongly satisfied by $A$.

\st By \mr{lb80} and \mr{lb81}, there is rule $r' \in R_{\mathcal{A}}(\Pi_0, A)^A$ that is obtained from $r$, and  the body of $r'$ is satisfied by $A$. However, the head of $r'$ is falsified by $A$. So, $A$ is not an answer set of $R_{\mathcal{A}}(\Pi_0, A)^A$, contradicting that $A$ is an answer set of $R_{\mathcal{A}}(\Pi_0, A)^A$. Therefore, we have 

\st \ml{lb9} $agg$ is not strongly satisfied by $A$. 
\end{enumerate}
  
\noindent $\Longleftarrow$: Assuming 

\st \ml{lb91} $A$ is compatible with $I_2$, $TA_2$ and $FA_2$, 

\st we prove $I_1 \subseteq compl(A)$, and $A$ is compatible with $TA_1$ and $FA_1$.

\st Since $I_1 \subseteq I_2$, and $A$ is compatible with $I_2$, we have $I_1 \subseteq compl(A)$. 

\st Since $TA_1 \subseteq TA_2$ and $FA_1 \subseteq FA_2$, that $A$ is compatible with $TA_2$ and $FA_2$ implies that $A$ is compatible with $TA_1$ and $FA_1$.  \hfill $\Box$

\st For $Cons$, we have the following property. 

\begin{lemma}[Property of $Cons$] \label{lm:cons}
Given a program $\Pi_0$, a partial interpretation $I_0$, and two sets $TA_0$ and $FA_0$ of aggregate atoms, let $I$, $TA$ and $FA$ be the values returned by $Cons(I_1, TA_0, FA_0, \Pi_0)$.  

\ee{
  \item $I_0 \subseteq I$, $TA_0 \subseteq TA$, $FA_0 \subseteq FA$, and
  \item 
  For every answer set $A$ of $\Pi_0$,  $A$ is compatible with $I_0$, $TA_0$ and $FA_0$ iff  $A$ is compatible with $I$, $TA$ and $FA$.
  \item If there is an answer set of $\Pi_0$ compatible with $I_0, TA_0, FA_0$, then {\em Cons($I_0, TA_0, FA_0, \Pi_0$)} returns $<I, TA, FA, true>$. 
}
\end{lemma}

\noindent Proof. 


\begin{enumerate}
\item  By lines 1 and 7, $I_0 \subseteq I$, $TA_0 \subseteq TA$, $FA_0 \subseteq FA$. 

\st

\item Let $I_1$, $TA_1$ and $FA_1$ be the value of $I$, $\Pi$, $TA$ and $FA$ respectively before the execution of line 7, and $I_2$, $TA_2$ and $FA_2$ the value of $I$, $TA$ and $FA$ after the execution of line 7. By Lemma~\ref{lm:lowerBound}, we have

\st \ml{con1} for any answer set $A$ of $\Pi_0$, $A$ is compatible with $I_1$, $TA_1$ and $FA_1$ iff $A$ is compatible with $I_2$, $TA_2$ and $FA_2$.

\st From line 7, the value of $I$ is non-decreasing, the finite number of ground atoms in the signature of $\Pi_0$ ensures that the termination condition in line 8 will be satisfied in a finite number of steps. 

\st Let $I_0, TA_0, FA_0, \Pi_0$ be the values of $I, TA, FA$ and $\Pi$ before the {\bf repeat} loop, and $I, TA, FA$ be the values of $I, TA$ and $FA$ after the loop. After the termination of the {\bf repeat} loop,  the property \mr{con1} ensures that for every answer set $A$ of $\Pi_0$,  it is compatible with $I_0$, $TA_0$ and $FA_0$ iff  it is compatible with $I$, $TA$ and $FA$. Lines 9 to 11 do not change the property above.  

\st

\item  Assume there is an answer set of $\Pi_0$ compatible with $I_0, TA_0, FA_0$. By the second part of this proposition, we have

\st \ml{con2} $A$ is compatible with $I$, $TA$ and $FA$ where $I$, $TA$ and $FA$ are the values of $I$, $TA$ and $FA$ after Line 8.

\st To prove {\em Cons($I_0, TA_0, FA_0, \Pi_0$)} returns $<I, TA, FA, true>$ by contradiction, we assume {\em Cons} returns $<I_0, TA_0, FA_0, false>$. By the {\bf if} clause starting from Line
9, one of the following holds: $I$ is not consistent, some aggregate atom of $TA$ is falsified by $I$, or some aggregate atom of $FA$ is satisfied by $I$, which contradicts with \mr{con2}. 
\hfill $\Box$
\end{enumerate}

\st For $IsAnswerSet$, we have the following property. 

\begin{lemma}[Property of $IsAnswerSet$] \label{lm:IsAnswerSet}
Given a program $\Pi$, a complete interpretation $I$, $IsAnswerSet(I, TA, FA, \Pi)$ returns true iff $\{l\in I: l \mbox{ is an atom}\}$ is an answer set of $\Pi$. 
\end{lemma}

\noindent This result follows from the definition of the answer sets of a program. 

\begin{claim}[Correctness of $Solver$]
Given a program $\Pi_0$, a partial interpretation $I_0$, and two sets $TA_0$ and $FA_0$ of aggregate atoms such that any answer set of $\Pi_0$ compatible with $I_0$ is compatible with $TA_0$ and $FA_0$, Solver($I_0, TA_0, FA_0, \Pi_0$) returns $<I, true>$
if and only if there is an answer set of $\Pi_0$ that is compatible with $I_0$, $TA_0$ and $FA_0$. 
\end{claim}

\noindent Proof. 

\st $\Longrightarrow$: Assuming

\st \ml{solv90} $Solver(I_0, TA_0, FA_0, \Pi_0)$ returns $<I, true>$, 

\st we will prove 

\st \ml{solv91} $I$ is an answer set of $\Pi_0$ that is compatible with $I_0$, $TA_0$ and $FA_0$

\st by induction on the number of undecided atoms in $I_0$. Let $I$, $TA_1$, $FA_1$, and $X$ be the results returned by {\em Cons} (Line 3 of $Solver$). 

\st Base case: there is no undecided atom in $I_0$. By \mr{solv90} and the algorithm $Solver$, line 8 of $Solver$ is executed before it returns $\langle I, true \rangle$. The function call {\em IsAnswerSet}($I, \Pi_0$) in Line 7 must return true. By Lemma~\ref{lm:IsAnswerSet}, $\{l\in I: l \mbox{ is an atom}\}$ is an answer set of $\Pi_0$. By Lemma~\ref{lm:cons},  $I_0 \subseteq I$ and thus $I$ is compatible with $I_0$. Hence, $I$ is compatible with $TA_0$ and $FA_0$ by the given assumption on $TA_0$ and $FA_0$ in the proposition. 

\st Inductive hypothesis: given a number $k > 0$, assume for all $n < k$, the claim holds for $I_0$ with $n$ undecided atoms. 

\st We next prove when $I_0$ has $k+1$ undecided atoms, \mr{solv91} holds. 

\st Let $I_1$, $TA_1$, $FA_1$, and $X$  be the results returned by $Cons$ in Line 3. The statement \mr{solv90} implies that $X$ (in Line 3) must be true. Now consider two cases.

\st Case 1: there is no undecided atoms in $I_1$. By \mr{solv90}, {\em IsAnswerSet$(I_1, TA_1, FA_1, \Pi_0)$} must return true. $\{l\in I_1: l \mbox{ is an atom}\}$  is an answer set of $\Pi_0$ by Lemma~\ref{lm:IsAnswerSet}. Since $I_0  \subseteq I_1$ by Lemma~\ref{lm:cons}, $I_1$ is compatible with $I_0$, and thus compatible with $TA_0$ and $FA_0$ by the assumption on $TA_0$ and $FA_0$. So is $I$ because $I_1 = I$. 

\st Case 2: there is at least one undecided atoms in $I_1$. 
\hide{
\st First, for any answer set $A$ of $\Pi_0$ compatible with $I_1 \cup \{p\}$ or  $I_1 \cup \{not~ p\}$, we prove \mr{solv911}. Clearly $A$ is compatible with $I_1$. Since $I_0 \subseteq I_1$, $A$ is compatible with $I_0$ and thus $TA_0$ and $FA_0$. By Lemma~\ref{lm:cons}, 

\st \ml{solv911}$A$ is compatible with $TA_1$ and $FA_1$.
}

\st By Line 11, we have two cases. 

\st Case 2.1: {\em Solver($I_1 \cup \{p\}, TA_1, FA_1, \Pi_0)$} returns $<I, true>$. Since $I_1 \cup \{p\}$ has at most $k$ undecided atoms, by inductive hypothesis, 

\st \ml{solv92} $\{l \in I: \mbox{$l$ is an atom}\}$ is an answer set of $\Pi_0$ compatible with $I_1 \cup \{p\}$, $TA_1$ and $FA_1$. 

\st Case 2.2: {\em Solver($I_1 \cup \{p\}, TA_1, FA_1, \Pi_0)$} returns $<I, false>$. In this case, {\em Solver($I_1 \cup \{not ~p\}, TA_1, FA_1, \Pi_0)$} must return $<I, true>$. Since $I_1 \cup \{not~ p\}$ has at most $k$ undecided atoms, by inductive hypothesis, 

\st \ml{solv93} $\{l \in I: \mbox{$l$ is an atom}\}$ is an answer set of $\Pi_0$ compatible with $I_1 \cup \{not ~p\}$, $TA_1$ and $FA_1$. 

\st Since $I_0 \subseteq I_1$, $I_1 \subseteq I_1 \cup \{p\}$, and $I_1 \subseteq I_1 \cup \{not~ p\}$, $I$ is compatible with $I_0$ by \mr{solv92} and \mr{solv93}, and thus compatible with $TA_0$ and $FA_0$. 

\st By case 1 and 2, $I$ is an answer set of $\Pi_0$ that is compatible with $I_0$, $TA_0$ and $FA_0$.

\st $\Longleftarrow$: Assuming

\st \ml{solv931} there is an answer set $A$ of $\Pi_0$ that is compatible with $I_0$, $TA_0$ and $FA_0$,

\st we will prove  {\em Solver}($I_0, TA_0, FA_0$, $\Pi_0$) returns $<I, true>$ and $I$ is an answer set of $\Pi_0$ compatible with $I_0$, $TA_0$ and $FA_0$, by induction on the number $n$ of undecided e-atoms in $I_0$. 

\st  Let $I_1$, $TA_1$, $FA_1$ and $X$ be the results returned by {\em Cons($I_0, TA_0, FA_0, \Pi_0$)} (Line 3 of $Solver$). By 
Lemma~\ref{lm:cons},

\st \ml{solv01} $A$ is compatible with $I_1$ and $TA_1$ and $FA_1$.

\st Base case $n=0$. There is no undecided e-atom in $I_0$. Since there is no undecided atoms in $I_0$ and $A$ is an answer set of $\Pi_0$ compatible with $I_0$, $I_0 = I_1 = compl(A)$. 
{\em Cons} must return true (among other parameters) by Lemma~\ref{lm:cons} and the fact that $A$ is an answer set of $\Pi_0$ compatible with $I_0$ (and thus $TA_0$ and $FA_0$). Since there is no undecided atoms in $I_1$, Line 7 will be executed. Since $\{l \in I_1: \mbox{$l$ is an atom}\}$ is an answer set of $\Pi_0$, {\em IsAnswerSet}($I_1, TA_1, FA_1, \Pi_0 $) must return true by Lemma~\ref{lm:IsAnswerSet}. Therefore, Line 8 is executed and $<I_1, true>$ is returned by $Solver(I_0, TA_0, FA_0, \Pi_0)$. 

\st Inductive hypothesis: given a number $k > 0$, for any number $n < k$, if $I_0$ has $n$ undecided atoms and there is an answer set of $\Pi_0$ that is compatible with $I_0$, $TA_0$ and $FA_0$, then $solver(I_0, TA_0, FA_0, \Pi_0)$ returns $<I, true>$ where $I$ is an answer set of $\Pi_0$ compatible with $I_0$, $TA_0$ and $FA_0$. 

\st We will prove, if $I_0$ has $k+1$ undecided atoms and there is an answer set of $\Pi_0$ that is compatible with $I_0$, then $solver(I_0, TA_0, FA_0, \Pi_0)$ returns $<I, true>$ where $I$ is an answer set of $\Pi_0$ that is compatible with $I_0$, $TA_0$ and $FA_0$. 


\st We consider two cases below. 

\st Case 1: there is no undecided atom in $I_1$. So, $I_1 = compl(A)$ by \mr{solv01}. Line 7 will be executed, and by \mr{solv931} and Lemma~\ref{lm:IsAnswerSet}, {\em IsAnswerSet} must return true and thus $<compl(A), true>$ is returned by $Solver1(I_0, TA_0, FA_0, \Pi_0)$. 
By Lemma~\ref{lm:cons}, $I_1$ is compatible with $I_0$, which in turn implies that $I_1$ is compatible with $TA_0$ and $FA_0$. 

\st Case 2: there is some undecided atom in $I_1$. Let the atom selected in Line 10 be $p$. 

\hide{
\st We show for any answer set $S$ of $\Pi_0$ compatible with $I_1 \cup \{p\}$ or  $I_1 \cup \{not~ p\}$, we prove \mr{solv20}. Clearly $S$ is compatible with $I_1$. Since $I_0 \subseteq I_1$, $S$ is compatible with $I_0$ and thus $TA_0$ and $FA_0$. By Lemma~\ref{lm:cons}, 

\st \ml{solv20}$S$ is compatible with $TA_1$ and $FA_1$.

\st $A$ is compatible with both $I_1 \cup \{p\}$ and $I_1 \cup \{not ~p\}$ because of \mr{solv01}.
}

\st Consider the following two cases.

\st Case 2.1: there is an answer set $B$ 
of $\Pi_0$ that is compatible with $I_2=I_1 \cup \{p\}$. By inductive hypothesis, $Solver(I2, TA_1, FA_1, \Pi_0)$ (Line 11) will return $<I, true>$ where 

\st \ml{solv2} $\{l \in I: \mbox{$l$ is an atom}\}$ is an answer set of $\Pi_0$ that is compatible with $I_2$.

\st Since $I_0 \subseteq I_1 \subseteq I_2$, \mr{solv2} implies that $I$ is compatible with $I_0$ (and thus $TA_0$ and $FA_0$). 

\st Case 2.2: there is no answer set compatible with $I_2$. $Solver(I_2, TA_1, FA_1, \Pi_0)$ will return false (by the sufficient condition we have proved). $A$ must be compatible with $I_3=I_1 \cup \{not~ p\}$ because $A$ is compatible with $I_1$ and not compatible with $I_2$.  
Since $I_3$ has at most $k$ undecided atoms, and $A$ is an answer set of $\Pi_0$ compatible with $I_1$, $TA_1$ and $FA_1$, $Solver(I3, TA_1, FA_1, \Pi_0)$ will return $<I, true>$ where 

\st \ml{solv3} $\{l \in I: \mbox{$l$ is an atom}\}$ is an answer set of $\Pi_0$ and compatible with $I_3$, $TA_1$ and $FA_1$. 

\st Since $I_0 \subseteq I_1 \subseteq I_3$, \mr{solv3} implies that $I$ is compatible with $I_0$ and thus $TA_0$ and $FA_0$. 

\st In summary, $Solver(I_0, TA_0, FA_0, \Pi_0)$ returns $<I, true>$ where $\{l \in I: \mbox{$l$ is an atom}\}$ is an answer set of $\Pi_0$ that is compatible with $I_0$, $TA_0$ and $FA_0$.  $\hfill \Box$

\subsection{Comparison with Other Approaches}

We give below the proofs for the results on comparing $\mathcal{A}log$ with existing approaches. 

\subsubsection{Comparison with $\mathcal{F}log$}
\label{app:flog}

\setcounter{claim}{5}

\setcounter{statement}{0}

\begin{claim}[From $\mathcal{A}log$ Answer Sets to $\mathcal{F}log$ Answer Sets] 
If $\Pi$ is  $\mathcal{AF}$ compatible program 
and $A$ is an $\mathcal{A}log$ answer set of $\Pi$ then it is an $\mathcal{F}log$ answer set of $\Pi$.  
\end{claim}

\noindent 
Proof. 

\st Given the properties of $\Pi$, its ground program in $\mathcal{F}log$ is the same as the one in $\mathcal{A}log$ except the form of the aggregate atoms. Given a set $S$ of literals and the assumption that no aggregate functions are partial, an aggregate atom in $\mathcal{F}log$ is satisfied by $S$ iff the same atom in $\mathcal{A}log$ is satisfied by $S$. 

\st To prove $A$ is an $\mathcal{F}log$ answer set of $\Pi$, we prove $A$ is a minimal set satisfying $R_{\mathcal{F}}(\Pi, A)$. 

\st We first show that $A$ satisfies $R_{\mathcal{F}}(\Pi, A)$, i.e., \mr{AF51}. 

\st For any rule

\st
\ml{AF1} $r \in R_{\mathcal{F}}(\Pi, A)$,  

\st
\ml{AF2} $r \in \Pi$ by definition of $\mathcal{F}log$ reduct. 

\st Since $A$ is an $\mathcal{A}log$ answer set of $\Pi$, 

\st \ml{AF3} $A$ satisfies every rule of $\Pi$ by Proposition~\ref{p1}. Hence, \mr{AF2} implies 
$A \models r$. Therefore,

\st \ml{AF51} $A  \models R_{\mathcal{F}}(\Pi, A)$. 

\st We next prove $A$ is a minimal set satisfying $R_{\mathcal{F}}(\Pi, A)$ by contradiction. Assume 

\st \ml{AF61} $B \subset A$, and

\st \ml{AF611} $B \models R_{\mathcal{F}}(\Pi, A)$. 

\st We now prove \mr{AF8}. 

\st For any rule 

\st \ml{AF62} $r \in R_{\mathcal A}(\Pi, A)^A$, assume 

\st \ml{AF63} $B \models body(r)$. We will prove $B \models head(r)$ \mr{AF7}.

\st Let $r_1 \in R_{\mathcal A}(\Pi, A)$ be the rule from which $r$ is obtained, and $r_2 \in \Pi$ the rule from which $r_1$ is obtained. 

\st For any $(not~ l) \in body(r_2)$, 

\st \ml{AF64} $A \models not ~ l$ because of \mr{AF62}, implying

\st \ml{AF640} $B \models not ~ l$. 

\st For any positive literal $l \in body(r_2)$, we have $l \in body(r)$ and thus 

\st \ml{AF641} $B \models l$ because of \mr{AF63}. Therefore, \mr{AF61} implies 

\st \ml{AF65} $A \models l$. 

\st For any aggregate atom $agg \in body(r_2)$, 

\st \ml{AF66} $A \models agg$ because of $r_1 \in R_{\mathcal A}(\Pi, A)$. 

\st By definition of the $\mathcal{F}log$ reduct, \mr{AF64}, \mr{AF65} and \mr{AF66} imply  

\st \ml{AF67} $r_2 \in R_{\mathcal{F}}(\Pi, A)$. 

\st Consider any aggregate atom 
$(f\{X: p(X)\} \odot  n) \in body(r_2)$. We will prove \mr{AF696}.  




\st By $B \models body(r)$ \mr{AF63}, 

\st \ml{AF6921} $B \models \{p(X): p(X) \in A \}$, i.e., 

$ \{p(X): p(X) \in A\} \subseteq \{p(X): p(X) \in B\}$.

\st So, $\{X:  p(X) \in A\} \subseteq \{X: p(X) \in B\}$. 

\st By \mr{AF61}, 



\st \ml{AF6922}  $\{X: p(X) \in B\} \subseteq \{X: p(X) \in A\}$. 

\st \mr{AF6921} and \mr{AF6922} imply that 

\st \ml{AF6923}  $\{X: p(X) \in A\} = \{X: p(X) \in B\}$, which implies

\st \ml{AF695} $f\{X: p(X) \in B\} \odot n$ holds because $f\{X: p(X) \in A\} \odot n$ holds by \mr{AF66}. Hence,

\st \ml{AF696} $B \models f\{X: p(X)\} \odot n$, which, together with \mr{AF641} and \mr{AF640}, implies

\st \ml{AF697} $B \models body(r_2)$. Therefore, 

\st \ml{AF698} $B \cap head(r_2) \neq \emptyset$, by \mr{AF611} and \mr{AF67}.

\st Since $head(r_2) = head(r)$,  

\st \ml{AF7} $B \cap head(r) \neq \emptyset$. Therefore, $B \models r$ and thus

\st \ml{AF8} $B \models R_{\mathcal A}(\Pi, A)^A$.

\st Together with \mr{AF61}, \mr{AF8} contradicts that $A$ is an answer set of $R_{\mathcal A}(\Pi, A)^A$. 

\st Hence, by \mr{AF51},
$A$ is a minimal set satisfying  $R_{\mathcal{F}}(\Pi, A)$, i.e., $A$ is an $\mathcal{F}log$ answer set of $\Pi$.
 \hfill $\Box$

\st Given a program $P$ that is aggregate stratified with leveling mapping $||~||$ and a predicate $p$, $|| p ||$ is called the {\em level} of $p$,
$P_i = \{r \in P:$  the level of the head of $r$ is the $ i^{th}$  minimum in terms of $||~|| \}$ is called the {\em $i^{th}$ strata} of $P$ with respect to $||~||$, and $P_1, P_2, ....$ form a partition of $P$. 
The answer sets of a program are related to those of the stratas of the program in the following way. 

\st In the rest of the appendix, by a set of literals being a {\em model} of a program, we mean that every rule of the program is satisfied by the set. 

\setcounter{statement}{0}

\begin{lemma}[Answer Sets of a Program and Its Stratas] \label{lm:answerSetsStrata}
Given a program $P$ aggregate stratified with a level mapping, let $P_i$ be the $i^{th}$ strata with respect to the level mapping, $Ha_i$ be the atoms occurring in the head of $P_i$, and $\Pi_i = \cup_{j \le i} P_j$. For any set $A$ of ground regular atoms such that $A \subseteq \cup_{j\in 1..\infty}Ha_j$, let $A_i = \cup_{j \le i} (Ha_j \cap A)$. $A$ is an $\mathcal{A}log$ ($\mathcal{F}log$ and $\mathcal{S}log$ respectively) answer set of $P$, iff for any $i$, $A_i$ is an $\mathcal{A}log$ ($\mathcal{F}log$ and $\mathcal{S}log$ respectively) answer set of $\Pi_i$. 
\end{lemma}

\noindent Proof. 

\st For any $i$, by definition of $A_i$, we have 

\st \ml{sas:11} $A_i \subseteq A_{i+1}$, and 

\st \ml{sas:12} no atoms of $A_{i+1}-A_{i}$ occur in $\Pi_i$. 

\begin{enumerate}

\item[a.] $\Longrightarrow$: Assume $A$ is an $\mathcal{A}log$ answer set. We have 

\st \ml{sas:21} $A$ is a minimal model of $R_{\mathcal{A}}(P, A)^A$. 

\st \ml{sas:22}$R_{\mathcal{A}}(P, A)^A$
\begin{minipage}[t]{0.85\textwidth}
= $(\cup_{j=1}^{j=i} R_{\mathcal{A}}(P_j, A)^A) \cup (\cup_{j=i+1}^{j=\infty} R_{\mathcal{A}}(P_j, A)^A)$ \\
$~~~~$ (let the latter be denoted by {\em restPi}) \\
= $R_{\mathcal{A}}(\Pi_i, A)^A \cup$ {\em restPi}.  
\end{minipage}

\st For any $i$, we will show that $A_i$ is  an answer set of $\Pi_i$ \mr{sas:2511} by showing $A_i$ is a minimal model of $R_{\mathcal{A}}(\Pi_i, A_i)^{A_i}$
\mr{sas:2510}. 

\st Since $A$ (= $A_i \cup (A-A_i)$) is a model of $R_{\mathcal{A}}(P, A)^A$ and no atoms of $A - A_i$ occur in $\Pi_i$ (by the definition of $A_i$ 
), \mr{sas:22} implies

\st \ml{sas:221} $A_i$ is a model of $R_{\mathcal{A}}(\Pi_i, A)^A$. 

\st We next show $A_i$ is minimal \mr{sas:25} by contradiction. Assume that there exists $B$ such that  

\st \ml{sas:231} $B \subset A_i$, and 

\st \ml{sas:232} $B$ is a model of $R_{\mathcal{A}}(\Pi_i, A)^A$.

\st For any rule $r \in$ {\em restPi}, assuming 

\st \ml{sas:233} $B \cup (A-A_i) \models body(r)$,

\st we prove $B \cup (A - A_i) \models head(r)$ \mr{sas:236}.

\st Since $B \cup (A-A_i) \subseteq A$ and there are no negative atoms or aggregate atoms in $r$, \mr{sas:233} implies

\st \ml{sas:234} $A \models body(r)$. 

\st Since $A$ is a model of {\em restPi} by \mr{sas:21} and \mr{sas:22}, we have $A \models r$ and thus \mr{sas:234} implies 

\st \ml{sas:235} $A \models head(r)$. Since $head(r)$ does not occur in $\Pi_i$, $A-A_i \models head(r)$, and thus

\st \ml{sas:236}  $B \cup (A-A_i) \models head(r)$. Therefore, 

\st \ml{sas:237} $B \cup (A - A_i) \models$ {\em restPi}, which, together with \mr{sas:232} and \mr{sas:22}, implies

\st \ml{sas:2371} $B \cup (A-A_i) \models R_{\mathcal{A}}(P, A)^A$.

\st By $B \subset A_i$ \mr{sas:231} and $A_i \subseteq A$, we have $B \cup (A-A_i) \subset A$, which, together with \mr{sas:2371}, contradicts 
that $A$ is a minimal model of $R_{\mathcal{A}}(P, A)^A$. Hence,

\st  \ml{sas:25} $A_i$ is a minimal
model of $R_{\mathcal{A}}(\Pi_i, A)^A$.

\st Since no atoms $A-A_i$ occurs in $\Pi_i$, $R_{\mathcal{A}}(\Pi_i, A) = R_{\mathcal{A}}(\Pi_i, A_i)$, and $R_{\mathcal{A}}(\Pi_i, A_i)^A = R_{\mathcal{A}}(\Pi_i, A_i)^{A_i}$. Therefore, 

\st \ml{sas:251} $R_{\mathcal{A}}(\Pi_i, A)^A = R_{\mathcal{A}}(\Pi_i, A_i)^{A_i}$, which, together with \mr{sas:25}, implies

\st \ml{sas:2510} $A_i$ is a minimal model of $R_{\mathcal{A}}(\Pi_i, A_i)^{A_i}$, i.e., 

\st \ml{sas:2511} $A_i$ is an $\mathcal{A}log$ answer set of $\Pi_i$. 

\st $\Longleftarrow$: Assuming for any $i \ge 0$, 

\st \ml{sas:261} $A_i$ is an $\mathcal{A}log$ answer set of $\Pi_i$, 

\st we prove $A$ is an $\mathcal{A}log$ answer set of $P$ \mr{sas:26z}.

\st \ml{sas:2611} $lim_{i\rightarrow \infty} A_i$ 
\begin{minipage}[t]{0.85\textwidth}
= $lim_{i\rightarrow \infty} \cup_{j \in 1..i} (Ha_j \cap A)$\\
=  $lim_{i\rightarrow \infty} (A \cap (\cup_{j \in 1..i} Ha_j))$\\
= $A \cap (lim_{i\rightarrow \infty} \cup_{j \in 1..i} (Ha_j ))$ \\
= $A$ (because $A \subseteq lim_{i\rightarrow \infty} \cup_{j \in 1..i} (Ha_j )$)
\end{minipage}

\st By \mr{sas:261}, $lim_{i\rightarrow \infty} A_i$ is a minimal model of  $lim_{i\rightarrow \infty} R_{\mathcal{A}}(\Pi_i, A_i)^{ A_i}$ which is $R_{\mathcal{A}}(P, A)^A$. Therefore, $A$ is a minimal model of $R_{\mathcal{A}}(\Pi, A)^{A}$. So,

\st \ml{sas:26z} $A$ is an $\mathcal{A}log$ answer set of $P$.

\st 

\item[b.] $\Longrightarrow$: Assume $A$ is an $\mathcal{F}log$ answer set of $P$. We have 

\st \ml{sas:31} $A$ is a minimal model of $R_{\mathcal{F}}(P, A)$.

\st \ml{sas:32} $R_{\mathcal{F}}(P, A) = \cup_{j=1}^{j=i} R_{\mathcal{F}}(P_j, A)~ \cup$ {\em restPi}, where {\em restPi} = $\cup_{j=i+1}^{j=\infty} R_{\mathcal{F}}(P_j, A)$.

\st Since no atoms of $A - A_i$ occur in $\Pi_i$, $A = A_i \cup (A - A_i)$ and \mr{sas:31}, 

\st \ml{sas:331} $A_i$ is a model of  $\cup_{j=1}^{j=i} R_{\mathcal{F}}(P_j, A)$, i.e., $R_{\mathcal{F}}(\Pi_i, A)$. 

\st We prove that $A_i$ is minimal \mr{sas:35} by contradiction. Assume 

\st \ml{sas:341} $B \subset A_i$, and 

\st \ml{sas:342} $B$ is a model of  $R_{\mathcal{F}}(\Pi_i, A)$. 

\st For any $r \in$ {\em restPi}, assuming 

\st \ml{sas:343} $B \cup (A- A_i) \models body(r)$, we prove $B  \cup (A- A_i) \models head(r)$ \mr{sas:3a}.

\st Since $r \in$ {\em restPi}, $A \models body(r)$ (by $\mathcal{F}log$ reduct). By \mr{sas:31}, 

\st \ml{sas:351} $A \models head(r)$. Since $head(r)$ does not occur in $\Pi_i$, it implies  $A - A_i \models head(r)$, and thus 

\st \ml{sas:3a} $B  \cup (A- A_i) \models head(r)$. Hence, 

\st \ml{sas:3a1} $B \cup (A- A_i) \models$ {\em restPi}, which, together with \mr{sas:342} and \mr{sas:32}, implies 

\st $B \cup (A- A_i) \models R_{\mathcal{F}}(P, A)$ which, together with $B \cup (A- A_i) \subset A$, contradicts that $A$ is a minimal model of $R_{\mathcal{F}}(P, A)$ \mr{sas:31}. Hence,  

\st \ml{sas:35} $A_i$ is a minimal model of $R_{\mathcal{F}}(\Pi_i, A)$. 

\st Since no atoms of $A - A_i$ occurs in $\Pi_i$, $R_{\mathcal{F}}(\Pi_i, A) = R_{\mathcal{F}}(\Pi_i, A_i)$, which, together with \mr{sas:35}, implies 

\st $A_i$ is a minimal model of $R_{\mathcal{F}}(\Pi_i, A_i)$. Therefore, $A_i$ is an $\mathcal{F}log$ answer set of $\Pi_i$.  

\st $\Longleftarrow$: Assuming for any $i$, 

\st \ml{sas:361} $A_i$ is an $\mathcal{F}log$ answer set of $\Pi_i$, 

\st we prove $A$ is an $\mathcal{F}log$ answer set of $P$ \mr{sas:36z}.

\st By \mr{sas:2611}, $A = lim_{i\rightarrow \infty} A_i$. By \mr{sas:361}, the latter is a minimal model of $lim_{i\rightarrow \infty}R_{\mathcal{F}}(\Pi_i,  A_i)$  which is $R_{\mathcal{F}}(P, A)$. Therefore, $A$ is a minimal model of $R_{\mathcal{F}}(\Pi, A)$. So,

\st \ml{sas:36z} $A$ is an $\mathcal{F}log$ answer set of $P$. 

\st 

\item[c.] $\Longrightarrow$: Assume $A$ is an $\mathcal{S}log$ answer set, i.e.,
 
\st \ml{sas:11}$A = lfp(K_A^P).$

\st We first prove that for any $I$, $\Pi_i$ and a rule $r \in \Pi_i$, $(I, A_{i+1}) \models body(r)$ iff $(I, A_i) \models body(r)$ \mr{sas:12a}. 

\st For any regular atom $a \in body(r)$, by \mr{sas:11} and \mr{sas:12},

\st \ml{sas:120} $(I, A_{i+1}) \models a$ iff $(I, A_i) \models a$. 

\st For any aggregate atom $agg \in body(r)$, by \mr{sas:12} and that $agg$ contains predicates occurring only in $\Pi_{i}$, we have

\st \ml{sas:121} $Base(agg) \cap (A_{i+1} - A_i) = \emptyset$. Therefore, 

\st \ml{sas:122} $Base(agg) - A_{i+1} = Base(agg) - A_{i}$.

\st \mr{sas:121} 
implies that 

\st \ml{sas:123} $I \cap A_{i+1} \cap Base(agg) = I \cap A_i \cap Base(agg)$, which, together with \mr{sas:122}, implies 

\st \ml{sas:124} $(I, A_{i+1}) \models agg$ iff $(I, A_{i}) \models agg$.

\st By \mr{sas:120} and \mr{sas:124},

\st \ml{sas:12a} $(I, A_{i+1}) \models body(r)$ iff $(I, A_i) \models body(r)$, which implies 

\st \ml{sas:12b} $(I, A) \models body(r)$ iff $(I, A_i) \models body(r)$.

\st We next prove that for any $\Pi_i$ and a rule $r \in \Pi_i$, \mr{sas:12ba} holds. 

\st For any regular atome $a \in body(r)$, by \mr{sas:12} and \mr{sas:11}, 

\st \ml{sas:12b1} $(A_{i+1}, A_i) \models a$ iff $(A_{i}, A_i) \models a$.

\st For any aggregate atom $agg \in body(r)$, since $A_{i+1} \cap A_i \cap Base(agg)$ = $A_{i} \cap A_i \cap Base(agg)$,

\st \ml{sas:12b2} $(A_{i+1}, A_i) \models agg$ iff $(A_{i}, A_i) \models agg$, which, together with \mr{sas:12b1}, implies

\st \ml{sas:12ba} $(A_{i+1}, A_i) \models body(r)$ iff $(A_i, A_i) \models body(r)$, which implies 

\st \ml{sas:12ba1} $(A, A_i) \models body(r)$ iff $(A_i, A_i) \models body(r)$

\st For any set $I$ of regular atoms,

\st \ml{sas:13} $K_{A_i}^{\Pi_i}(I)$ \begin{minipage}[t]{0.85\textwidth}
  = $\{head(r): r \in~ ^{A_i}\Pi_i, (I, A_i) \models body(r)\}$ \\
  
  = $\underline{\cup_{j=1}^{j=i}} \{head(r): r \in~ ^{A_i}\underline{P_j}, (I, A_i) \models body(r) \}$ since $P_1$ to $P_i$ are a partition of $\Pi_i$ \\
  
  = $\cup_{j=1}^{j=i} \{head(r): r \in~ ^{\underline{A_j}}\!P_j, (I, A_i) \models body(r) \}$ since no atoms of $A_i - A_j$ occur in $P_j$\\
  
\end{minipage}
(Note, underlines in the above highlight the difference between the current line and the previous.)

\st For any $i \ge 0$, 

\st \ml{sas:14} $A$ 
 \begin{minipage}[t]{0.85\textwidth}
  = $A_i \cup restA$, where $restA = A- A_i$\\
  
  = $K_{A}^{P}(A)$ since $A$ is an $\mathcal{S}log$ answer set of $P$ \\
  
  = $lim_{i \rightarrow \infty} K_{A_i}^{\Pi_i}(A)$ (by \mr{sas:2611})\\
  
  = $\cup_{j=1}^{j=
  \infty} \{head(r): r \in~ ^{A_j}\!P_j, (A, A_j) \models body(r) \}$ (by \mr{sas:13}), \\
  
  = $(\cup_{j=1}^{j=i} \{head(r): r \in~ ^{A_j}\!P_j, (A, A_j) \models body(r) \})$ (denoted by {\em HA}$_i$)
   \\
  $~~~~\cup (\cup_{j=i+1}^{j=\infty} \{head(r): r \in~ ^{A_j}\!P_j, (A, A_j) \models body(r) \})$ (denoted by {\em restHA}$_i$).\\
\end{minipage}

\st By \mr{sas:12}, no atoms in $A - A_i$, i.e., in $restA$, occur in $\Pi_i$. Therefore, 

\st \ml{sas:15} no atoms of $restA$ occur in {\em HA}$_i$. 

\st By definition of $A_i$ and strata $P_j (j > i)$, 

\st \ml{sas:16} $A_i \cap$  {\em restHA}$_j$ = $\emptyset$.

\st Therefore, we have 

\st \ml{sas:17} $A_i$ 
 \begin{minipage}[t]{0.85\textwidth}
  =  {\em HA}$_i$ by \mr{sas:15}, \mr{sas:16} and \mr{sas:14}.\\
  
  = $\cup_{j=1}^{j=i} \{head(r): r \in~ ^{\underline{A_i}}\!P_j, (A, A_j) \models body(r) \}$ \\
  
  = $(\cup_{j=1}^{j=i} \{head(r): r \in~ ^{A_i}P_j, (A, \underline{A_i}) \models body(r) \})$ by \mr{sas:12a}.
  
  = $(\cup_{j=1}^{j=i} \{head(r): r \in~ ^{A_i}P_j, (\underline{A_i}, A_i) \models body(r) \})$ by \mr{sas:12ba1}.

  = $K_{A_i}^{\Pi_i}(A_i)$.\\
\end{minipage}

\st Therefore, 

\st \ml{sas:171} $K_{A_i}^{\Pi_i} (A_i) = A_i$. 

\st We next prove, by contradiction, that $A_i$ is the least fixed point of $K_{A_i}^{\Pi_i}$. Assume 

\st \ml{sas:172}$B \subset A_i$, and 

\st \ml{sas:173} $B$ is a fixed point of $K_{A_i}^{\Pi_i}$. 

\st \ml{sas:180} $K_{A}^{P}(B \cup (A - A_i))$
 \begin{minipage}[t]{0.7\textwidth}
 
  = $(\cup_{j=1}^{j=i} \{head(r): r \in~ ^{A}\!P_j, (B \cup (A-A_i), A_j) \models body(r) \})$ (denoted by {\em HA$^\prime$}$_i$)
   \\
  $~~~~\cup (\cup_{j=i+1}^{j=\infty} \{head(r): r \in~ ^{A}\!P_j, (B \cup (A - A_i), A_j) \models body(r) \})$ (denoted by {\em restHA$^\prime$}$_i$).\\
\end{minipage}

\st \ml{sas:181} {\em HA$^\prime$}$_i$
 \begin{minipage}[t]{0.85\textwidth}
 
  = $(\cup_{j=1}^{j=i} \{head(r): r \in~ ^{A_i}P_j, (B, A_i) \models body(r) \})$ because no atoms of $(A-A_i)$ occur in $\Pi_i$ and no atoms of $(A_i - A_j)$ occurs in $\Pi_j$.
   \\
  = $K_{A_i}^{\Pi_i}(B)$ \\
  = $B$, by assumption \mr{sas:173}. 
\end{minipage}

\st \ml{sas:182} {\em restHA$^\prime$}$_i$
 \begin{minipage}[t]{0.85\textwidth}
 
  = $(\cup_{j=i+1}^{j=\infty} \{head(r): r \in~ ^{A}\!P_j, (B \cup (A - A_i), A_j) \models body(r) \})$\\
  $\subseteq (\cup_{j=i+1}^{j=\infty} \{head(r): r \in~ ^{A}\!P_j, (A, A_j) \models body(r) \})$ since $B \cup (A - A_i) \subseteq A$ (by \mr{sas:172}). \\
  = $A - A_i$ by by \mr{sas:15}, \mr{sas:16} and \mr{sas:14}.
  
\end{minipage}

\st By \mr{sas:181}, \mr{sas:182} and \mr{sas:180}, 

$K_{A}^{P}(B \cup (A - A_i))$
 \begin{minipage}[t]{0.85\textwidth}
 
  = $HA^\prime_i + restHA^\prime$ $\subseteq B \cup (A-A_i)$ \\
   $ \subset A_i \cup (A-A_i)$ = $A$,
\end{minipage}

contradicting $A$ = $lfp(K_A^P)$ \mr{sas:11}. Therefore,

\st \ml{sas:19} $A_i$ is the least fixed point of $K_{A_i}^{\Pi_i}$. 

\st $\Longleftarrow$: assuming for any $i$, 

\st \ml{sas:1a} $A_i$ is an $\mathcal{S}log$ answer set of $\Pi_i$, 

\st we prove $A$ is an $\mathcal{S}log$ answer set of $P$ \mr{sas:1z}.

\st By \mr{sas:2611}, $A = lim_{i\rightarrow \infty} A_i$. By \mr{sas:1a}, $lim_{i\rightarrow \infty} A_i$ is the least fixed point of $lim_{i\rightarrow \infty}K_{A_i}^{\Pi_i}$ which is $K_{A}^P$. Therefore,  $A$ is $lfp(K_A^P)$. So,

\st \ml{sas:1z} $A$ is an $\mathcal{S}log$ answer set of $P$.
\hfill $\Box$ 
\end{enumerate}

\hide{
\begin{lemma} \label{lm:answerSetsStrata}
Given a stratified aggregate program $P$ and a stratification of $P$, let $P_0, P_1, ..., P_i, ...$, where $P_i$ is the $i^{th}$ strata, be a partition of $P$,$Ha_i$ be the atoms occurring in the head of $P_i$, $A_i = \cup_{j \le i} (Ha_j \cap A)$, and $\Pi_i = \cup_{j \le i} P_j$. If $A$ is an $\mathcal{A}log$ ($\mathcal{F}log$ and $\mathcal{S}log$ respectively) answer set of $P$, then for any $i$, $A_i$ is an $\mathcal{A}log$ ($\mathcal{F}log$ and $\mathcal{S}log$ respectively) answer set of $\Pi_i$. 
\end{lemma}

\noindent Proof. 

\st By definition of $A_i$, we have 

\st \ml{sas:11} $A_i \subseteq A_{i+1}$, and 

\st \ml{sas:12} no atoms of $A_{i+1}-A_{i}$ occur in $\Pi_i$. 

\st a. Assume $A$ is an $\mathcal{A}log$ answer set. We have 

\st \ml{sas:21} $A$ is a minimal model of $R_{\mathcal{A}}(P, A)^A$. 

\st \ml{sas:22}$R_{\mathcal{A}}(P, A)^A$
\begin{minipage}[t]{0.85\textwidth}
= $(\cup_{j=1}^{j=i} R_{\mathcal{A}}(P_j, A)^A) \cup (\cup_{j=i}^{j=\infty} R_{\mathcal{A}}(P_j, A)^A)$ (denoted by {\em restPi}) \\
= $R_{\mathcal{A}}(\Pi_i, A)^A \cup$ {\em restPi}.  
\end{minipage}

\st Since $A$ (= $A_i \cup (A-A_i)$) is a model of $R_{\mathcal{A}}(P, A)^A$ and no atoms of $A - A_i$ occur in $\Pi_i$ (by \mr{sas:12}), \mr{sas:22} implies

\st \ml{sas:221} $A_i$ is a model of $R_{\mathcal{A}}(\Pi_i, A)^A$. 

\st We next show $A_i$ is minimal \mr{sas:25} by contradiction. Assume that 

\st \ml{sas:231} $B \subset A_i$, and 

\st \ml{sas:232} $B$ is a model of $R_{\mathcal{A}}(\Pi_i, A)^A$.

\st For any rule $r \in$ {\em restPi}, assuming 

\st \ml{sas:233} $B \cup (A-A_i) \models body(r)$ 

\st Since $B \cup (A-A_i) \subseteq A$, \mr{sas:233} implies

\st \ml{sas:234} $A \models body(r)$. 

\st Since $A$ is a model of {\em restPi} by \mr{sas:21}, $A \models r$ and thus \mr{sas:234} implies 

\st \ml{sas:235} $A \models head(r)$. Since $head(r)$ does not occur in $\Pi_i$, $A-A_i \models head(r)$, and thus

\st \ml{sas:236}  $B \cup (A-A_i) \models head(r)$. Therefore, 

\st \ml{sas:237} $A \models$ {\em restPi}, which, together with \mr{sas:232} and \mr{sas:22}, implies

$B \cup (A-A_i) \models R_{\mathcal{A}}(P, A)^A$, contradicting that $B \cup (A-A_i)$ and $A$ is a minimal model of $R_{\mathcal{A}}(P, A)^A$. Hence,

 \st \ml{sas:25} $A_i$ is a minimal
model of $R_{\mathcal{A}}(\Pi_i, A)^A$.

\st Since no atoms $A-A_i$ occurs in $\Pi$, $R_{\mathcal{A}}(\Pi_i, A) = R_{\mathcal{A}}(\Pi_i, A_i)$, and $R_{\mathcal{A}}(\Pi_i, A_i)^A = R_{\mathcal{A}}(\Pi_i, A_i)^{A_i}$. Therefore, 

\st \ml{sas:251} $R_{\mathcal{A}}(\Pi_i, A)^A = R_{\mathcal{A}}(\Pi_i, A_i)^{A_i}$, which, together with \mr{sas:25}, implies

$A_i$ is a minimal model of $R_{\mathcal{A}}(\Pi_i, A_i)^{A_i}$, which implies 

$A_i$ is an $\mathcal{A}log$ answer set of $\Pi_i$. 

\st b. Assume $A$ is an $\mathcal{F}log$ answer set. We have 

\st \ml{sas:31} $A$ is a minimal model of $R_{\mathcal{F}}(P, A)$.

\st \ml{sas:32} $R_{\mathcal{F}}(P, A) = \cup_{j=1}^{j=i} R_{\mathcal{F}}(P_j, A)$ + {\em restPi}, where {\em restPi} = $\cup_{j=i+1}^{j=\infty} R_{\mathcal{F}}(P_j, A)$.

\st Since no atoms of $A - A_i$ occur in $\Pi_i$, $A = A_i \cup (A - A_i)$ and \mr{sas:31}, 

\st \ml{sas:331} $A$ is a model of  $\cup_{j=1}^{j=i} R_{\mathcal{F}}(P_j, A)$, i.e., $R_{\mathcal{F}}(\Pi_i, A)$. 

\st We prove that $A_i$ is minimal \mr{sas:35} by contradiction. Assume 

\st \ml{sas:341} $B \subset A_i$, and 

\st \ml{sas:342} $B$ is a model of  $R_{\mathcal{F}}(\Pi_i, A)$. 

\st For any $r \in$ {\em restPi}, assuming 

\st \ml{sas:343} $B \cup (A- A_i) \models body(r)$, we prove $B  \cup (A- A_i) \models head(r)$ \mr{sas:3a}.

\st Since $r \in$ {\em restPi}, $A \models body(r)$. By \mr{sas:31}, 

\st \ml{sas:351} $A \models head(r)$. Since $head(r)$ does not occur in $\Pi_i$, it implies  $A - A_i \models head(r)$, and thus 

\st \ml{sas:3a} $B  \cup (A- A_i) \models head(r)$. Hence, 

\st \ml{sas:3a1} $B \cup (A- A_i) \models$ {\em restPi}, which, together with \mr{sas:342} and \mr{sas:32}, implies 

$B \cup (A- A_i) \models R_{\mathcal{F}}(P, A)$ which, together with $B \cup (A- A_i) \subset A$, contradicts that $A$ is minimal \mr{sas:31}. Hence,  

\st \ml{sas:35} $A_i$ is a minimal model of $R_{\mathcal{F}}(\Pi_i, A)$. 

\st Since no atoms of $A - A_i$ occurs in $\Pi_i$, $R_{\mathcal{F}}(\Pi_i, A) = R_{\mathcal{F}}(\Pi_i, A_i)$, which, together with \mr{sas:35}, implies 

$A_i$ is a minimal model of $R_{\mathcal{F}}(\Pi_i, A_i)$. Therefore, $A_i$ is an $\mathcal{F}log$ answer set of $\Pi_i$.  

\st c. Assume $A$ is an $\mathcal{S}log$ answer set, i.e.,
 
$A = lfp(K_A^P)$

\st We first prove that for any $\Pi_i$ and a rule $r \in \Pi_i$, $(I, A_{i+1}) \models body(r)$ iff $(I, A_i) \models body(r)$ \mr{sas:12a}. 

\st For any regular atoms $a \in body(r)$, by \mr{sas:11} and \mr{sas:12},

\st \ml{sas:120} $(I, A_{i+1}) \models a$ iff $(I, A_i) \models a$. 

\st For any aggregate atom $agg \in body(r)$, by \mr{sas:12} and that $agg$ contains predicates occurring only in $\Pi_{i}$. 

\st \ml{sas:121} $Base(agg) \cap (A_{i+1} - A_i) = \emptyset$. Therefore, 

\st \ml{sas:122} $Base(agg) - A_{i+1} = Base(agg) - A_{i}$.

\st \mr{sas:121} and \mr{sas:11} imply that 

\st \ml{sas:123} $I \cap A_{i+1} \cap Base(agg) = I \cap A_i \cap Base(agg)$, which, together with \mr{sas:122}, implies 

\st \ml{sas:124} $(I, A_{i+1}) \models agg$ iff $(I, A_{i}) \models agg$.

\st By \mr{sas:120} and \mr{sas:124},

\st \ml{sas:12a} $(I, A_{i+1}) \models body(r)$ iff $(I, A_i) \models body(r)$.

\st \mr{sas:12a} implies 

\st \ml{sas:12b} $(I, A) \models body(r)$ iff $(I, A_i) \models body(r)$.

\st We next prove that for any $\Pi_i$ and a rule $r \in \Pi_i$, \mr{sas:12ba} holds. 

\st For any regular atome $a \in body(r)$, by \mr{sas:12} and \mr{sas:11}, 

\st \ml{sas:12b1} $(A_{i+1}, A_i) \models a$ iff $(A_{i}, A_i) \models a$.

\st For any aggregate atom $agg \in body(r)$, since $A_{i+1} \cap A_i \cap Base(agg)$ = $A_{i} \cap A_i \cap Base(agg)$,

\st \ml{sas:12b2} $(A_{i+1}, A_i) \models agg$ iff $(A_{i}, A_i) \models agg$, which, together with \mr{sas:12b1}, implies

\st \ml{sas:12ba} $(A_{i+1}, A_i) \models body(r)$ iff $(A_i, A_i) \models body(r)$, which implies 

\st \ml{sas:12ba1} $(A, A_i) \models body(r)$ iff $(A_i, A_i) \models body(r)$

\st For any set $I$ of regular atoms,

\st \ml{sas:13} $K_{A_i}^{\Pi_i}(I)$ \begin{minipage}[t]{0.85\textwidth}
  = $\{head(r): r \in~ ^{A_i}\Pi_i, (I, A) \models body(r)\}$ \\
  
  = $\underline{\cup_{j=1}^{j=i}} \{head(r): r \in~ ^{A_i}\underline{P_j}, (I, A) \models body(r) \}$ since $P_0$ to $P_i$ are a partition of $\Pi_i$ \\
  
  = $\cup_{j=1}^{j=i} \{head(r): r \in~ ^{\underline{A_j}}\!P_j, (I, A) \models body(r) \}$ since no atoms of $A_i - A_j$ occur in $P_j$ (by \mr{sas:11} and \mr{sas:12})\\
  
  = $\cup_{j=1}^{j=i} \{head(r): r \in~ ^{A_j}\!P_j, (I,  \underline{A_i}) \models body(r)\}$ by \mr{sas:12b}.\\
\end{minipage}
(Note, underlines in the above highlight the difference between the current line and the previous.)

\st For any $i$, 

\st \ml{sas:14} $A$ 
 \begin{minipage}[t]{0.85\textwidth}
  = $A_i \cup restA_{i+1}$, where $restA = A- A_i$,\\
  
  = $K_{A}^{P}(A)$ since $A$ is an $\mathcal{S}log$ answer set of $P$ \\
  
  = $lim_{i \rightarrow \infty} K_{A_i}^{\Pi_i}(A)$ \\
  
  = $\cup_{j=1}^{j=
  \infty} \{head(r): r \in~ ^{A_j}\!P_j, (A, A_j) \models body(r) \}$ (by \mr{sas:13}), \\
  
  = $(\cup_{j=1}^{j=i} \{head(r): r \in~ ^{A_j}\!P_j, (A, A_j) \models body(r) \})$ (denoted by {\em HA}$_i$)
   \\
  $~~~~\cup (\cup_{j=i+1}^{j=\infty} \{head(r): r \in~ ^{A_j}\!P_j, (A, A_j) \models body(r) \})$ (denoted by {\em restHA}$_i$).\\
\end{minipage}

\st By \mr{sas:12}, no atoms in $A - A_n$, and thus in $restA$, occurs in $\Pi_i$. Therefore, 

\st \ml{sas:15} no atoms of $restA$ occur in {\em HA}$_i$. 

\st By definition of $A_i$ and strata $P_j (j \ge i)$, 

\st \ml{sas:16} $A_i \cap$  {\em restHA}$_i$ = $\emptyset$.

\st Therefore, we have 

\st \ml{sas:17} $A_i$ 
 \begin{minipage}[t]{0.85\textwidth}
  =  {\em HA}$_i$, by \mr{sas:15}, \mr{sas:16} and \mr{sas:14}.\\
  
  = $\cup_{j=1}^{j=i} \{head(r): r \in~ ^{\underline{A_i}}\!P_j, (A, A_j) \models body(r) \}$ by \mr{sas:13}. \\
  
  = $(\cup_{j=1}^{j=i} \{head(r): r \in~ ^{A_i}P_j, (A, \underline{A_i}) \models body(r) \})$, by \mr{sas:12a}.
  
  = $(\cup_{j=1}^{j=i} \{head(r): r \in~ ^{A_i}P_j, (\underline{A_i}, A_i) \models body(r) \})$, by \mr{sas:12ba1}.

  = $K_{A_i}^{\Pi_i}(A_i)$.\\
\end{minipage}

\st Therefore, 

\st \ml{sas:171} $K_{A_i}^{\Pi_i} (A_i) = A_i$. 

\st We next Prove, by contradiction, that $A_i$ is the least fixed point of $K_{A_i}^{\Pi_i}$. Assume 

\st \ml{sas:172}$B \subset A_i$, and 

\st \ml{sas:173} $B$ is a fixed point of $K_{A_i}^{\Pi_i}$. 

\st \ml{sas:180} $K_{A}^{P}(B \cup (A - A_i))$
 \begin{minipage}[t]{0.7\textwidth}
 
  = $(\cup_{j=1}^{j=i} \{head(r): r \in~ ^{A}\!P_j, (B \cup (A-A_i), A_j) \models body(r) \})$ (denoted by {\em HA}$_i$)
   \\
  $~~~~\cup (\cup_{j=i+1}^{j=\infty} \{head(r): r \in~ ^{A}\!P_j, (B \cup (A - A_i), A_j) \models body(r) \})$ (denoted by {\em restHA}$_i$).\\
\end{minipage}

\st \ml{sas:181} {\em HA}$_i$
 \begin{minipage}[t]{0.85\textwidth}
 
  = $(\cup_{j=1}^{j=i} \{head(r): r \in~ ^{A_i}P_j, (B, A_i) \models body(r) \})$ because no atoms of $(A-A_i)$ occur in $\Pi_i$ and no atoms of $(A_i - A_j)$ occurs in $\Pi_j$.
   \\
  = $K_{A_i}^{\Pi_i}(B)$ \\
  = $B$, by assumption \mr{sas:173}. 
\end{minipage}

\st \ml{sas:182} {\em restHA}$_i$
 \begin{minipage}[t]{0.85\textwidth}
 
  = $(\cup_{j=i+1}^{j=\infty} \{head(r): r \in~ ^{A}\!P_j, (B \cup (A - A_i), A_j) \models body(r) \})$\\
  = $(\cup_{j=i+1}^{j=\infty} \{head(r): r \in~ ^{A}\!P_j, (B \cup (A - A_i), A) \models body(r) \})$ \\
  $\subseteq (\cup_{j=i+1}^{j=\infty} \{head(r): r \in~ ^{A}\!P_j, (A, A) \models body(r) \})$ since $B \cup (A - A_i) \subseteq A$ (by \mr{sas:172}). \\
  = $A - A_i$ because of reasons similar to those leading to \mr{sas:17}.
  
\end{minipage}

\st By \mr{sas:181}, \mr{sas:182} and \mr{sas:180}, $K_{A}^{P}(B \cup (A - A_i)) = B \cup (A-A_i) \subset A_i \cup (A-A_i)$ = $A$, contradicting $A$ = $lfp(K_A^P)$. Therefore,

\st \ml{sas:19} $A_i$ is the least fixed point of $K_{A_i}^{\Pi_i}$.

\hfill $\Box$ 
}

\st In the proofs for the following proposition and Proposition~\ref{prop:stratA2S}, we use the following notations without redefining them.  $Ha_i$ denotes the atoms occurring in the head of $i^{th}$ strata $P_i$, and $\Pi_i = \cup_{j \le i} P_j$. For any set $A$ of ground regular atoms, $A_i$ denotes $\cup_{j \le i} (Ha_j \cap A)$. 

\setcounter{statement}{0}

\begin{claim}
[Equivalence of $\mathcal{A}log$ and
   $\mathcal{F}log$ Semantics for Aggregate Stratified Programs]\label{prop:AlogEqFlog}
If $\Pi$ is  an aggregate stratified $\mathcal{AF}$-compatible program then
$A$ is an $\mathcal{A}log$ answer set of $\Pi$ iff it is an $\mathcal{F}log$ answer set of $\Pi$.    
\end{claim}

\noindent Proof. 

\st $\Longrightarrow$: Assuming 

\st \ml{sfa1} $A$ is an $\mathcal{F}log$ answer set of $P$,

\st we prove $A$ is an $\mathcal{A}log$ answer set of $P$ \mr{sfa99}. 

\st By \mr{sfa1} and Lemma~\ref{lm:answerSetsStrata},  

\st \ml{sfa11} for any $i$, $A_i$ is an $\mathcal{F}log$ answer set of $\Pi_i$. 

\st To prove \mr{sfa99}, for any $i(\ge 1)$, we prove $A_i$ is an $\mathcal{A}log$ answer set of $\Pi_i$ by induction on $i$.

\st Base case: $i=1$. $A_i$ is an $\mathcal{A}log$ answer set of $\Pi_i$ because $\Pi_i$ contains no aggregate e-atoms. 

\st Inductive hypothesis: for any number $n > 1$, we assume 

\st \ml{sfa12} for any $k$ such that $1 \le k < n$, $A_k$ is an $\mathcal{A}log$ answer set of $\Pi_k$. 

\st We will prove that $A_n$ is an $\mathcal{A}log$ answer set of $\Pi_n$ \mr{sfa91}. 

\st We first prove $A_n$ is a model of $R_{\mathcal{A}}(\Pi_n, A_n)^{A_n}$ \mr{sfa8}.

\st For any rule $r'' \in R_{\mathcal{A}}(\Pi_n, A_n)^{A_n}$,  assuming 

\st \ml{sfa2} $A_n \models body(r'')$, 

\noindent we prove $A_n \models head(r'')$ \mr{sfa372}. 

\st Since $r'' \in R_{\mathcal{A}}(\Pi_n, A_n)^{A_n}$, there exists a rule $r \in P$ from which $r''$ is obtained after the aggregate reduct and the classical reduct. Let $r$ be of the form 

\st \ml{sfa31} $head(r)$ :- $posReg(r)$, $negReg(r)$, $aggs(r)$.  

\st Since $r'' \in R_{\mathcal{A}}(\Pi_n, A_n)^{A_n}$. 

\st \ml{sfa32} $A_n \models aggs(r)$, and 

\st \ml{sfa33} $A_n \models negReg(r)$. 

\st By \mr{sfa31}, the form of $r''$ is 

\st \ml{sfa34} $head(r)$ :- $posReg(r)$, $\cup_{agg \in aggs(r)}ta(agg, A_n)$.

\st By \mr{sfa2}, 

\st \ml{sfa35} $A_n \models posReg(r)$, which, together with \mr{sfa33} and \mr{sfa32}, implies the existence of rule $r' \in R_{\mathcal{F}}(\Pi_n, A_n)$ which is of the same form as $r$:

\st \ml{sfa36} $head(r)$ :- $posReg(r)$, $negReg(r)$, $aggs(r)$, and 

\st \ml{sfa37} $A_n \models body(r')$, which, together with \mr{sfa36} and $A_n \models r$ (by \mr{sfa1}), i.e., $A_n \models r'$, implies 

\st \ml{sfa371} $A_n \models head(r')$. 

\st By \mr{sfa34} and \mr{sfa36}, $head(r') = head(r'')$. So, \mr{sfa371} implies 

\st \ml{sfa372} $A_n \models head(r'')$, which implies 

\st \ml{sfa7z} $A_n \models r''$. Therefore, 

\st \ml{sfa8}$A_n$ is a model of $R_{\mathcal{A}}(\Pi_n, A_n)^{A_n}$.

\st We next show $A_n$ is minimal \mr{sfa8z} by contradiction. Assume there exists $B$ such that 

\st \ml{sfa811} $B \subset A_n$, and 

\st \ml{sfa812} $B$ is a model of $R_{\mathcal{A}}(\Pi_n, A_n)^{A_n}$.

\st We note   


\st \ml{sfa7931} $A_n - Ha_n = A_{n-1}$ by the definition of $A_n$.

\st Since $A_n$ is an $\mathcal{F}log$ answer set of $\Pi_n$ \mr{sfa11}, \mr{sfa811} implies that there is some rule $r$ of $R_{\mathcal{F}}(\Pi_n, A_n)$ which is not satisfied by $B$, i.e.,

\st \ml{sfa821} $B \models body(r)$, and 

\st \ml{sfa822} $B \not \models head(r)$. 

\st Since  $r \in R_{\mathcal{F}}(\Pi_n, A_n)$, $r \in \Pi_n$. 
Let $r$ be of the form: 

\st \ml{sfa823} $head(r)$ :- $posReg(r)$, $negReg(r)$, $aggs(r)$. 

\st Since $r \in R_{\mathcal{F}}(\Pi_n, A_n)$, 

\st \ml{sfa831} $A_n \models posReg(r)$, 

\st \ml{sfa832} $A_n \models negReg(r)$, and

\st \ml{sfa833} $A_n \models aggs(r)$.

\st \mr{sfa831} to \mr{sfa833} imply that there is a rule $r'' \in R_{\mathcal{A}}(\Pi_n, A_n)^{A_n}$ which is obtained from $r$. Rule $r''$ is of the form:

\st \ml{sfa835} $head(r)$ :- $posReg(r)$, $\cup_{agg \in aggs(r)} ta(agg, A_n)$. 

\st We now prove an intermediate result $A_{n-1} = B - Ha_n$ \mr{sfa84242}.

\st Since $B \subset A_n$ \mr{sfa811} and $A_n - Ha_n = A_{n-1}$ \mr{sfa7931},

\st \ml{sfa8421} $B - Ha_n \subseteq A_{n-1}$.

\st By definition of $\Pi_{n}$ and 
that $B$ is a model of $R_{\mathcal{A}}(\Pi_{n}, A_n)^{A_n}$ \mr{sfa812}, 

\st \ml{sfa8422} $B$ is a model of $R_{\mathcal{A}}(\Pi_{n-1}, A_n)^{A_n}$.

\st Since atoms of $Ha_n$ do not occur in $\Pi_{n-1}$, \mr{sfa8422}

\st \ml{sfa8423} $B-Ha_n$ is a model of $R_{\mathcal{A}}(\Pi_{n-1}, A_n)^{A_n}$ = $R_{\mathcal{A}}(\Pi_{n-1}, A_{n-1})^{A_{n-1}}$ (because no atoms of $A_n - A_{n-1}$ occur in $\Pi_{n-1}$).

\st By induction hypothesis, $A_{n-1}$ is an $\mathcal{A}log$ answer set of $\Pi_{n-1}$. Therefore, 

\st \ml{sfa8424} $A_{n-1}$ is a minimal model of $R_{\mathcal{A}}(\Pi_{n-1}, A_{n-1})^{A_{n-1}}$, which, together with \mr{sfa8423}, implies

\st \ml{sfa84241} $(B-Ha_n) \not  \subset A_{n-1}$, which together with \mr{sfa8421}, implies 

\st \ml{sfa84242} $A_{n-1} = B - Ha_n$. 

\hide{ 
\st We now prove \mr{sfa8425} by proving \mr{sfa8425}. For any $r \in R_{\mathcal{F}}(\Pi_{n-1}, A_{n-1})$, assume 

\st \ml{sfa84231} $B - Ha_n \models body(r)$. 

\st Since $r \in R_{\mathcal{F}}(\Pi_{i-1}, A_{i-1})$, $r \in \Pi_{i-1}$ and there is $r'' \in R_{\mathcal{A}}(\Pi_{i-1}, A_{i-1})^{A_{i-1}}$ such that $r''$ is obtained from $r$. Let $r$ be of the form:

\st \ml{sfa84232} $head(r)$ :- $posReg(r), negReg(r), aggs(r)$. 

\st \mr{sfa84231} implies 

\st \ml{sfa84233} $B-Ha_n \models posReg(r)$,

\st \ml{sfa84234} $B - Ha_n \models negReg(r)$, and 

\st \ml{sfa84235} $B - Ha_n \models aggs(r)$. 

\st \mr{sfa84235} implies

......

\st $B - Ha_n \models ta(agg, A_{i-1})$. 

$A_{n-1}$ is not a minimal model of $R_{\mathcal{A}}(\Pi_{n-1}, A)^A$, contradicting that $A_{i-1}$ is a minimal model \mr{sfa791}. Therefore, 

\st \ml{sfa8424} $B - Ha_n \models R_{\mathcal{F}}(\Pi_{i-1}, A_{i-1})$.

\st \ml{sfa8425} $A_{n-1} \subseteq B-Ha_n$, which together with \mr{sfa8421}, implies 

\st \ml{sfa8426} $A_{n-1} = B-Ha_n$. 
}

\st We next prove $B \models \cup_{agg \in aggs(r)} ta(agg, A_n)$ \mr{sfa861}.



\st Since $r \in \Pi_{n}$, by definition of $Ha_n$,  for any $agg \in aggs(r)$, we have

\st \ml{sfa851} $Base(agg) \cap Ha_n = \emptyset$.  Therefore,

\st \ml{sfa852} $ta(agg, A_n)$ 
  \begin{minipage}[t]{0.85\textwidth}
	$= ta(agg, A_n-Ha_n)$ \\
	$= ta(agg, A_{n-1})$ because $A_n - Ha_n = A_{n-1}$ \mr{sfa7931} \\
	$= ta(agg, B - Ha_n)$ because $A_{n-1} = B - Ha_n$ \mr{sfa84242} \\
	$= ta(agg, B)$ by \mr{sfa851}.
  \end{minipage}

\st Hence, 

\st \ml{sfa854} $ta(agg, A_n) = ta(agg, B)$. Since $B \models ta(agg, B)$, we have  

\st \ml{sfa86} $B \models ta(agg, A_n)$. Hence, 

\st \ml{sfa861} $B \models \cup_{agg \in aggs(r)} ta(agg, A_n)$. 

\st By $B \models body(r)$ \mr{sfa821}, 

\st \ml{sfa862} $B \models posReg(r)$. 

\st \mr{sfa862} and \mr{sfa861} imply the body of rule $r''$ \mr{sfa835} is satisfied.  Since $B$ is a model of $R_{\mathcal{A}}(\Pi_n, A_n)^{A_n}$ \mr{sfa812}, $B \models r''$. Therefore, $B \models head(r'')$, i.e., 

\st \ml{sfa8x} $B \models head(r)$ because $head(r'')=head(r)$, which contradicts $B \not \models head(r)$ \mr{sfa822}. Hence,

\st \ml{sfa8z} $A_n$ is a minimal model of $R_{\mathcal{A}}(\Pi_n, A_n)^{A_n}$. Therefore, 

\st \ml{sfa91} $A_n$ is an $\mathcal{A}log$ answer set of $\Pi_n$. So, 

\st \ml{sfa92} For any $i \ge 1$, $A_i$ is an $\mathcal{A}log$ answer set of $\Pi_i$. 

\st By Lemma~\ref{lm:answerSetsStrata}, \mr{sfa92} implies 

\st \ml{sfa99} $A$ is an $\mathcal{A}log$ answer set of $P$. 

\st $\Longleftarrow$: this is implied by Proposition~\ref{prop:a2flog}. 
\hfill $\Box$

\subsubsection{Comparison with $\mathcal{S}log$} \label{app:slog}
Some notations. Consider a set $S$ of ground regular literals and an aggregate atom $agg$. $Base(agg)$ denotes the set of the ground instantiations of atoms occurring in the set name of $agg$. 
We define $ta(agg, S) = \{l: l \in S, l \mbox{ occurs in } agg\}$, i.e., $S \cap Base(agg)$, and $fa(agg,S) = Base(agg) - S$. We have the following lemma whose correctness is immediate. 

\begin{lemma}[Satifiability of An Aggregate Atom] \label{lm:aggSufficientSatisfaction}
Given a set $S$ of ground regular literals and an aggregate e-atom $agg$,  
if $S \models agg$, for any $T$ such that $ta(agg,S) \subseteq T$ and $T \cap fa(agg, S)=\emptyset$, $T \models agg$.   
\end{lemma}

\hide{
\noindent The follow lemma says that if $S$ satisfies an aggregate atom $agg$, we can shrink $S$ to any set $C$, and as long as $C$ satisfies the atoms of $agg$ that occur in $S$, $C$ still satisfies $agg$. This result is used in proving that an $\mathcal{A}log$ answer set is an $\mathcal{F}log$ answer set. 

\begin{lemma} \label{lm:shrink}
Given a set $S$ of ground regular literals, $C \subseteq S$ and an aggregate atom $agg$,  
if $S \models agg$ and $C \models ta(agg, S)$, then $C \models agg$. 
\end{lemma}

\noindent Proof. Let $agg$ be of the form $f\{x: p(x)\} \odot n$. 

\st Since $C \subseteq S$, 

\st \ml{shr1} $\{x: p(x) \in C\} \subseteq \{x: p(x) \in S\}$. 

\st Since $C \models ta(agg, S)$, 

$ta(agg, S) \subseteq C$, i.e., 

\st \ml{shr2} $\{p(x): p(x) \in S\} \subseteq C$, which implies

\st \ml{shr3} $\{x: p(x) \in S\} \subseteq \{x: p(x) \in C \}$, which, together with \mr{shr1}, implies 

\st \ml{shr4} $\{x: p(x) \in S\} = \{x: p(x) \in C\}$.

\st Since $S \models agg$, 

\st \ml{shr5} $f\{x: p(x)\in S\} \odot n$ is true, which, together with \mr{shr4}, implies 

$f\{x: p(x)\in C\} \odot n$ is true, i.e., 

\st \ml{shr9} $C \models agg$. \hfill $\Box$
}

\noindent The following lemma is also useful. 

\begin{lemma}[Monotonicity of $K^P_S$ {\cite{SonP07}}] \label{lm:KMPmonotone}
Let $P$ be an $\mathcal{S}log$ program and $S$ be a set of ground regular atoms. Then, $K^P_
S$ is monotone and
continuous over the lattice $\langle 2^{B_P}, \subseteq \rangle$, where $B_P$ is the Herbrand base of $P$. 
\end{lemma}

\noindent Given an $\mathcal{S}log$ program, we first present a result relating $K_A^P$ operator to the classical consequence operator on $R_{\mathcal{A}}(P, A)^A$, denoted by $T_{R_{\mathcal{A}}(P, A)^A}$. 

\setcounter{statement}{0}

\begin{lemma}[$K_A^P$ Operator and The Classical Consequence Operator]
\label{lm:a2sfixedPoint}
Given a program $P$ of $\mathcal{S}log$ syntax and a set $A$ of regular atoms, for any set $S_1$ of regular atoms,  $T_{R_{\mathcal{A}}(P, A)^A} (S_1)   \subseteq K_A^P(S_1)$. 
\end{lemma}

\noindent Proof. Without loss of generality we assume $P$ is a ground program. 

\st Let $P^\prime = R_{\mathcal{A}}(P, A)^A$. 

\st For any $a \in T_{P^\prime}(S_1)$, we prove $a \in K_A^P(S_1)$ \mr{as11z}. 

\st Since $a \in T_{P^\prime}(S_1)$, there exists a rule $r'$ of $P^\prime$ such that 

\st \ml{as111} $head(r') = a$, and 

\st \ml{as112} $S_1 \models body(r')$.

\st Since $r' \in P^\prime$, there exists $r \in P$ such that $r'$ is obtained from $r$. Let $r$ be of the form 

\st \ml{as113} $a$ :- $posReg(r), negReg(r), aggs(r)$, where $posReg(r)$ denotes the positive regular atoms, $negReg(r)$ the negative regular atoms and $aggs(r)$ the aggregate e-atoms. 

\st Since $r' \in P^\prime$, 

\st \ml{as114} $A \models negReg(r)$, and 

\st \ml{as115} $A \models aggs(r)$.  

\st \mr{as114} and \mr{as115} imply that the form of $r'$ is 

\st \ml{as116} $a$ :-  $posReg(r), faggs(r)$, where $faggs(r) = \cup_{agg \in aggs(r)} ta(agg, A)$.

\st By \mr{as114}, there exists a rule $r'' \in~ ^{A}\!P$, obtained from $r$, of the form:

\st \ml{as117} $a$ :- $posReg(r), aggs(r)$.

\st By \mr{as112} and \mr{as116}, $S_1 \models posReg(r)$, which implies 

\st \ml{as118} $(S_1, A) \models posReg(r)$.

\st For any aggregate atom $agg \in aggs(r)$, \mr{as112} and \mr{as116} implies 

\st \ml{as119} $S_1 \models ta(agg,A)$. 

\st \mr{as115} implies 

\st \ml{as1191} $A \models agg$, and thus

\st \ml{as11a} $A \models ta(agg, A)$, which, together with \mr{as119}, implies

\st \ml{as11b} $(S_1 \cap A) \models ta(agg, A)$, which, together with the definition of $ta(agg, A)$, implies 

\st \ml{as11c} $(S_1 \cap A \cap Base(agg)) \models ta(agg, A)$

\st For any $S$ such that 

\st \ml{as11d1}$S_1 \cap A \cap Base(agg) \subseteq S$, and

\st \ml{as11d2} $S \cap (Base(agg)-A) = \emptyset$,

\st we will show $S \models agg$ \mr{as11g}.

\st First, we have 

\st \ml{as11e} $Base(agg) - A$
\begin{minipage}[t]{\textwidth}
= $Base(agg) - (A \cap Base(agg))$ by definition of $-$. \\
= $Base(agg) - ta(agg, A)$ by definition of $ta(agg,A)$. 
\end{minipage}

\st \ml{as11f} $Base(agg) - (S_1 \cap A \cap Base(agg))$

$~~~~~~$
\begin{minipage}[t]{\textwidth}
 $ \subseteq Base(agg) - ta(agg, A)$ because $ta(agg, A) \subseteq S_1 \cap A \cap Base(agg)$ (by \mr{as11c})
 
 $= Base(agg) - A$ by $Base(agg) - A = Base(agg) - ta(agg,A)$ \mr{as11e}
 
 $\subseteq Base(agg) -(S_1 \cap A \cap Base(agg))$ because $(S_1 \cap A \cap Base(agg)) \subseteq A$.
\end{minipage}

\st Therefore, 


\st \ml{as11f0} $Base(agg) - (S_1 \cap A \cap Base(agg)) = Base(agg) - A$, which also implies 

\st \ml{as11f01} $Base(agg) - (S_1 \cap A \cap Base(agg)) = Base(agg) - ta(agg,A)$ 

\st because $Base(agg) - A = Base(agg) - ta(agg,A)$ \mr{as11e}. Therefore,

\st \ml{as11f1} $(S_1 \cap A \cap Base(agg)) \cap (Base(agg) - ta(agg, A)) = \emptyset$.

\st By 
\mr{as11c} and \mr{as11f1}, we have

\st \ml{as11f2} $(S_1 \cap A \cap Base(agg)) \models agg$. 

\st By \mr{as11d2} and \mr{as11f0},

\st \ml{as11f3} $S \cap (Base(agg) - (S_1 \cap A \cap Base(agg))) = \emptyset$

\st By Lemma~\ref{lm:aggSufficientSatisfaction}, \mr{as11f2}, \mr{as11d1} and \mr{as11f3} imply 

\st \ml{as11g}  
$S \models agg$. 

\st Therefore,

$\langle S_1 \cap A \cap Base(agg), Base(agg)-A \rangle$ is a solution of $agg$. Hence, 

\st \ml{as11h} $(S_1, A) \models agg$. Therefore, 

\st $(S_1, A) \models aggs(r)$, which, together with \mr{as118} and \mr{as117}, implies 

\st \ml{as11i} $(S_1, A) \models body(r'')$. 

\st By definition of $K_A^P$, \mr{as11i} and \mr{as117},

\st \ml{as11z} $a \in K_A^P(S_1)$. Hence,

\st \ml{as11} $T_{P^\prime} (S_1) \subseteq K_A^P(S_1)$. \hfill $\Box$

\setcounter{statement}{0}

\begin{claim}[From $\mathcal{A}log$ Answer Sets to $\mathcal{S}log$ Answer Sets]
Consider an $\mathcal{S}log$ program $P$ without multisets in its rules. If $A$ is an $\mathcal{A}log$ answer set of $P$, it is an $\mathcal{S}log$ answer set of $P$.
\end{claim}

\noindent Proof. Assuming

\st \ml{a2s11} $A$ is an $\mathcal{A}log$ answer set of $P$,

\st we prove $A$ is an $\mathcal{S}log$ answer set of $P$ \mr{a2s9}, by showing first $A$ is a fixed point of $K_A^P$ \mr{a2s5} and then the least fixed point of $K_A^P$  \mr{a2s7}. 

\st To show \mr{a2s5}, we will show $K^P_A(A) \subseteq A$.  In turn, for any rule $r' \in~ ^A\!P$, assuming 

\st \ml{a2s12} $(A, A) \models body(r')$, 

\st we prove $head(r') \in A$ \mr{a2s4x}.

\st Since $r' \in~ ^A\!P$, there exists $r \in P$ such that $r'$ is obtained from $r$. Let the form of $r$ be 

\st \ml{a2s21} $head(r) :- posReg(r), negPos(r), aggs(r)$.

\st Since $r' \in~ ^A\!P$, 

\st \ml{a2s22} $A \models negReg(r)$. 

\st For any $agg \in aggs(r)$, by $(A, A) \models body(r')$ \mr{a2s12},

\st \ml{a2s23} $(A, A) \models agg$. Therefore, 

\st \ml{a2s24} $A \models agg$. Hence, 

\st \ml{a2s241} $A \models ta(agg,A)$. 

\st \mr{a2s24} implies 

\st \ml{a2s251} $A \models aggs(r)$, which, together with \mr{a2s22}, implies

\st \ml{a2s26} there exists $r'' \in P^\prime$, where $P^\prime = R_{\mathcal{A}}(P, A)^A$, such that $r''$ is obtained from $r$. The form of $r''$ is 

\st \ml{a2s27} $head(r)$ :- $posReg(r), \cup_{agg\in aggs(r)} ta(agg, A)$. 

\st \mr{a2s241} implies 

\st \ml{a2s25} $A \models \cup_{agg \in aggs(r)}ta(agg, A)$.

\st By $(A, A) \models body(r')$ \mr{a2s12}, 

\st \ml{a2s28} $A \models posReg(r)$, which, together with \mr{a2s25} and \mr{a2s27}, implies 

\st \ml{a2s3} $A \models body(r'')$. 

\st Since $A$ is an $\mathcal{A}log$ answer set of $P$ \mr{a2s11}, $A$ is a model of $P^\prime$. So, $A \models r''$, which, together with \mr{a2s3}, implies 

\st \ml{a2s31} $A \models head(r'')$, which, together with $head(r'') = head(r')$, implies 

$A \models head(r')$, i.e., 

\st \ml{a2s4x} $head(r') \in A$.  Hence, $K^P_A(A) \subseteq A$, which, together with $A \subseteq K^P_A(A)$ by Lemma~\ref{lm:KMPmonotone}, implies

\st \ml{a2s5}$A$ is a fixed piont of $K_A^P$.

\st We next prove that $A$ is the least  \mr{a2s7} by contradiction. Assume, there exists $B$ such that 

\st \ml{a2s51} $B \subset A$, and 

\st \ml{a2s52} $B$ is a fixed point of $K_A^P$, i.e., $B = K_A^P(B)$. 

\st By Lemma~\ref{lm:a2sfixedPoint}, $T_{P^\prime}(B) \subseteq K_A^P(B)$, which, together with \mr{a2s52}, 

\st \ml{a2s53} $T_{P^\prime}(B) \subseteq B$. 

\st Since $T_{P^\prime}$ is monotone, $B \subseteq T_{P^\prime}(B)$, which, together with \mr{a2s53}, implies

\st \ml{a2s54} $B = T_{P^\prime}(B)$. Hence 

\st \ml{a2s55} $B$ is a model of $P^\prime$, which, together with $B \subset A$ \mr{a2s51}, implies

\st \ml{a2s56} $A$ is not a minimal model of $P^\prime$.

\st Since $A$ is an $\mathcal{A}log$ answer set of $P$ \mr{a2s11}, $A$ is a minimal model of $P'$, contradicting \mr{a2s56}. 
Therefore, 

\st \ml{a2s7} $A$ is the least fixed point of $K_A^P$, i.e., 

\st \ml{a2s9} $A$ is an $\mathcal{S}log$ answer set of $P$.

\hfill $\Box$



\setcounter{statement}{0}
\st The following lemma will  be useful in proving Proposition~\ref{prop:AlogEqSlog}. 

\begin{lemma} [From $\mathcal{S}log$ to $\mathcal{F}log$ on Stratified Programs]\label{lm:Slog2Flog}
Given an aggregate stratified $\mathcal{S}log$ program $P$, $A$ is an $\mathcal{F}log$ answer set of $P$ if $A$ is an $\mathcal{S}log$ answer set of $P$.
\end{lemma}

\noindent Proof. 

\st Assuming 

\st \ml{af:1} $A$ is an $\mathcal{S}log$ answer set of $P$,

\st we prove $A$ is an $\mathcal{A}log$ answer set of $P$ \mr{af:4z1}.

\st By Lemma~\ref{lm:answerSetsStrata} and \mr{af:1},

\st \ml{af:11} for any $i( \ge 1)$, $A_i$ is an $\mathcal{S}log$ answer set of $\Pi_i$. 

\st We will show, for any $i \ge 1$, $A_i$ is an $\mathcal{F}log$ answer set of $\Pi_i$ \mr{af:4z}, by first showing that $A_i$ is a model of $R_{\mathcal{F}}(\Pi_i, A_i)$ \mr{af:4f} and then proving it is minimal \mr{af:4m}.

\hide{
\st When $i=1$, there is no aggregate e-atoms in $\Pi_i$ because the program is aggregate stratified. For a program with aggregate atoms, $\mathcal{S}log$ and $\mathcal{A}log$ agree with the classical answer set semantics. Therefore, 
$A_i$ is an $\mathcal{A}log$ answer set of $\Pi_1$ because $A_i$ is an $\mathcal{S}log$ answer set of $\Pi_1$.}

\st For any rule $r' \in R_{\mathcal{F}}(\Pi_i, A_i)$, if 

\st \ml{af:4e1} $A_i \models body(r')$, then we prove $A_i \models head(r')$ \mr{af:4ez}.

\st Since $r' \in R_{\mathcal{F}}(\Pi_i, A_i)$, there exists $r \in \Pi_i$ such that $r'$ is obtained from $r$. Let $r$ be of the form: 

\st \ml{af:4e11} $head(r)$ :- $posReg(r), negReg(r), aggs(r)$. 

\st Since $r' \in R_{\mathcal{F}}(\Pi_i, A_i)$, we have  

\st \ml{af:4e2} $A_i \models posReg(r)$,

\st \ml{af:4e3} $A_i \models negReg(r)$, and 

\st \ml{af:4e4} $A_i \models aggs(r)$.

\st By \mr{af:4e3} and definition of $^{A_i}P_i$, there exists a rule $r'' \in~  ^{A_i}P_i$ of the form:   

\st \ml{af:4e41} $head(r)$ :- $posReg(r), aggs(r)$.   

\st We now prove $(A_i, A_i) \models body(r'')$ \mr{af:4eo}.

\st We first prove, for any $agg \in aggs(r)$, $(A_i, A_i) \models agg$ \mr{af:4em}.

\st For any $S$ such that 

\st \ml{af:4e51} $A_i \cap Base(agg) \subseteq S$, and 
  
\st \ml{af:4e52} $S \cap (Base(agg) - A_i) = \emptyset$,
  
\st we prove \mr{af:4e9}.

\st \mr{af:4e4} implies that $A_i \cap Base(agg) \models agg$, which, together with \mr{af:4e51} and \mr{af:4e52} and Lemma~\ref{lm:aggSufficientSatisfaction},implies

\st \ml{af:4e9} $S \models agg$. Therefore, by definition of aggregate solution, we have 

\st \ml{af:4ea} $\langle A_i \cap Base(agg), Base(agg) - A_i \rangle$ is a solution of $agg$. 

\st Hence, by definition of conditional satisfaction,


\st \ml{af:4em} $(A_i, A_i) \models agg$. Therefore,

\st \ml{af:4en} $(A_i, A_i) \models aggs(r)$. 

\st \mr{af:4e2} implies $(A_i, A_i) \models posReg(r)$, which, together with \mr{af:4en} and the form \mr{af:4e41} of $r''$, implies 

\st \ml{af:4eo} $(A_i, A_i) \models body(r'')$.  

\st Since $A_i$ is an $\mathcal{S}log$ answer set of $\Pi_i$ \mr{af:11}, $K_{A_i}^{\Pi_i}(A_i) = A_i$, which, together with \mr{af:4eo} and the definition of $K_{A_i}^{\Pi_i}$, implies $head(r) \in A_i$, i.e., $A_i \models head(r)$. Since $head(r')=head(r)$, 

\st \ml{af:4ez} $A_i \models head(r')$. Hence,

\st \ml{af:4f} $A_i$ is a model of $R_{\mathcal{F}}(\Pi_i, A_i)$. 

\st We next prove $A_i$ is a minimal model of $R_{\mathcal{F}}(\Pi_i, A_i)$ \mr{af:4m} by contradiction. Assume there exists $B$ such that 

\st \ml{af:4h1} $B \subset A_i$, and 

\st \ml{af:4h2} $B$ is a model of $R_{\mathcal{F}}(\Pi_i, A_i)$.

\st Since $K_{A_i}^{\Pi_i}$ is monotone (by Lemma~\ref{lm:KMPmonotone}), $B \subseteq K_{A_i}^{\Pi_i}(B)$, which, together with $A_i = lfp(K_{A_i}^{\Pi_i})$, implies the existence of a rule $r' \in~ ^{A_i}{\Pi_i}$ such that 

\st \ml{af:4h3} $(B, A_i) \models body(r')$, and 

\st \ml{af:4h4} $B \not \models head(r')$. 

\st Since $r' \in~ ^{A_i}{\Pi_i}$, there exists $r \in \Pi_i$ such that $r'$ is obtained from $r$. 
Let the form of $r$ be 

\st \ml{af:4h5} $head(r)$ :- $posReg(r), negReg(r), aggs(r)$. 

\st Since $r' \in~ ^{A_i}{\Pi_i}$, 

\st \ml{af:4h51} $A_i \models negReg(r)$, which, together with \mr{af:4h1}, implies 

\st \ml{af:4h511} $B \models negReg(r)$. 

\st By \mr{af:4h3},  

\st \ml{af:4h52} $(B, A_i) \models posReg(r)$, and 

\st \ml{af:4h53} $ (B, A_i) \models aggs(r)$. 

\st By definition of conditional satisfiability and that $posReg(r)$ contains only positive regular atoms, \mr{af:4h52} implies 

\st \ml{af:4h54} $B \models posReg(r)$. 

\st For any $agg \in aggs(r)$, \mr{af:4h53} implies 

\st \ml{af:4h55} $(B, A_i) \models agg$, which implies that 

\st \ml{af:4h56} $\langle B \cap A_i \cap Base(agg), Base(agg) - A_i \rangle$ is a solution of $agg$. 

\st Since $B \subset A_i$ \mr{af:4h1}, 

\st \ml{af:4h57} $B \cap (Base(agg) - A_i) = \emptyset$, which, together with $B \cap A_i \cap Base(agg) \subseteq B$ and \mr{af:4h56}, implies 

\st \ml{af:4h6} $B \models agg$. Therefore, 

\st \ml{af:4h61} $B \models aggs(r)$, which, together with \mr{af:4h54} and \mr{af:4h511}, implies 

\st \ml{af:4h62} $B \models body(r)$, which implies $r \in R_{\mathcal{F}}(\Pi_i, A_i)$. 

\st Since $B$ is a model of $R_{\mathcal{F}}(\Pi_i, A_i)$ \mr{af:4h2}, 

\st \ml{af:4h63} $B \models r$, which, together with \mr{af:4h62} implies 

\st \ml{af:4h64} $B \models head(r)$. 

\st Since $head(r) = head(r')$, \mr{af:4h64} implies 

\st \ml{af:4h65} $B \models head(r')$,  contradicting $B \not \models head(r')$ \mr{af:4h4}. Hence, 

\st \ml{af:4m} $A_i$ is a minimal model of $R_{\mathcal{F}}(\Pi_i, A_i)$. Therefore, 

\st \ml{af:4z} $A_i$ is the $\mathcal{F}log$ answer set of $R_{\mathcal{F}}(P_i, A_i)$. Hence, 

\st \ml{af:4z1} 
$A$ is an $\mathcal{F}log$ answer set of $P$, by Lemma~\ref{lm:answerSetsStrata}. 
\hfill $\Box$

\setcounter{statement}{0}

\begin{claim} [Equivalence of $\mathcal{A}log$ and
   $\mathcal{S}log$ Semantics for Aggregate Stratified Programs] \label{prop:AlogEqSlog}
Given an aggregate stratified $\mathcal{S}log$ program $P$, $A$ is an $\mathcal{S}log$ answer set of $P$ iff $A$ is an $\mathcal{A}log$ answer set of $P$. 
\end{claim}

\noindent Proof. 

\st $\Longrightarrow$: Since $A$ is an $\mathcal{S}log$ answer set of $P$, $A$ is an $\mathcal{F}log$ answer set of $P$ by Lemma~\ref{lm:Slog2Flog}. By the equivalence of $\mathcal{F}log$ and $\mathcal{A}log$ on aggregate stratified programs ( Proposition~\ref{prop:AlogEqFlog}), $A$ is an $\mathcal{A}log$ answer set of $P$. 

\st $\Longleftarrow$: this is a direct result of Proposition~\ref{prop:a2slog}. 

\hfill $\Box$

\end{document}